\documentclass{article} %
\usepackage{iclr2025_conference,times}

\usepackage[backref=page]{hyperref}
\usepackage{url}

\usepackage{algorithm}
\usepackage{algpseudocode}

\usepackage{amsmath}
\usepackage{amsfonts}
\usepackage{amssymb}
\usepackage{amsthm}
\usepackage{bm}

\usepackage{graphicx}
\usepackage{caption}
\usepackage{subcaption}
\usepackage{booktabs}

\usepackage{tikz}
\usetikzlibrary{decorations.pathmorphing}

\usepackage[T1]{fontenc}

\newtheorem{manualtheoreminner}{Question}
\newenvironment{question}[1]{%
  \IfBlankTF{#1}
    {\renewcommand{\themanualtheoreminner}{\unskip}}
    {\renewcommand\themanualtheoreminner{#1}}%
  \manualtheoreminner
}{\endmanualtheoreminner}

\definecolor{darkgreen}{rgb}{0.0, 0.5, 0.0}
\definecolor{darkorange}{rgb}{0.99, 0.54, 0.0}

\newcommand{\vth}{\bm{\theta}}

\newcommand{\vm}{\bm{m}}

\newcommand{\vg}{\bm{g}}

\newcommand{\cD}{\mathcal{D}}

\newcommand{\bbE}{\mathbb{E}}

\newcommand{\vgo}{\vg^{O}}

\title{Beyond the Boundaries of Proximal Policy \\ Optimization}

\def\myspace{\hskip0.7\fontdimen6\font}

\author{
Charlie B. Tan\textsuperscript{1}
\myspace
Edan Toledo\textsuperscript{2,3}
\myspace
Benjamin Ellis\textsuperscript{1}
\myspace
Jakob N. Foerster\textsuperscript{1}
\myspace
Ferenc Huszár\textsuperscript{4}\\
\textsuperscript{1}University of Oxford
\myspace
\textsuperscript{2}University College London
\myspace
\textsuperscript{3}Meta
\myspace
\textsuperscript{4}University of Cambridge
}

\iclrfinalcopy %
\begin{document}

\maketitle

\begin{abstract}

Proximal policy optimization (PPO) is a widely-used algorithm for on-policy reinforcement learning. This work offers an alternative perspective of PPO, in which it is decomposed into the inner-loop \emph{estimation} of update vectors, and the outer-loop \emph{application} of updates using gradient ascent with unity learning rate. Using this insight we propose \emph{outer proximal policy optimization} (outer-PPO); a framework wherein these update vectors are applied using an arbitrary gradient-based optimizer. The decoupling of update estimation and update application enabled by outer-PPO highlights several implicit design choices in PPO that we challenge through empirical investigation. In particular we consider non-unity learning rates and momentum applied to the outer loop, and a momentum-bias applied to the inner estimation loop. Methods are evaluated against an aggressively tuned PPO baseline on Brax, Jumanji and MinAtar environments; non-unity learning rates and momentum both achieve statistically significant improvement on Brax and Jumanji, given the same hyperparameter tuning budget.

\end{abstract}

\section{Introduction}
\label{sec:introduction}

Proximal policy optimization (PPO) \citep{schulman_proximal_2017} is ubiquitous within modern reinforcement learning (RL), having found success in domains such as robotics \citep{andrychowicz2020learning}, gameplay \citep{berner2019dota}, and  research applications \citep{mirhoseini2021graph}. Given it's ubiquity, significant research effort has explored the theoretical \citep{hsu_revisiting_2020, kuba_mirror_2022} and empirical \citep{engstrom_implementation_2020, andrychowicz_what_2020} properties of PPO.

PPO is an on-policy algorithm; at each iteration it collects a dataset using the current (behavior) policy. This dataset is used to construct a surrogate to the true objective, enabling gradient-based optimization while seeking to prevent large changes in policy between iterations, similar to trust region policy optimization \citep{schulman_trust_2017}. The solution to the surrogate objective is then taken as the behavior parameters for the following iteration, defining the behavior policy with which to collect the following dataset. The behavior policies are therefore \emph{exactly coupled} with the preceding surrogate objective solution.

\begin{figure}[t]
      \tikzset{fixed size node/.style={rectangle, minimum width=0.5cm, minimum height=0.5cm, text centered}}
  
\begin{tikzpicture}[line width=1.5pt,>=latex, every node/.style={}, xscale=0.7, yscale=0.7]
 
  \begin{scope}
  [shift={(0,0)}]

  \node[] at (-2.55,-1.6) {\bf (i)};

  \fill[blue!20,opacity=0.5] (0,0) ellipse (2.75cm and 1.75cm);

  \node[fixed size node] (thetak) at (0,0) {$\vth_k$};
  \node[fixed size node] (thetastar) at (2.2,-0.5) {$\vth_k^*$};

    \draw[blue, ->, dashed] (thetak) .. controls (0.9,2) and (2,1) .. (1.2,0) .. controls (-0.5,-1.5) and (1,-1.7) .. (2, -0.8);

  \end{scope}
  
    \begin{scope}
  [shift={(6.3,0)}]
  
  \node[] at (-2.55,-1.6) {\bf (ii)};
  
  \fill[blue!20,opacity=0.5] (0,0) ellipse (2.75cm and 1.75cm);

  \node[fixed size node] (thetak) at (0,0) {$\vth_k$};
  \node[fixed size node] (thetastar) at (2.2,-0.5) {$\vth_k^*$};

  \draw[red,->, shorten >= -1mm] (thetak) -- (thetastar) node[pos=0.5, below, xshift=-1mm] {\(\vgo_k\)};

  \end{scope}

   \begin{scope}[shift={(12.6,0)}]  %
  
  \node[] at (-2.55,-1.6) {\bf (iii)};
  
  \fill[blue!20,opacity=0.5] (0,0) ellipse (2.75cm and 1.75cm);

  \node[fixed size node] (thetak) at (0,0) {$\vth_k$};
  \node[fixed size node] (thetakp1) at (3.65,-0.83) {$\vth_{k+1}$};

  \draw[darkgreen,->, shorten >= -1mm] (thetak) -- (thetakp1) node[pos=0.5, below, xshift=-1mm] {\(\sigma\vgo_k\)};

  \draw[white] (0,-1.8) -- (0, -2.2);
  
  \end{scope}
   
\end{tikzpicture}
    \caption{{\bf Diagram of outer-PPO estimating and applying the outer gradient as an update.} (i) Transitions are collected with policy \(\pi(\vth_k)\) defining a surrogate objective and corresponding `trust-region' (shaded) surrounding \(\vth_k\); inner-loop optimization of the surrogate objective (blue dashed) yields \(\vth_k^*\). (ii) Outer-PPO computes outer gradient as \(\vgo_k \gets \vth_k^*-\vth_k\). (iii) Outer-PPO updates behavior parameters using an arbitrary gradient based optimizer applied to the outer gradient to give \(\vth_{k+1}\), in this case gradient ascent with a learning rate \(\sigma > 1\). Standard PPO can be understood as directly taking \(\vth_{k+1} \gets \vth_k^*\), or as a special case of outer-PPO corresponding to gradient ascent with learning rate \(\sigma = 1\).}
    \label{fig:intro_diagram}
\end{figure}
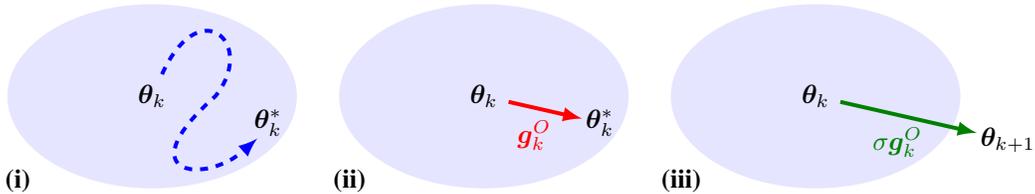

In this work we instead consider the inner-loop optimization of each surrogate objective to estimate an update vector, which we name the \textit{outer gradient}. A trivial result follows that the outer loop of PPO can be viewed to update the behavior parameters using unity learning rate \(\sigma = 1\) gradient ascent on the outer gradients. Using this insight we propose outer-PPO, a novel variation of PPO that employs an arbitrary gradient-based optimizer in the outer loop of PPO. Outer-PPO \emph{decouples} the estimation and application of updates in way not possible in standard PPO. An illustration of outer-PPO applying a learning rate greater than unity is provided in figure \ref{fig:intro_diagram}. The new behaviors enabled by outer-PPO raise several questions related to implicit design choices of PPO:

\begin{question}{1}
Is the unity learning rate always optimal?
\end{question}
\begin{question}{2}
Is the independence (lack of prior trajectory information e.g momentum) of each outer update step always optimal?
\end{question}
\begin{question}{3}
Is initializing the inner loop surrogate objective optimization at the behavior parameters (without exploiting prior trajectory / momentum) always optimal? 
\end{question}

\clearpage

This work forms an empirical investigation of the aforementioned questions. To motivate this investigation, consider the clipping parameter \(\epsilon\) of PPO, controlling the size of the `trust region' within which we seek to restrict our update. If \(\epsilon\) is set too low, we restrict ourselves to small policy updates. Conversely, if we set \(\epsilon\) too high we decrease the reliability of our update direction. In outer-PPO, introducing an outer learning rate decouples these two effects; we are able to reliably estimate an update vector using a moderate \(\epsilon\), but then take step of large magnitude in this direction.

We emphasize that we do not seek to identify the most performant configuration possible but to understand the performance of outer-PPO relative to a well-tuned PPO baseline. To this end we restrict the tuning of outer-PPO to simple grid searches applied to fixed base PPO hyperparameters.

Our contributions are as follows:

\begin{itemize}
    \item We propose \emph{outer proximal policy optimization} (outer-PPO), in which an arbitrary gradient-based optimizer is applied to the `outer gradients' of PPO. By tracking the outer trajectory, outer-PPO further permits a momentum bias to be applied to the inner-loop initialization.
    \item We optimize a PPO baseline through extensive hyperparameter sweeps (total of 38,400 agent trained) on subsets of Brax (6 tasks) \citep{freeman_brax_2021}, Jumanji (4 tasks) \citep{bonnet2024jumanji}, and MinAtar (4 tasks) \citep{young_minatar_2019}. We open-source the sweep database files to facilitate future research against strongly tuned baselines.
    \item We perform three lightweight outer-PPO grid searches on non-unity outer learning rates, outer Nesterov momentum and biased-initialization, each addressing questions 1, 2 and 3 respectively.
    \item We evaluate the outer-PPO methods against the baseline, using 64 seeds per task over the 14 different tasks. We find non-unity outer learning rates to yield the greatest improvement (5-10\%) on both Brax and Jumanji. Outer Nesterov also improves performance on Brax and Jumanji. Biased initialization achieves a moderate improvement on Jumanji alone. No method improves over the baseline on MinAtar.
    \item Given the stated empirical results we conclude the \emph{negative} for questions 1, 2 and 3. Relaxing each of these PPO design choices can lead to consistent, statistically significant improvement of performance over at least one of the evaluated environment suites.
    \item We propose that practitioners able to experiment may explore non-unity outer learning rates given the simplicity (single hyperparameter) and consistent improvement achieved on Brax and Jumanjji.
\end{itemize}

\section{Background}
\label{sec:background}

\subsection{Reinforcement Learning}

We consider the standard reinforcement learning formulation of a Markov decision process \(\mathcal{M} = \langle \mathcal{S}, \mathcal{A}, \mathcal{T}, r, \gamma \rangle\), where \(\mathcal{S}\) is the set of states, \(\mathcal{A}\) is the set of actions, \(\mathcal{T}: \mathcal{S} \times \mathcal{A}  \rightarrow \Delta(\mathcal{S})\) is the state transition probability function, \(r: \mathcal{S} \times \mathcal{A} \rightarrow \Delta(\mathbb{R})\) is the reward function, and \(\gamma \in [0, 1]\) is the discount factor. We use the notation \(\Delta(\bm{X})\) to denote the probability distribution over a set \(\bm{X}\). The reinforcement learning objective is to maximize the expected return \(\bbE_\pi[G_t] = \bbE_\pi[\sum_{t}\gamma^tr_t]\) given a \emph{policy} \(\pi : \mathcal{S} \rightarrow \Delta(\mathcal{A})\) defining the agent behavior. In actor-critic policy optimization the policy is explicitly represented as a parametric function \(\pi : \mathcal{S} \times \vth^\pi \rightarrow \Delta(\mathcal{A})\), and a \emph{value function} \(V:\mathcal{S} \times \vth^V \rightarrow \mathbb{R}\) is employed to guide optimization. In deep RL ~\citep{mnih2015human, silver2017mastering} neural networks are used for the policy and  value functions, for ease of notation we consider \(\vth \in \mathbb{R}^{(d_\pi + d_V)}\) as the concatenation of the respective weight vectors.

\subsection{Proximal Policy Optimization}

Proximal policy optimization was proposed by \citet{schulman_proximal_2017}, and has since become one of the most popular algorithms for on-policy reinforcement learning. At each iteration \(k\) a dataset of transitions \(\cD_k\) is collected using policy \(\pi(\vth_k)\), and advantages \(\hat{A}_k\) are estimated using generalized advantage estimation (GAE) \citep{schulman_high-dimensional_2018}.
The transition dataset and advantages are then used within an inner optimization loop, in which the policy parameters \(\vth^\pi\) are optimized with respect to a given surrogate objective along with the value parameters \(\vth^V\).
Psuedocode for a single iteration of PPO is provided in algorithm \ref{alg:ppo}, where \textproc{InnerOptimizationLoop} is defined in appendix \ref{app:ppo_details}. The full algorithm updates parameters iteratively by \(\vth_{k+1} \gets \textproc{PPOIteration}(\vth_k)\).

\begin{algorithm}[H]
    \caption{Proximal policy optimization iteration}
    \begin{algorithmic}[1]
    \Function{PPOIteration}{$\vth$}
    \State Collect set of trajectories \(\cD\) by running policy \(\pi(\vth)\)
    \State Estimate advantages \(\hat{A}\) with GAE.
    \State \(\vth^* \gets \textproc{InnerOptimizationLoop}(\vth, \cD, \hat{A})\)
    \State \Return \(\vth^*\) 
    \EndFunction
    \end{algorithmic}
    \label{alg:ppo}
\end{algorithm}

PPO permits the use of any arbitrary surrogate objective, though it is most commonly associated with the \emph{clipped objective} \citet{schulman_proximal_2017} stated in equation \ref{eq:clip_policy}.

\begin{equation}
\label{eq:clip_policy}
L^\pi(\vth^\pi) = \mathbb{E}_{s,a \sim \mathcal{D}_k} \left[ \min \left( \rho(\vth^\pi) \hat{A}, \ \text{clip}(\rho(\vth^\pi), 1 - \epsilon, 1 + \epsilon) \hat{A} \right) \right]
\end{equation}

Here \(\rho(\vth^\pi) = \frac{\pi(a | s)}{\pi_k(a | s)}\) is the ratio between our current policy \(\pi\) and the behavior policy \(\pi_k\), and \(\epsilon\) is the clipping threshold. The value function is similarly optimized using either simple regression. \(L^{V}(\vth^V) = (V_{\vth_k} - V_\text{targ})^2\) or the clipped objective defined in appendix \ref{app:ppo_details}.

\subsection{Trust regions}

A \emph{trust region} is a region surrounding an optimization iterate \(\vth_k\) within which we permit our algorithm to update the parameters to \(\vth_{k+1}\). In TRPO, a trust region surrounding the behavior parameters is explicitly defined as the region in parameter space \(\vth \in \Theta\) satisfying 
\(\mathbb{E}_{s\sim \cD_k} \left[D_{\text{KL}}\left(\pi(\vth_k|s) \parallel \pi(\vth|s) \right) \right] \leq \delta\). Optimizing subject to this constraint prevents large changes in the policy between successive iterations, and gives rise to a guarantee of \emph{monotonic improvement}. Similarly, if the clipped surrogate objective of PPO is replaced with a KL penalty \(L^\pi(\vth) = \mathbb{E}_{s,a\sim \cD_k}[\rho(\vth) \hat{A} - \beta D_{\text{KL}}\left(\pi(\vth_k|s) \parallel \pi(\vth|s) \right) ]\), a trust-region is implicitly defined for some \(\delta\). Both TRPO and PPO-KL approximate the natural policy gradient \citep{kakade_natural_2001}, \citep{hsu_revisiting_2020}; the steepest direction in the non-Euclidean geometry of policy space induced by the Fisher information metric.

Unlike the KL penalized surrogate, the clipped surrogate objective of equation \ref{eq:clip_policy} \emph{does not} define a formal trust region. We can however define the region of non-zero gradients, with gradient defined as in equation \ref{eq:non-zero}.

\begin{equation}
\nabla_{\theta^\pi} L^\pi(\vth^\pi) = \mathbb{E}_{s,a \sim \cD_k} \left[ \hat{A} \nabla_{\vth^\pi } \rho(\vth^\pi) \cdot \mathbb{I}\left( \left| \rho(\vth^\pi) - 1 \right| \leq \epsilon\ \text{or}\ \left( \rho(\vth^\pi) - 1 \right) \hat{A} \leq 0 \right) \right]
\label{eq:non-zero}
\end{equation}

Here \(\mathbb{I}(\cdot)\) is an indicator function that equals 1 if and only if the argument is true, and 0 otherwise. Whilst the subspace \(\nabla_{\vth} L^\pi > 0\) can be considered analogous to a trust region, it is possible to irreversibly step arbitrarily far beyond this region \citep{hsu_revisiting_2020}. Nonetheless, where not ambiguous we shall abuse notation and refer to  \(\nabla_{\vth} L^\pi > 0\) as the trust region of the clipped surrogate. Whilst not defining a formal trust region, the clipped objective enjoys theoretical motivation as a valid drift function in the mirror learning framework \citep{kuba_mirror_2022}, hence also benefits from monotonic improvement and convergence guarantees.

\section{Outer-PPO}

\label{sec:outer_ppo}

In equation \ref{eq:ppo_update} we define the outer gradient of PPO.  

\begin{equation}
\vgo(\vth) = \textproc{PPOIteration}(\vth) - \vth
\label{eq:ppo_update}
\end{equation}

The behavior parameter update of PPO \(\vth_{k+1} \gets \textproc{PPOIteration}(\vth_k)\) can now be equivalently expressed as \(\vth_{k+1} \gets \vth_k + \vgo(\vth_k)\). Evidently, PPO is exactly gradient ascent, with a constant learning rate \(\sigma=1\), on its outer gradients. With this simple result established, we propose a family of methods employing arbitrary optimizers on the PPO outer loop, denoted as \emph{outer-PPO}. As an illustrating example, a comparison of standard PPO and outer-PPO with non-unity learning rates is provided in algorithms \ref{alg:std-ppo} and \ref{alg:outer-lr}. We additionally propose a closely-related method for biasing the inner estimation loop using the prior (outer) trajectory, denoted as \emph{biased initialization}.

\subsection{Outer Learning Rates}

Varying the outer learning rate scales the update applied to the behavior parameters, as defined in algorithm \ref{alg:outer-lr} and illustrated in figure \ref{fig:intro_diagram}. The behavior of scaling the outer gradient can not be directly recovered by varying the PPO hyperparameters.

\vspace{-1em}

\begin{center}
\begin{minipage}{0.49\textwidth}
    \begin{algorithm}[H]
        \caption{Standard PPO}
        \label{alg:std-ppo}
        \begin{algorithmic}[1]
        \item[Input: \(\vth_0\) (parameters)]
        \For{\(k = 0,1,2, \dots\)}
            \State \(\vth^* \gets \textproc{PPOIteration}(\vth_k)\)
            \State \(\vth_{k+1} \gets \vth^*\)
        \EndFor
        \end{algorithmic}
    \end{algorithm}
\end{minipage}
\hfill
\begin{minipage}{0.49\textwidth}
    \begin{algorithm}[H]
        \caption{Outer-LR PPO}
        \label{alg:outer-lr}
        \begin{algorithmic}[1]
        \item[Input: \(\vth_0\) (parameters), \(\sigma\) (outer learning rate)]
        \For{\(k = 0,1,2, \dots\)}
            \State \(\vgo_k \gets \textproc{PPOIteration}(\vth_{k}) - \vth_k\)
            \State \(\vth_{k+1} \gets \vth_{k} + \sigma \vgo_k\)
        \EndFor
        \end{algorithmic}
    \end{algorithm}
\end{minipage}
\hfill
\end{center}

An outer learning rate \(\sigma < 1\) interpolates between the behavior parameters \(\vth_k\) and inner-loop solution \(\vth_k^*\), encoding a lack of trust in the outer gradient estimation. Whilst the magnitude of the outer gradient can be reduced by varying hyperparameters, such as the clipping \(\epsilon\) or number of inner loop iterations, the outer gradients are inherently noisy due to stochastic data collection and inner-loop optimization. PPO is additionally able to irreversibly escape its clipping boundary \citep{engstrom_implementation_2020}, and can drift far from the behavior policy given sub-optimal surrogate objective parameters. These effects motivate the exploration of methods that attenuate the outer updates, irrespective of the outer gradient magnitude. In contrast, a learning rate \(\sigma > 1\) amplifies the update vector, encoding confidence in its direction. Whilst the outer gradient magnitude could be increased by varying the PPO hyperparameters, in particular \(\epsilon\), increasing the size of the trust region may lead the policy to drift to beyond the region of policy space where the dataset \(\cD_k\) collected with policy \(\pi_k\) can be considered representative of the environment dynamics, motivating the amplification of well-estimated outer gradients over increases to trust region size.

\subsection{Momentum}

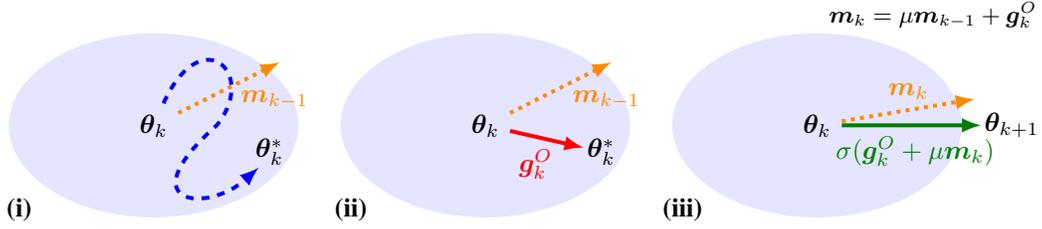
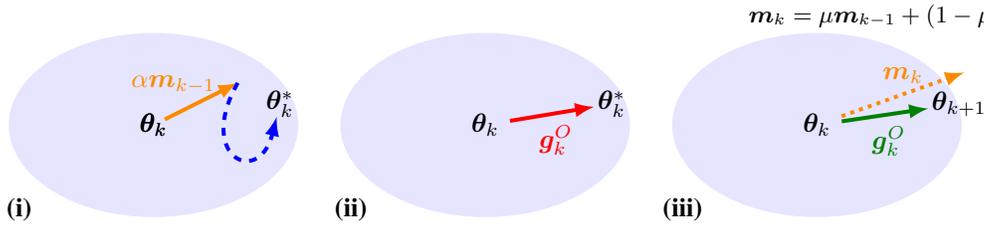
\begin{figure}
    \begin{subfigure}{\textwidth}
        \centering
          \tikzset{fixed size node/.style={rectangle, minimum width=0.5cm, minimum height=0.5cm, text centered}}
  
\begin{tikzpicture}[line width=1.5pt,>=latex, every node/.style={}, scale=0.7]
 
  \begin{scope}
  [shift={(0,0)}]

  \node[] at (-2.55,-1.6) {\bf (i)};

  \fill[blue!20,opacity=0.5] (0,0) ellipse (2.75cm and 1.75cm);

  \node[fixed size node] (thetak) at (0,0) {$\vth_k$};
  \node[fixed size node] (thetastar) at (2.2,-0.5) {$\vth_k^*$};
  
  \coordinate (mom) at (2.4, 1.2);
   \draw[darkorange, ->, dotted] (thetak) -- (mom) node[pos=0.8, below, xshift=2mm, yshift=-1mm] {\(\vm_{k-1}\)};

    \draw[blue, ->, dashed] (thetak) .. controls (0.9,2) and (2,1) .. (1.2,0) .. controls (-0.5,-1.5) and (1,-1.7) .. (2, -0.8);

  \end{scope}
  
    \begin{scope}
  [shift={(6.3,0)}]
  
  \node[] at (-2.55,-1.6) {\bf (ii)};
  
  \fill[blue!20,opacity=0.5] (0,0) ellipse (2.75cm and 1.75cm);

  \node[fixed size node] (thetak) at (0,0) {$\vth_k$};
  \node[fixed size node] (thetastar) at (2.2,-0.5) {$\vth_k^*$};
  
  \coordinate (mom) at (2.4, 1.2);
   \draw[darkorange, ->, dotted] (thetak) -- (mom) node[pos=0.8, below, xshift=2mm, yshift=-1mm] {\(\vm_{k-1}\)};

  \draw[red,->, shorten >= -1mm] (thetak) -- (thetastar) node[pos=0.5, below, xshift=-1mm] {\(\vgo_k\)};

  \end{scope}
  
   \begin{scope}[shift={(12.6,0)}]  %
  
  \node[] at (-2.55,-1.6) {\bf (iii)};
  
  \fill[blue!20,opacity=0.5] (0,0) ellipse (2.75cm and 1.75cm);

  \node[fixed size node] (thetak) at (0,0) {$\vth_k$};
  \node[fixed size node] (thetakp1) at (3.7,0.0)  {$\vth_{k+1}$};

  \draw[darkgreen,->, shorten >=-1mm] (thetak) -- (thetakp1) node[pos=0.5, below, xshift=1mm]{\(\sigma (\vg^O_k + \mu \vm_k)\)};
  
  \node[] () at (2.2,2.1) {\small \(\vm_k = \mu \vm_{k-1} + \vgo_k\)};
  
  \draw[darkorange, ->, dotted] (thetak) -- (3.0, 0.5) node[pos=0.5, above, xshift=0mm, yshift=0mm] {\(\vm_k\)};

  \draw[white] (0,-1.8) -- (0, -2.2);
  
  \end{scope}
   
\end{tikzpicture}
        \caption{{\bf Outer-Nesterov PPO}. 
        At each iteration Outer-Nesterov PPO estimates an outer gradient \(\vgo_k\), updates the momentum \(\vm_k\), and steps the parameters using the Nesterov momentum update. The momentum step therefore \emph{precedes} the construction of the following trust region, since it defines the following behavior policy \(\pi(\vth_{k+1})\).}
        \label{fig:momentum}
    \end{subfigure}
    \begin{subfigure}{\textwidth}
        \centering
          \tikzset{fixed size node/.style={rectangle, minimum width=0.5cm, minimum height=0.5cm, text centered}}
  
\begin{tikzpicture}[line width=1.5pt,>=latex, every node/.style={}, scale=0.7]
 
  \begin{scope}
  [shift={(0,0)}]

  \node[] at (-2.55,-1.6) {\bf (i)};

  \fill[blue!20,opacity=0.5] (0,0) ellipse (2.75cm and 1.75cm);

  \node[fixed size node] (thetak) at (0,0) {$\vth_k$};

  \coordinate (init) at (1.6, 0.8);
 
   \draw[darkorange, ->, shorten <= -2mm] (thetak) -- (init) node[pos=0.3, above, xshift=-3mm, yshift=0mm] {\(\alpha \vm_{k-1}\)};

  \node[fixed size node] (thetak) at (0,0) {$\vth_k$};
  \node[fixed size node] (thetastar) at (2.4,0.4) {$\vth_k^*$};
  
   \draw[blue, ->, dashed, shorten >= -1.5mm] (init) .. controls (0.8,-0.6) and (2,-1.2) .. (thetastar);

  \end{scope}
  
    \begin{scope}
  [shift={(6.3,0)}]
  
  \node[] at (-2.55,-1.6) {\bf (ii)};
  
  \fill[blue!20,opacity=0.5] (0,0) ellipse (2.75cm and 1.75cm);

  \node[fixed size node] (thetak) at (0,0) {$\vth_k$};
  \node[fixed size node] (thetastar) at (2.4,0.4) {$\vth_k^*$};

  \draw[red,->, shorten >= -1mm] (thetak) -- (thetastar) node[pos=0.5, below, xshift=1mm] {\(\vgo_k\)};

  \end{scope}
 
   \begin{scope}[shift={(12.6,0)}]  %
  
  \node[] at (-2.55,-1.6) {\bf (iii)};
  
  \fill[blue!20,opacity=0.5] (0,0) ellipse (2.75cm and 1.75cm);

  \node[fixed size node] (thetak) at (0,0) {$\vth_k$};
  \node[fixed size node] (thetakp1) at (2.7,0.4)  {$\vth_{k+1}$};

  \draw[darkgreen,->, shorten >=-1mm] (thetak) -- (thetakp1) node[pos=0.5, below, xshift=1mm]{\(\vg_k^O\)};
  \node[] () at (1.4,2.1) {\small \(\vm_k = \mu \vm_{k-1} + (1 - \mu)\vgo_k\)};
  
  \draw[darkorange, ->, dotted] (thetak) -- (2.8, 1.0) node[pos=0.5, above, xshift=0mm, yshift=0mm] {\(\vm_k\)};

  \draw[white] (0,-1.8) -- (0, -2.2);
  
  \end{scope}
   
\end{tikzpicture}
        \caption{{\bf Biased initialization}. Each iteration commences with a momentum step (solid orange); the inner optimization (blue dashed) is therefore initialized at \(\vth_k + \alpha \vm_{k-1}\). The momentum step therefore \emph{occurs within} the trust region as the dataset \(\cD_k\) was collected prior, and the surrogate objective remains defined relative to \(\pi(\vth_k)\).}
        \label{fig:bias}
    \end{subfigure}
    \caption{{\bf Comparison of Nesterov-PPO and biased initialization}.} %
\end{figure}

Whilst permitting novel behavior, outer-LR PPO still only exploits information from a single PPO iteration when updating the parameters. Applying momentum breaks this design choice; instead of directly updating the parameters with the scaled outer gradient \(\sigma \vgo_k\), we update using the Nesterov momentum rule as in algorithm \ref{alg:outer-ppo} and illustrated in figure \ref{fig:momentum}.

\begin{algorithm}[H]
    \caption{Outer-Nesterov PPO}
    \label{alg:outer-ppo}
    \begin{algorithmic}[1]
    \State Input: \(\vth_0\) (parameters), \(\sigma\) (learning rate), \(\mu\) (momentum factor)
    \State \(\vm_0 \gets \bm{0} \in \mathbb{R}^d\)
    \For{\(k = 0,1,2, \dots\)}
        \State \(\vgo_k \gets \textproc{PPOIteration}(\vth_{k}) - \vth_k\)
        \State \(\vm_k \gets \mu \vm_{k-1} + \vgo_k\)
        \State \(\vth_{k+1} \gets \vth_{k} + \sigma (\vm_k + \mu \vgo_k)\)
    \EndFor
    \end{algorithmic}
\end{algorithm}

In supervised learning momentum is motivated using pathological curvature, and the ability to `build up speed' \citep{sutskever2013importance}. Given that the outer gradient is the solution to a surrogate objective, we do not anticipate pathological curvature presenting to the outer optimizer. However, similar to learning rates \(\sigma > 1\) the increase in effective learning rate of momentum may assist in learning. Momentum can also be motivated here using resilience to noise; since any given collected dataset will be noisy, the outer gradient is also noisy. As using a learning rate \(\sigma < 1\) corresponded to a lack of trust in any given outer gradient, using momentum corresponds to a smoothing process, where we at no point solely trust a single outer gradient to be accurate.

\subsection{Biased Initialization}

Outer-PPO Nesterov applies a  momentum-based update to the outer loop of PPO. This update occurs \emph{before} the successive iteration's dataset \(\cD_{k+1}\) is collected, hence the momentum directly determines the behavior parameters \(\pi_{k+1}\) for the following surrogate objective. Beyond the effects of stateful inner-loop optimizers such as Adam \citep{kingma2014adam}, each outer gradient estimation is independent of the prior trajectory. In contrast we propose biased initialization to apply an outer momentum-based update \emph{after} data is collected, hence \emph{inside} the following trust region problem as in algorithm \ref{alg:general_pgi_freestep},where \(\vm_k = \mu \vm_{k-1} + (1 - \mu)\vgo_k\) is the momentum vector, and in figure \ref{fig:bias}.

\begin{algorithm}[H]
    \caption{PPO iteration with biased initialization}
    \begin{algorithmic}[1]
    \Function{BiasedPPOIteration}{$\vth,\vm,\alpha$}
    \State Collect set of trajectories \(\cD\) by running policy \(\pi(\vth)\)
    \State Compute advantages \(\hat{A}\).
    \State \(\vth \gets \vth + \alpha \vm\)
    \State \(\vth^* \gets \textproc{InnerOptimizationLoop}(\vth,\cD, \hat{A})\)
    \State \Return \(\vth^*\)
    \EndFunction
    \end{algorithmic}
    \label{alg:general_pgi_freestep}
\end{algorithm}

Biased initialization bears a strong similarity to the conjugate gradient initialization employed in Hessian-free optimization \citep{martens_deep_2010}. The primary motivation for such techniques would be to better estimate the update vector in a given budget of inner-loop iterations.

\section{Experiments}

\subsection{Evaluation Procedure}

We experiment on subsets of the Brax \citep{freeman_brax_2021}, Jumanji \citep{bonnet2024jumanji}, and MinAtar \citep{young_minatar_2019} environment suites, selected as diverse examples of continuous and discrete control problems. We employ the absolute evaluation procedure recommended by \citet{colas2018gep} and \citet{gorsane_towards_2022}. Absolute evaluation entails intermediate evaluations during training and a final, large-scale evaluation using the best policy identified to give the `absolute' performance. We train with a budget of \(1 \times 10^7\) transitions, perform 20 intermediate evaluations, and conduct final evaluation using 1280 episodes.

Recognizing the hyperparameter sensitivity of deep reinforcement learning \citep{hsu_revisiting_2020, engstrom_implementation_2020, andrychowicz_what_2020}, we commit significant resources to establishing a strong PPO baseline and fair evaluation. We sweep for a budget of 600 trials per task using the tree-structured Parzen estimator \citep{bergstra_algorithms_2011, watanabe_tree-structured_2023}. Each trial is the mean of 4 agents, trained using seeds randomly sampled from \([0, 10000]\), for a total of 2400 agents trained per task during baseline tuning. A total of 11 hyperparameters are tuned, each with extensive ranges considered. Full descriptions of the hyperparameter sweep ranges, and the optimal values identified are provided in appendix \ref{app:hyperparams}.

After hyperparameter tuning a final 64 agents are trained per environment task, where the set of evaluation seeds is non-overlapping with seeds used for hyperparameter tuning. To compare methods we aggregate performance over the tasks of an environment suite following the procedure recommended by \citet{agarwal_deep_2021}, normalizing with the min/max return found for each task across all trained agents (including sweep agents), a table of which is presented in appendix \ref{app:additional_results}.

\subsection{Defined Experiments}

We consider the three outer-PPO methods defined in section \ref{sec:outer_ppo}; outer-LR, outer-Nesterov and biased initialization, addressing questions 1, 2, and 3 respectively. The outer-PPO methods are grid searched using increments of 0.1 for all hyperparameters. Outer-LR has a single hyperparameter; outer learning rate \(\sigma\), which is swept over the range [0.1, 4.0] (40 trials). Nesterov-PPO two hyperparameters; \(\sigma\) [0.1, 1.0] and momentum factor \(\mu\) [0.1, 0.9] (90 trials). Biased initialization also has two hyperparameters; bias learning rate \(\alpha\) [0.1, 1.0], bias momentum \(\mu\) [0.0, 0.9] (100 trials). The base PPO hyperparameters are frozen from the baseline sweep up to the 500th trial, such that no method is tuned using a budget greater than the 600 trials used by the baseline. The optimal hyperparameters identified for each sweep are provided in the figures of appendix \ref{app:sweep_performances}.

\begin{figure}[h]
\centering
    \begin{subfigure}{\textwidth}
        \centering
        \includegraphics[width=0.98\textwidth]{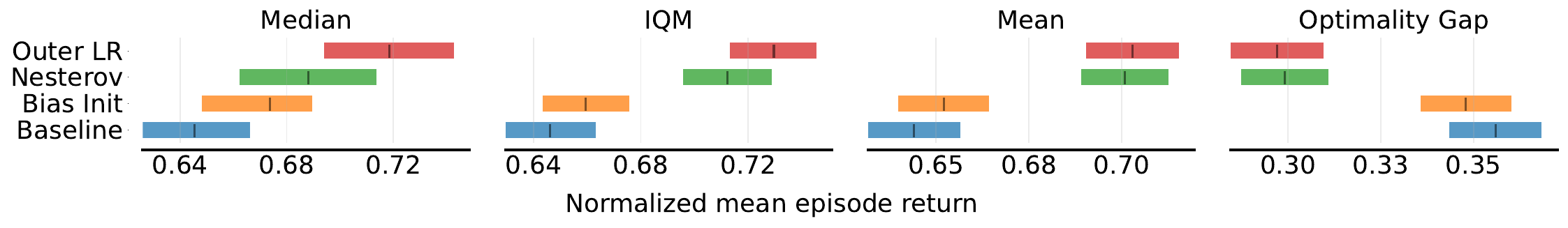}
        \label{fig:brax_aggregate_scores}
    \end{subfigure}
    \begin{subfigure}{\textwidth}
        \centering
        \includegraphics[width=0.98\textwidth]{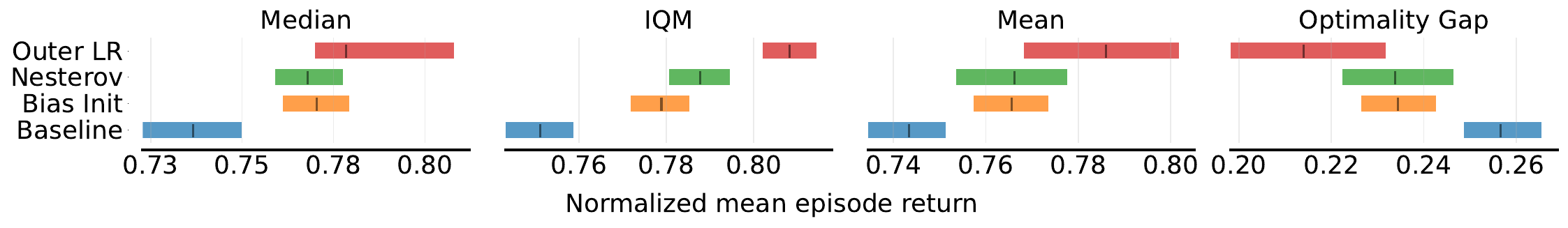}
        \label{fig:minatar_aggregate_scores}
    \end{subfigure}
    \begin{subfigure}{\textwidth}
        \centering
        \includegraphics[width=0.98\textwidth]{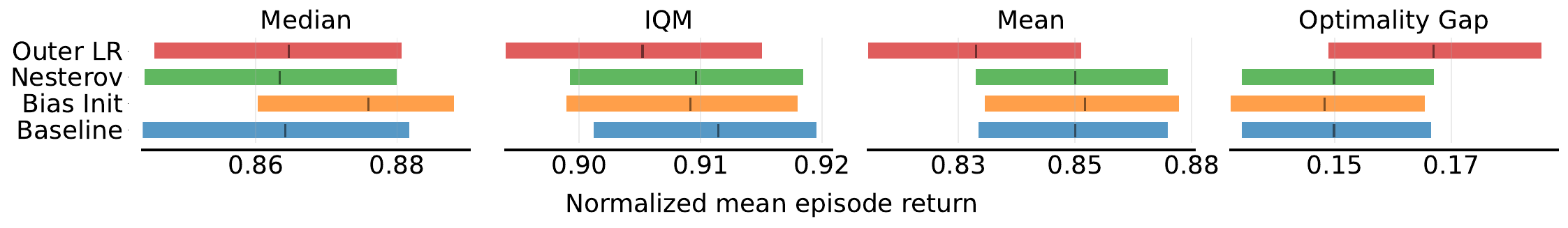}
        \label{fig:minatar_aggregate_scores}
    \end{subfigure}

\caption{{\bf Aggregate point estimates} for Brax (upper), Jumanji (center), and MinAtar (lower). Optimal hyperparameters \emph{per-environment} are used. Normalized to task min/max across all experiments.} 
\label{fig:agg_perf}
\end{figure}

\begin{figure}[h]
\centering
    \begin{subfigure}{0.30\textwidth}
        \includegraphics[width=\textwidth]{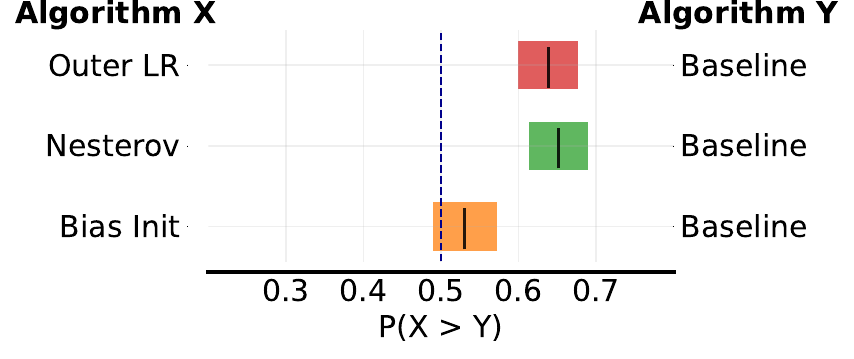}
        \label{fig:brax_prob_improvement}
    \end{subfigure}
    \hfill
    \begin{subfigure}{0.30\textwidth}
        \includegraphics[width=\textwidth]{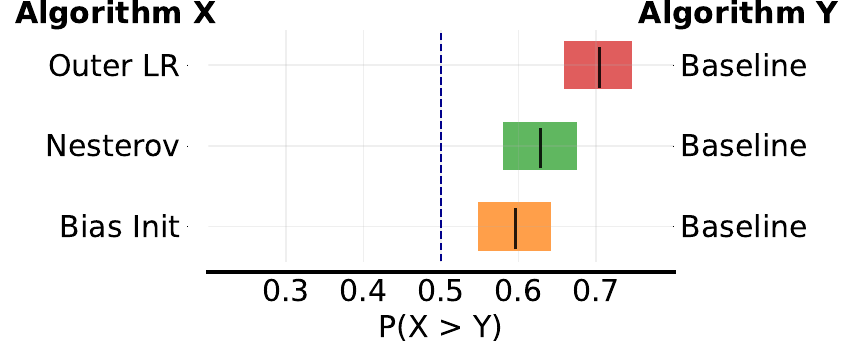}
        \label{fig:jumanji_prob_improvement}
    \end{subfigure}
    \hfill
    \begin{subfigure}{0.30\textwidth}
        \includegraphics[width=\textwidth]{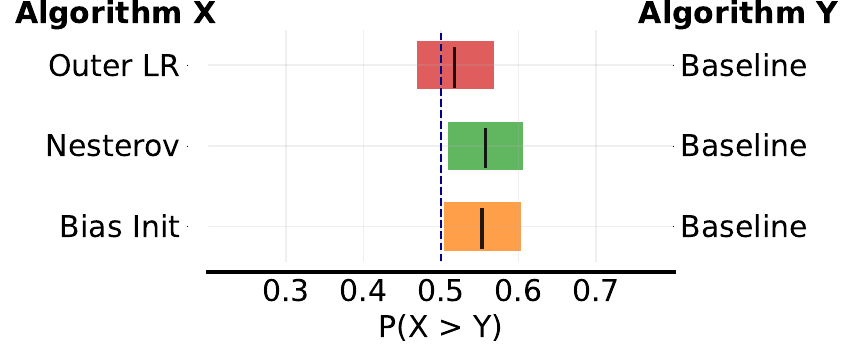}
        \label{fig:minatar_prob_improvement}
    \end{subfigure}

\caption{{\bf Probability of improvement} for Brax (left), Jumanji (center), and MinAtar (right). Optimal hyperparameters \emph{per-environment} are used. Normalized to task min/max across all experiments.}
    \label{fig:prob_improvement}
\end{figure}

\section{Results}

\subsection{Empirical performance}

We first consider the performance of the three outer-PPO methods, where the optimal hyperparameters identified from the grid sweeps \emph{per-environment} are employed. 
In figures \ref{fig:agg_perf} and \ref{fig:prob_improvement} we present the aggregate point estimates and probability of improvement.
Further results including sample efficiency curves are provided in appendix \ref{app:additional_results}.

{\bf Aggregate point estimates}. Outer-LR demonstrates a statistically significant improvement over the PPO baseline on Brax and Jumanji for all point estimates considered (median, IQM, mean, optimality gap). Outer-Nesterov also demonstrates enhanced performance on Brax and Jumanji; this improvement is less substantial than that of outer-LR but remains statistically significant on all point estimates aside from the Brax median. Biased initialization is the weakest of the outer-PPO instantiations, with minor improvements lacking statistical significance on Brax and moderate but significant improvements on Jumanji. No method improves over baseline on MinAtar.

{\bf Probability of improvement}. All methods have a probability of improvement (over baseline) greater than 0.5. In most cases this improvement is statistically significant, aside from biased initialization on Brax and outer-LR on MinAtar. Notably, outer-LR has a probability of improvement greater than 0.6 on Brax and greater than 0.7 on Jumanji. 

\setcounter{topnumber}{3}    %

\begin{figure}[t]
\centering
\begin{subfigure}{0.32\textwidth}
    \includegraphics[width=\textwidth]{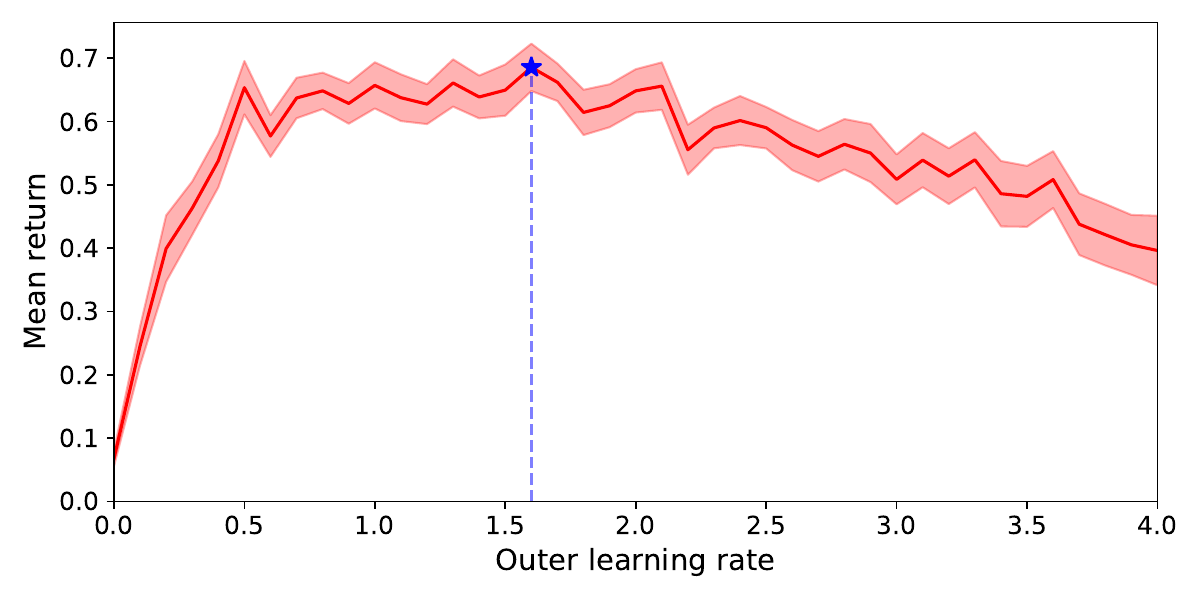}
\end{subfigure}
\hfill
\begin{subfigure}{0.32\textwidth}
    \includegraphics[width=\textwidth]{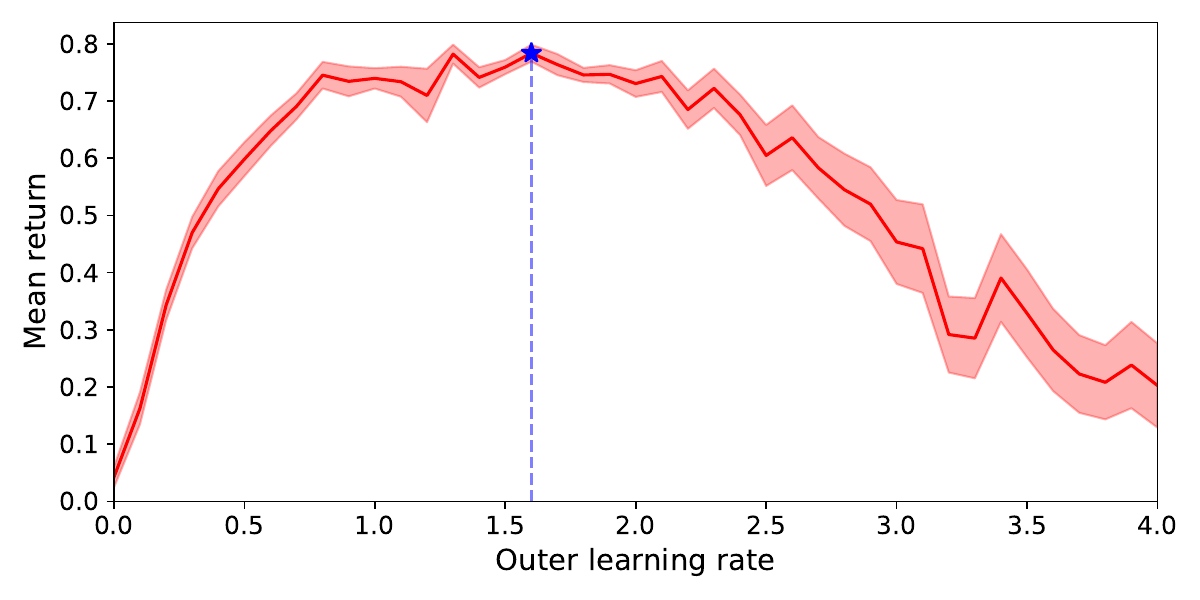}
\end{subfigure}
\hfill
\begin{subfigure}{0.32\textwidth}
    \includegraphics[width=\textwidth]{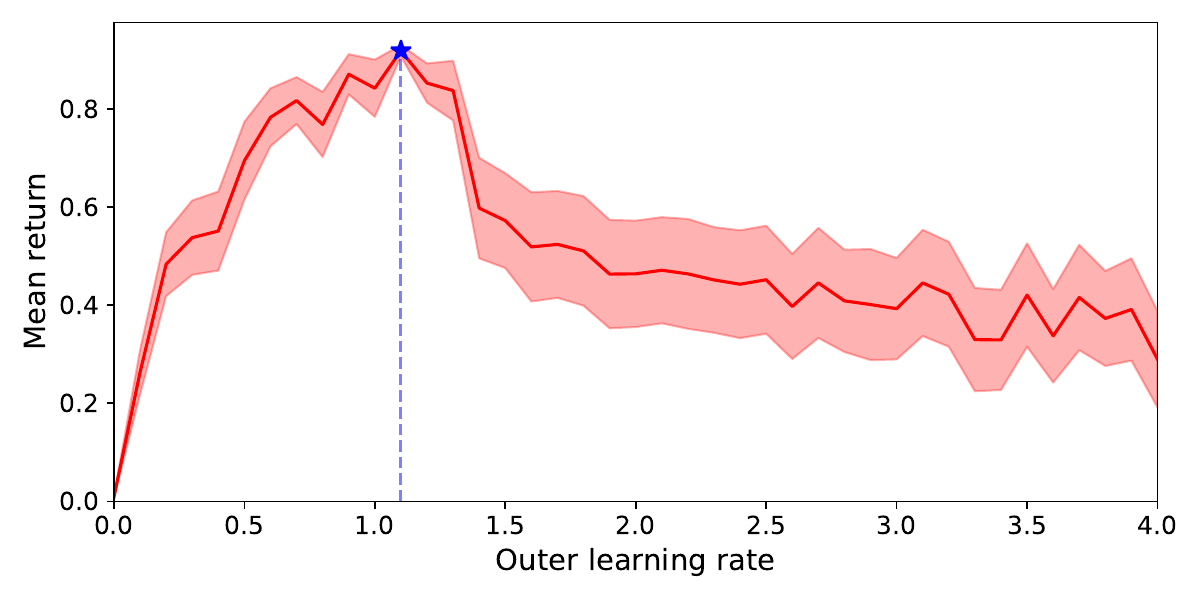}
\end{subfigure}
    \caption{{\bf Outer-LR hyperparameter sensivity.}
    Mean normalized return across the Brax (left), Jumanji (center), MinAtar (right) tasks as a function of outer learning rate \(\sigma\). Mean of 4 seeds plotted with standard error shaded. Normalized to task min/max across all experiments.}
    \label{fig:sensitivity_outer_lr}
\end{figure}
\begin{figure}[t]
\centering
\begin{subfigure}{0.32\textwidth}
    \includegraphics[width=\textwidth]{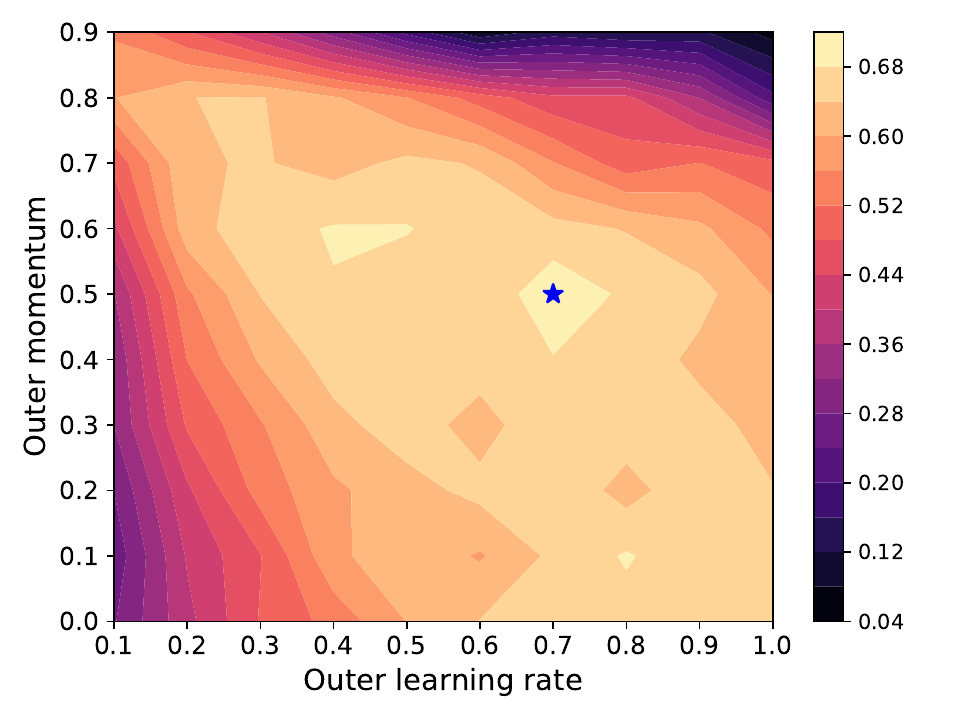}
\end{subfigure}
\hfill
\begin{subfigure}{0.32\textwidth}
    \includegraphics[width=\textwidth]{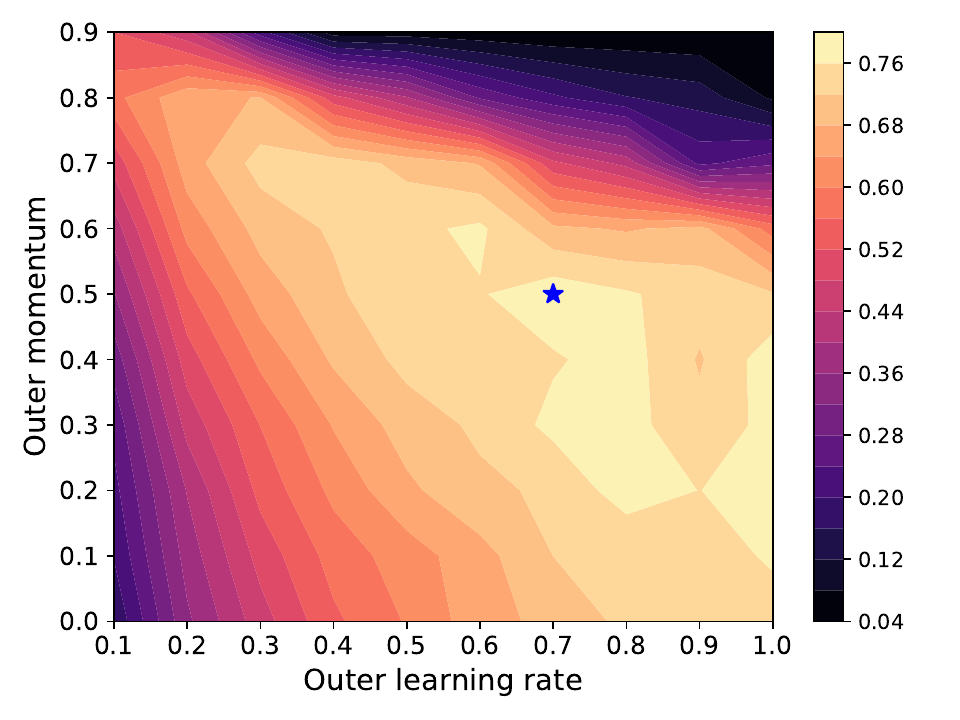}
\end{subfigure}
\hfill
\begin{subfigure}{0.32\textwidth}
    \includegraphics[width=\textwidth]{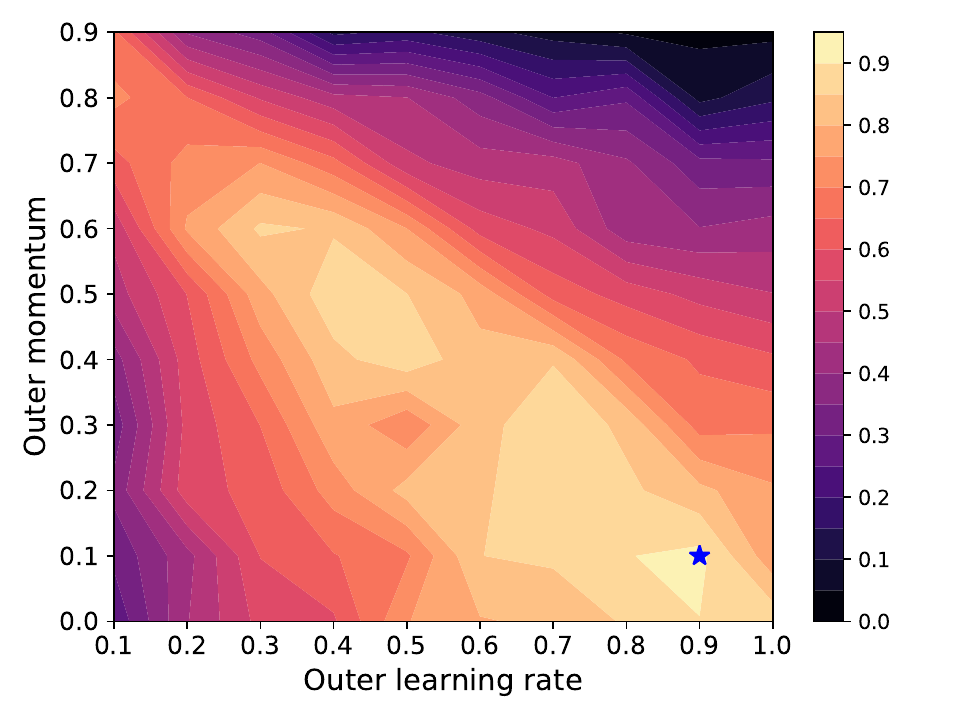}
\end{subfigure}
    \caption{{\bf Outer-Nesterov hyperparameter sensitivity.} Mean normalized return across the Brax (left), Jumanji (center), MinAtar (right) tasks as a function of outer learning rate \(\sigma\) and outer momentum \(\mu\). Mean of 4 seeds plotted. Normalized to task min/max across all experiments.}
    \label{fig:sensitivity_nest}
\end{figure}
\begin{figure}[t]
\centering
\begin{subfigure}{0.32\textwidth}
    \includegraphics[width=\textwidth]{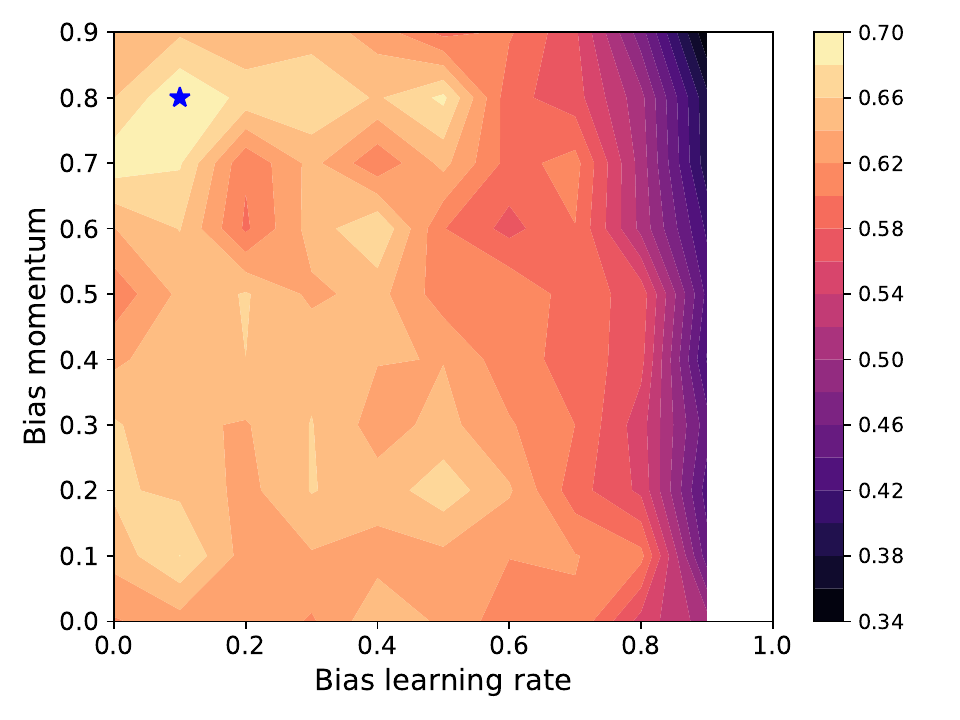}
\end{subfigure}
\hfill
\begin{subfigure}{0.32\textwidth}
    \includegraphics[width=\textwidth]{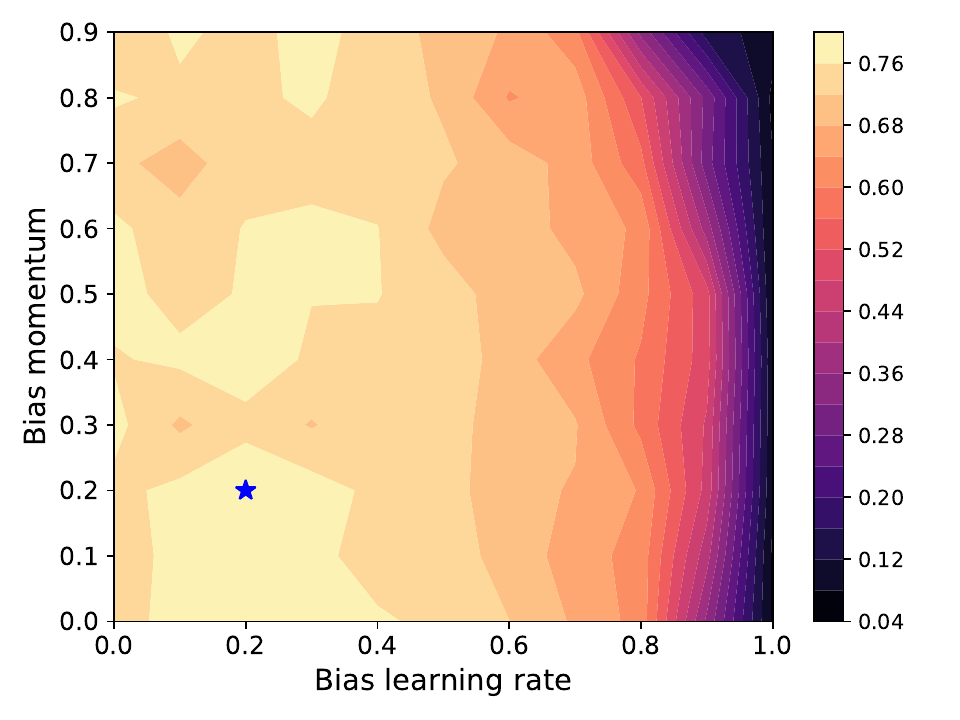}
\end{subfigure}
\hfill
\begin{subfigure}{0.32\textwidth}
    \includegraphics[width=\textwidth]{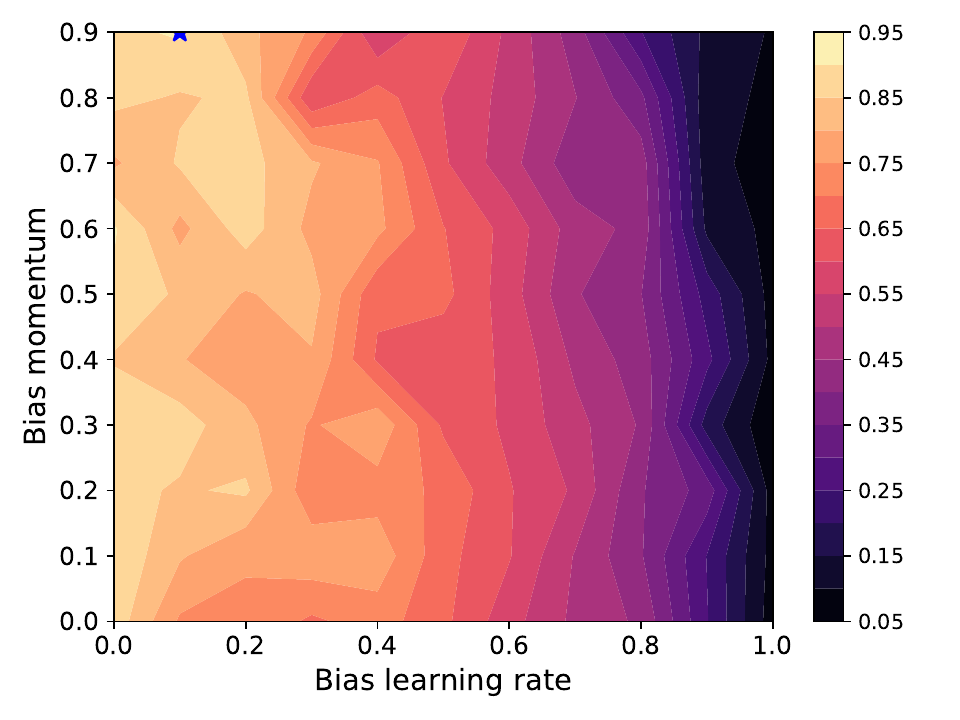}
\end{subfigure}
    \caption{{\bf Biased initialization hyperparameter sensitivity.} Mean normalized return across the Brax (left), Jumanji (center), MinAtar (right) tasks as a function of bias init learning rate \(\alpha\) and bias momentum \(\mu\). Mean of 4 seeds plotted. Normalized to task min/max across all experiments.}
    \label{fig:sensitivity_bias_init}
\end{figure}

\subsection{Hyperparameter Sensitivity}

In the results of figures \ref{fig:agg_perf} and \ref{fig:prob_improvement}, the optimal hyperparameters from each \emph{per-environment} outer-PPO grid search are used. We now consider the sensitivity of outer-PPO to these hyperparameters. In figures \ref{fig:sensitivity_outer_lr}, \ref{fig:sensitivity_nest}, and \ref{fig:sensitivity_bias_init} we present the return,  \emph{normalized across each environment suite}, as a function of the sweep hyperparameters for outer-LR, outer-Nesterov and biased initialization. Normalization is again performed using the extreme values presented in appendix \ref{app:additional_results}. Analogous plots for the individual tasks are provided in appendix \ref{app:sweep_performances}.

\paragraph{Outer learning rate.} When normalized across all tasks, Brax has low sensitivity to outer learning rate. The range of values \(\sigma \in [0.8, 2.0]\) has comparable performance to the peak located at \(\sigma = 1.6\). Notably, performance is not greatly reduced when using values up to \(\sigma = 3.0\). Jumanji again exhibits near optimal-performance over a broad range of values \(\sigma \in [0.5, 2.2]\), with the peak again located at \(\sigma = 1.6\). Unlike in Brax, performance on Jumanji is greatly diminished for values \(\sigma > 2.5\). MinAtar has a sharp peak in performance around standard PPO (\(\sigma = 1.0\)), with a rapid decrease in performance for values greater than this.

\paragraph{Nesterov.} All three suites have a ridge-like trend in normalized performance, with poor performance where \(\sigma\) and \(\mu\) are both small or both large. Both Brax and Jumanji have their peak at \((\sigma, \mu) = (0.7, 0.5)\), with a relatively broad plateau of near-optimal performance. The peak of MinAtar is at \((\sigma, \mu) = (0.9, 0.1)\), with a narrow ridge of near-optimal performance.

\paragraph{Biased initialization.} The dominant trend on all three suites is decreasing normalized performance for large bias learning rate \(\alpha\). The optima for all suites at either \(\alpha = 0.1\) (Brax, MinAtar) or \(\alpha = 0.2\) (Jumajji). There is comparably little variation with respect to bias momentum \(\mu\), with the suite optima dispersed through the available range. Jumanji has a broader region of near optimal performance than Brax or MinAtar, covering \(\alpha < 0.4\).

\section{Discussion}

We now reflect on the questions posed in section \ref{sec:introduction}. The PPO baselines in this work were tuned aggressively for each task, greatly increasing the confidence in the experimental findings. Given the baseline strength, and performance demonstrated in figures \ref{fig:agg_perf} and \ref{fig:prob_improvement}, we conclude in the \emph{negative} for all three questions as evidenced by:
\begin{itemize}
    \item [\bf Q1.] Varying the outer learning rate leads to an statistically significant increase on all point estimates on Brax and Jumanji, with corresponding increases to probability of improvement.
    \item [\bf Q2.] Employing Nesterov momentum on the outer loop, with outer learning rate attenuation, achieves statistically significant increases to all point estimates on Brax and Jumanji. We also observe a statistically significant probability of improvement on all three suites.
    \item [\bf Q3.] Momentum-biased initialization achieves statistically significant increase on all point estimates on Jumanji, with a probability of improvement of 0.6 on this suite.
\end{itemize}

\paragraph{Common hyperparameters.} The sensitivity plots in figures \ref{fig:sensitivity_outer_lr}, \ref{fig:sensitivity_nest} demonstrate robust normalized performance across the Brax and Jumanji suites for outer-LR and outer-Nesterov. However, they do not indicate any significant increase in normalized return could be achieved over standard PPO for a set of common hyperparameters shared across a suite. To achieve the improved aggregate metrics in figure \ref{fig:agg_perf} it was necessary to use task-specific hyperparameters. We do however emphasize the aggressive, task-specific, tuning of the baseline, and view the robustness of normalized return across a range of hyperparameters as a strength of the methods.

\paragraph{Task-specific hyperparameters.} Task-specific hyperparameter sensitivity plots are provided in appendix \ref{app:sweep_performances}. For outer-LR the optimal per-task values for \(\alpha\) range between 0.5 (corresponding to cautious updates) and 2.3 (corresponding to confident updates). That values of \(\alpha\) up to 2.3 can be optimal is surprising, as an \(\alpha\) greater than unity directly violates the trust region established by our previous behavior policy. This precludes the provable monotonic improvement of PPO \citep{kuba_mirror_2022}; by stepping beyond the trust region we may in principle select a policy that is worse than the previous. For outer-Nesterov co-varying \(\sigma\) with \(\mu\) can be understood through the effective learning rate \(\sigma / (1 - \mu)\). The task-specific effective learning rate varies from 0.7 to 2.3. Lastly, for biased initialization the sharp peaks in performance on Brax tasks suggest the method suffers from high variance on this suite, hence the hyperparameters selected may not be optimal in expectation. On Jumanji the method is significantly less hyperparameter sensitive as evidenced by the smooth contours, providing an explanation for the performance gap observed between these suites.

\paragraph{MinAtar results.} No outer-PPO method improved over baseline on MinAtar. We comment that other works committing substantial resources to baseline tuning on MinAtar have struggled to achieve improvements on the suite \citet{jesson_relu_2023}. Furthermore, the hyperparameter sensitivity plots in figures \ref{fig:sensitivity_outer_lr}, \ref{fig:sensitivity_nest} and \ref{fig:sensitivity_bias_init} demonstrate all methods achieve peak normalized return greater than 0.9 on MinAtar. Since here we are normalizing to the maximum performing agents across all sweeps, this indicates there is less variance in the optimal performance of MinAtar compared to Brax and Jumanji with peak normalized returns around 0.7 and 0.8 respectively. A final explanation for the failure to surpass baseline on MinAtar could be `brittle' base hyperparameters, not suited to the modified dynamics introduced by outer-PPO, supported by the sharp peak observed in outer-LR and concentration of performance in outer-Nesterov about standard PPO in figures \ref{fig:sensitivity_outer_lr} and \ref{fig:sensitivity_nest}.

\paragraph{Limitations.} We identify two core limitations to this work; the \emph{fixed transition budget} and the \emph{lack of co-optimization} of base and outer-PPO hyperparameters. We only consider a timestep budget of \(1 \times 10^7\) transitions. Whilst sample efficiency plots are provided in appendix \ref{app:additional_results} the hyperparameters have not been tuned to maximize performance in the data-limited regime. Furthermore, we do not consider the asymptotic performance for larger transition budgets, where it is possible the improvement achieved by outer-PPO methods may be diminished. With respect to co-optimization, given the dependence of the outer gradients on the base hyperparameters there is undoubtedly significant interaction between these and the outer-PPO hyperparameters. Exploring these interactions would yield better understanding and potentially improved performance. We additionally highlight the presence of learning rate annealing on the inner Adam instances in all experiments. This implies the outer gradients tend to zero, the implications of which we do not explore in this work.

\section{Related Work}
\label{sec:related_work}

The usage of the difference between initial parameters and those after gradient-based optimization as a `gradient' has been explored for meta-learning in the Reptile algorithm \citep{nichol2018first}. Reptile aims to find an initialization that can be quickly fine-tuned across a distribution of tasks. Unlike outer-PPO, which applies this idea within a single RL task, Reptile performs gradient steps on different supervised learning tasks to determine the `Reptile gradient'. One could interpret outer-PPO as performing serial Reptile whereby each sampled task is the next PPO iteration alongside the collected dataset.

Whilst to the best of our knowledge we are the first to apply momentum to the outer loop of PPO, momentum-based optimizers such as RMSProp \citet{RMSProp} and Adam \citet{kingma2014adam} are commonly applied in other areas of RL. Recent work has examined the interaction of momentum based optimizers and RL objectives. \citet{bengio2021correcting} identify that a change in objective (such as by updating a target network or dataset), may lead to momentum estimates anti-parallel to the current gradient thereby hindering progress, and propose a correction term to mitigate this effect. \citet{asadi2023resetting} propose to reset the momentum estimates periodically throughout training and demonstrate improved performance on the Atari Learning Environment \citet{bellemare_arcade_2012} with Rainbow \citet{hessel_rainbow_2017} doing so. However, none of these approaches focuses on PPO specifically, and instead address temporal difference learning or value based-methods.  

Lastly, the biased initialization explored in this work is similar to the conjugate gradient initialization technique employed in hessian-free optimization \citet{martens_deep_2010}, although this used only the prior iterate and not a momentum vector. Hessian-free optimization can be considered a supervised learning version of TRPO \citep{schulman_trust_2017}.

\section{Conclusion}
\label{sec:conclusion}

In this work, we introduced outer-PPO, a novel perspective of proximal policy optimization that applies arbitrary gradient-based optimizers to the outer loop of PPO. We posed three key research questions regarding the optimization process in PPO and conducted an empirical investigation across 14 tasks from three environments suites. Our experiments revealed that non-unity learning rates and momentum in the outer loop both yielded statistically significant performance improvements across a variety of evaluation metrics in the Brax and Jumanji environments, with gains ranging from 5-10\% over a heavily tuned PPO baseline. Biased initialization provided improvements upon the baseline on Jumanji tasks but not Brax.

The most immediate direction for future research would be the exploration of interactions between base hyperparameters and outer-PPO hyperparameters. Since the optimal base hyperparameters may be unsuited to the modified dynamics of outer-PPO, the co-optimization of hyperparameters may yield performance improvements and deeper understanding of the method. Other possible future directions include the use of outer-PPO with alternatives to the clipped surrogate loss function, such as KL-penalized PPO \citet{hsu_revisiting_2020} or discovered policy optimization \citet{lu_discovered_2022}, and the use of adaptive optimizers on the outer loop such as RMSProp or Adam. Indeed, an `outer' variant of many dual-loop RL algorithms can be defined, and we hope that this work will stimulate further research into optimizing RL algorithms through more sophisticated outer-loop strategies.

\section*{Acknowledgments}

This work was supported by a Turing AI World-Leading Researcher Fellowship G111021. Research supported with Cloud TPUs from Google's TPU Research Cloud (TRC)

\clearpage

\bibliography{zotero_references, non_zotero_references}
\bibliographystyle{iclr2025_conference}

\appendix

\clearpage

\section{Further Details on PPO}
\label{app:ppo_details}

\subsection{Inner Optimization Loop}

\begin{algorithm}[H]
    \caption{PPO Inner Optimization Loop}
    \begin{algorithmic}[1]
    \State \textbf{Input:} \(\vth\) (initial parameters), \(\cD\) (collected trajectories), \(\hat{A}\) (estimated advantages)
    \State $\vth^{\pi}, \vth^{V} \gets \vth$
    \For{epoch \(i = 1, 2, \dots, N\)}
        \State Shuffle (\(\cD\), \(\hat{A}\)) and create \(M\) minibatches \(\{(\cD_1, \hat{A}_1), (\cD_2, \hat{A}_2), \dots, (\cD_M, \hat{A}_M)\}\)
        \For{\(j = 1, 2, \dots, M\)}
            \State \(\vth^{\pi} \gets \vth^{\pi} + \eta \nabla_{\vth^{\pi}} L^{\pi}(\vth^\pi, \cD_j, \hat{A}_j)\)
            \State \(\vth^{V} \gets \vth^{V} + \eta \nabla_{\vth^{V}} L^{V}(\vth^{V}, \cD_j, \hat{A}_j)\)
        \EndFor
    \EndFor
    \State $\vth \gets \vth^{\pi}, \vth^{V}$
    \State \textbf{Return:} \(\vth^* \gets \vth\)
    \end{algorithmic}
    \label{alg:inner_optim}
\end{algorithm}

Algorithm \ref{alg:inner_optim} describes the inner optimization loop of proximal policy optimization, where \(L^{\pi}\) and \(L^{V}\) are defined in equations \ref{eq:clip_policy} and \ref{eq:clip_value} respectively. For notational ease this presentation is slightly simplified. Typically, instead of the gradient ascent steps taken in lines 5 and 6 typically each of \(\vth^\pi\) and \(\vth^V\) are optimized using independent instances of Adam \citep{kingma2014adam}, with potentially distinct learning rates \(\eta^\pi \neq \eta^V\).

\subsection{Clipped Value Objective}

\begin{equation}\label{eq:clip_value}
L^{V}(\vth^V) = \max \left[ \left( V_{\vth_k} - V_\text{targ} \right)^2,  \left( \text{clip}\left(V_{\vth_k}, V_{\vth_{k-1}} - \varepsilon, V_{\vth_{k-1}} + \varepsilon \right) - V_\text{targ} \right)^2 \right]
\end{equation}

\section{Implementation details}

We implement our experiments using the JAX-based Stoix library \citep{toledo2024stoix}. Our implementation is such that several seeds can be trialed / evaluated simultaneously for the same hyperparameters using a single device. We used Google Cloud TPU (v4-8) for these experiments. The runtime varied between environments, and given different hyperparameters (e.g parallel environments, rollout length, batch size, number of inner epochs) with an average of approximately 10 minutes per 4-seed trial.

\clearpage
\section{Hyperparameters}
\label{app:hyperparams}

\subsection{Sweep Ranges}

The sweep ranges for baseline hyperparameter sweeps are presented in table \ref{tab:sweep_ranges}.

\begin{table}[H]
\centering
\caption{Sweep ranges for baseline hyperparameters.}
\label{tab:sweep_ranges}
\begin{tabular}{ll}
    \toprule
    \textbf{Parameter} & \textbf{Sweep Range} \\
    \midrule
    Parallel environments & \(2^6\) to \(2^{10}\) \\
    Rollout & \(2^2\) to \(2^8\) \\
    Num. epoch & 1 to 16 \\
    Num. minibatch & \(2^0\) to \(2^6\) \\
    Actor learning rate & \(1 \times 10^{-5}\) to \(1 \times 10^{-3}\) (log scale) \\
    Critic learning rate & \(1 \times 10^{-5}\) to \(1 \times 10^{-3}\) (log scale) \\
    Discount factor (\(\gamma\)) & 0.9 to 1.0 \\
    GAE $\lambda$ & 0.0 to 1.0 \\
    Clip $\epsilon$ & 0.1 to 0.5 \\
    Max gradient norm & 0.1 to 5.0 \\
    Reward scaling & \(0.1\) to \(100\) (log scale) \\
    \bottomrule
\end{tabular}
\end{table}

\subsection{Optimal Values}
\label{app:opt_hyperparams}

The optimal values identified by the baseline sweep, up to trial 500, are included in table \ref{tab:sweep_results}. These values are the `base' hyperparameters used for outer-PPO methods. 

\begin{table}[H]
\centering
\small
\caption{Optimal values from baseline sweep up to trial 500}
\label{tab:sweep_results}
\begin{tabular}{lccccccccccc}
\toprule
\rotatebox{0}{\textbf{Task}} &
\rotatebox{90}{\textbf{Parallel env.}} &
\rotatebox{90}{\textbf{Rollout}} &
\rotatebox{90}{\textbf{Num. epoch}} &
\rotatebox{90}{\textbf{Num. m-batch}} &
\rotatebox{90}{\textbf{Actor lr}} &
\rotatebox{90}{\textbf{Critic lr}} &
\rotatebox{90}{\textbf{Discount $\gamma$}} &
\rotatebox{90}{\textbf{GAE $\lambda$}} &
\rotatebox{90}{\textbf{Clip $\epsilon$}} &
\rotatebox{90}{\textbf{Max g. norm}} &
\rotatebox{90}{\textbf{Reward scale}} \\
\midrule
            ant &             128 &               8 &       2 &               32 &  3.0e-04 &   1.4e-04 &  0.98 &       0.70 &     0.21 &          4.85 &           0.14 \\
    halfcheetah &              64 &              64 &       3 &               16 &  3.9e-04 &   4.4e-04 &  0.99 &       0.94 &     0.13 &          2.40 &           0.46 \\
         hopper &              64 &              64 &       2 &               64 &  6.3e-04 &   3.6e-04 &  1.00 &       0.96 &     0.17 &          3.54 &           3.95 \\
       humanoid &             256 &              64 &       4 &               64 &  1.0e-04 &   1.0e-04 &  0.98 &       0.89 &     0.34 &          3.30 &           0.14 \\
humanoidstandup &              64 &              64 &       3 &               32 &  3.0e-04 &   8.2e-04 &  0.99 &       0.98 &     0.10 &          4.65 &           0.35 \\
       walker2d &             256 &              32 &       4 &               64 &  5.4e-04 &   8.2e-04 &  1.00 &       0.92 &     0.12 &          3.74 &          22.54 \\
       \midrule
        asterix &             128 &             128 &       3 &               64 &  8.3e-04 &   2.1e-05 &  1.00 &       0.20 &     0.30 &          2.28 &           6.62 \\
       breakout &              64 &              16 &      14 &               16 &  1.8e-04 &   1.2e-04 &  0.90 &       0.53 &     0.16 &          0.25 &           5.19 \\
        freeway &              64 &             128 &      10 &                2 &  6.9e-04 &   1.3e-04 &  0.98 &       0.70 &     0.15 &          4.71 &           6.64 \\
 space\_invaders &             128 &              32 &      16 &                2 &  3.0e-05 &   1.1e-04 &  0.98 &       1.00 &     0.25 &          0.35 &           0.61 \\
 \midrule
      game\_2048 &            1024 &               8 &       9 &               32 &  4.9e-04 &   3.8e-04 &  0.99 &       0.04 &     0.28 &          2.56 &           0.13 \\
           maze &             256 &              32 &       7 &               64 &  6.5e-04 &   4.3e-04 &  0.98 &       0.66 &     0.14 &          2.46 &           1.97 \\
    rubiks\_cube &              64 &             256 &      13 &                4 &  9.0e-04 &   2.2e-04 &  0.99 &       0.55 &     0.14 &          3.45 &          11.03 \\
          snake &            1024 &               8 &      11 &                4 &  6.0e-04 &   6.0e-04 &  1.00 &       0.46 &     0.12 &          2.52 &          20.48 \\
\bottomrule
\end{tabular}
\end{table}

\begin{table}[H]
\centering
\label{tab:combined}
\caption{Optimal hyperparameters per task for each outer-PPO method}
\begin{tabular}{@{}lrrrrr@{}}
\toprule
\textbf{Task} & \textbf{Outer-LR} & \multicolumn{2}{c}{\textbf{Outer-Nesterov}} & \multicolumn{2}{c}{\textbf{Biased Initialization}} \\ \cmidrule(lr){3-4} \cmidrule(lr){5-6}
              &      \textbf{$\sigma$}      & \textbf{$\sigma$} & \textbf{$\mu$}          &  \textbf{$\alpha$}     & \textbf{$\mu$}  \\ \midrule
Ant           & 0.5               & 0.7               & 0.2                     & 0.1                    & 0.8                     \\
HalfCheetah   & 0.5               & 0.4               & 0.5                     & 0.2                    & 0.8                     \\
Hopper        & 1.5               & 0.9               & 0.4                     & 0.5                    & 0.8                     \\
Humanoid      & 1.9               & 0.5               & 0.7                     & 0.1                    & 0.4                     \\
HumanoidStandup & 2.1             & 0.5               & 0.3                     & 0.5                    & 0.8                     \\
Walker2d      & 2.0               & 0.9               & 0.6                     & 0.4                    & 0.0                     \\
\midrule
2048          & 1.3               & 0.8               & 0.4                     & 0.3                    & 0.9                     \\
Snake         & 2.3               & 1.0               & 0.4                     & 0.7                    & 0.5                     \\
Rubik's Cube  & 1.7               & 0.5               & 0.7                     & 0.4                    & 0.3                     \\
Maze          & 0.9               & 0.9               & 0.0                     & 0.1                    & 0.5                     \\
\midrule
Asterix       & 1.1               & 0.6               & 0.5                     & 0.1                    & 0.4                     \\
Breakout      & 1.1               & 0.9               & 0.1                     & 0.0                    & 0.5                     \\
Freeway       & 1.6               & 0.9               & 0.3                     & 0.2                    & 0.5                     \\
Space Invaders & 1.3              & 0.8               & 0.2                     & 0.1                    & 0.9                     \\ \bottomrule
\end{tabular}
\end{table}

\clearpage

\section{Additional Results}

\label{app:additional_results}

\begin{figure}[h]
\centering
\begin{subfigure}{0.32\textwidth}
    \includegraphics[width=\textwidth]{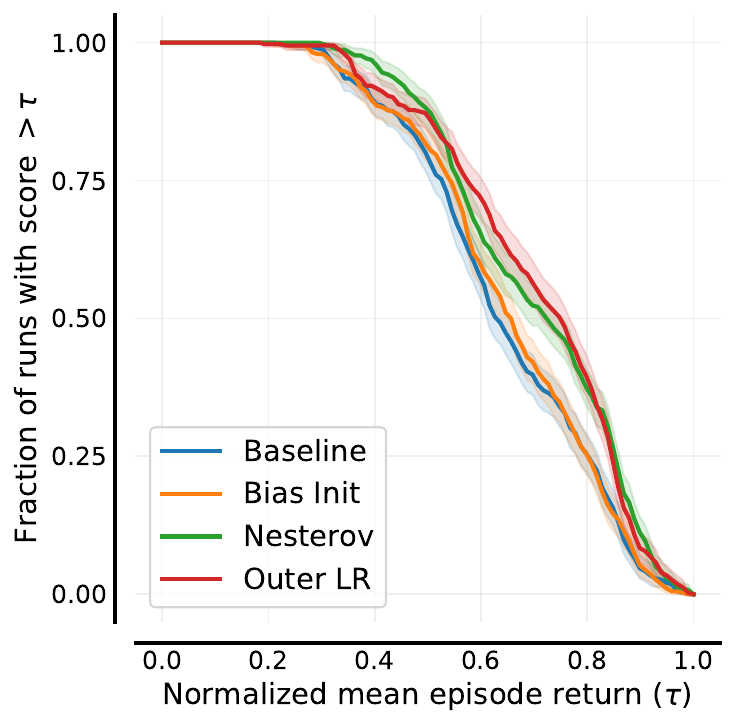}
    \label{fig:brax_aggregate_scores}
\end{subfigure}
\begin{subfigure}{0.32\textwidth}
    \includegraphics[width=\textwidth]{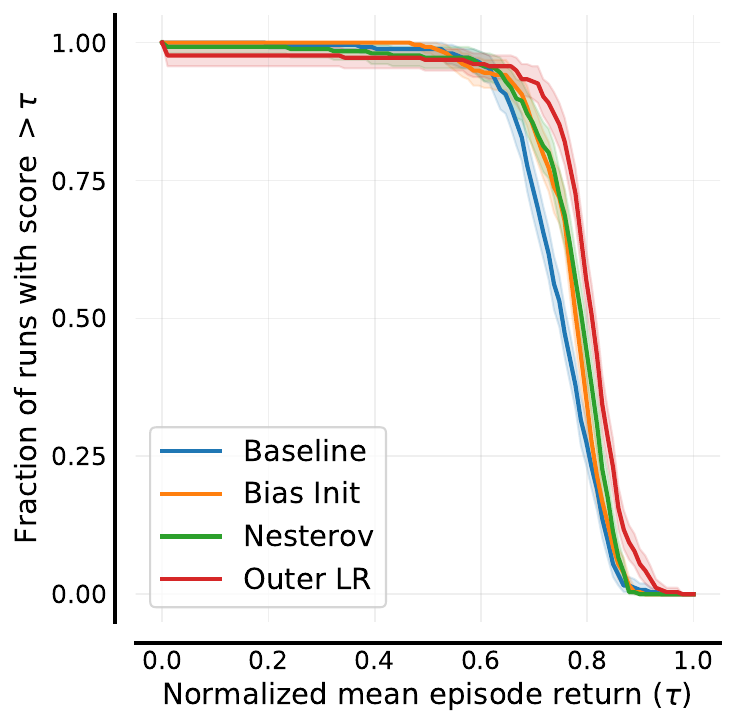}
    \label{fig:jumanji_aggregate_scores}
\end{subfigure}
\begin{subfigure}{0.32\textwidth}
    \includegraphics[width=\textwidth]{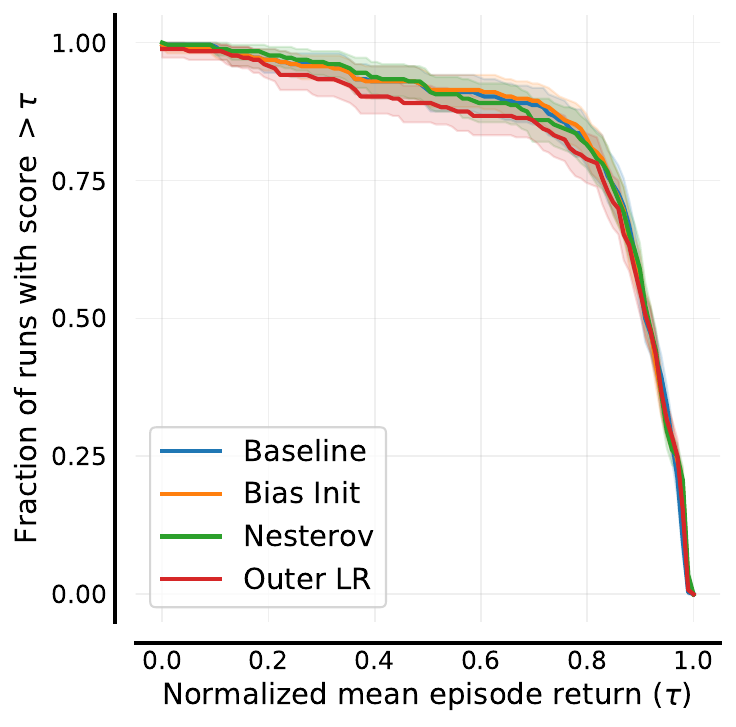}
    \label{fig:minatar_aggregate_scores}
\end{subfigure}
    \caption{\textbf{Performance profiles for Brax (left), Jumanji (center), and MinAtar (right).} 6 / 4 / 4 tasks used from Brax / Jumanji / MinAtar respectively. For each task, agents are trained and evaluated using 64 seeds.}
\end{figure}

\begin{figure}[h]
\centering
    \begin{subfigure}{0.32\textwidth}
        \includegraphics[width=\textwidth]{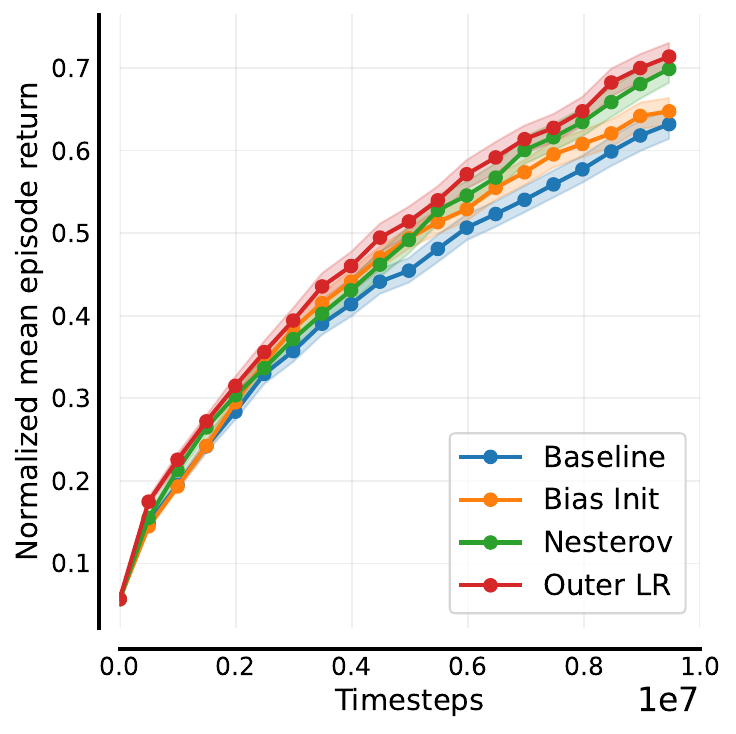}
        \label{fig:brax_sample_efficiency}
        \end{subfigure}
    \begin{subfigure}{0.32\textwidth}
        \includegraphics[width=\textwidth]{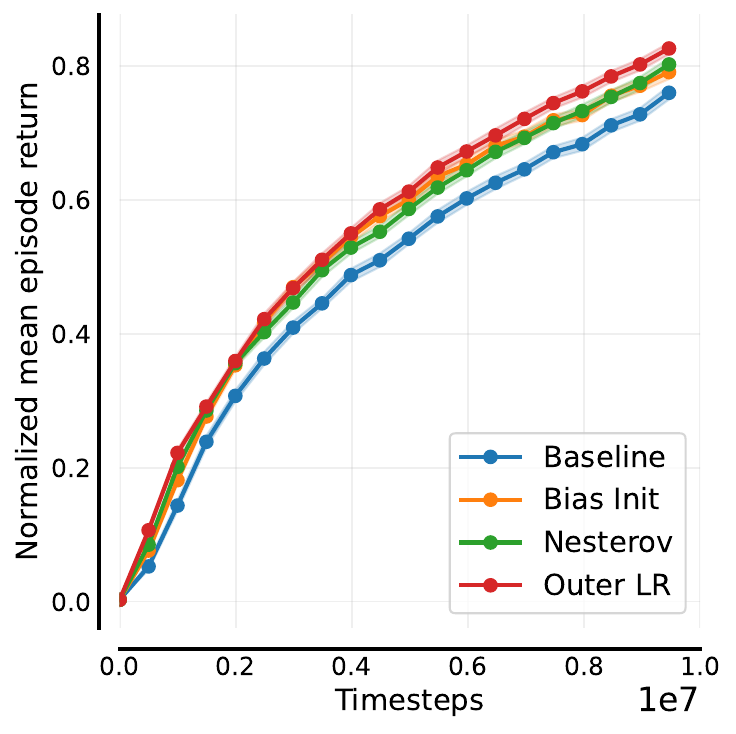}
        \label{fig:jumanji_sample_efficiency}
    \end{subfigure}
    \begin{subfigure}{0.32\textwidth}
        \includegraphics[width=\textwidth]{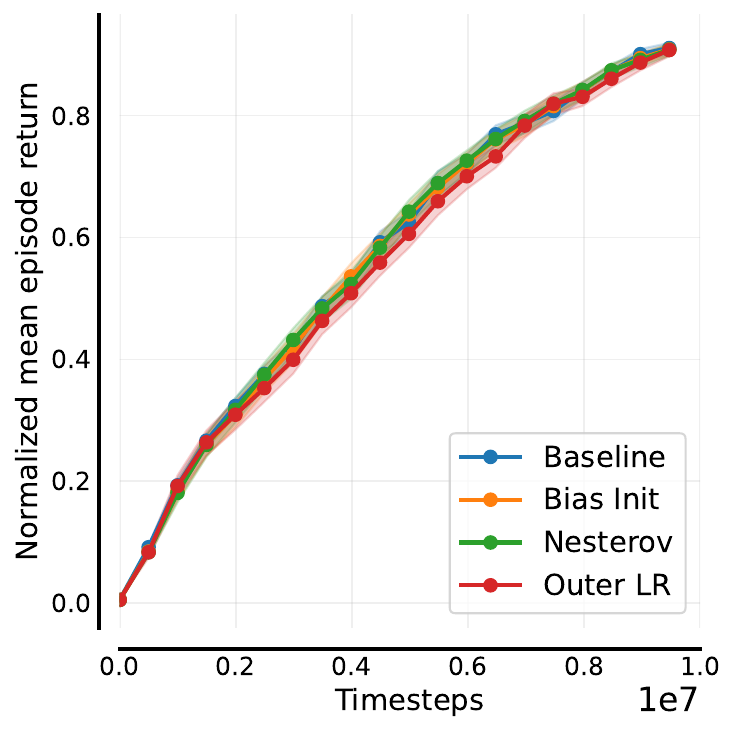}
        \label{fig:minatar_sample_efficiency}
    \end{subfigure}
\caption{Sample efficiency curves for Brax (left), Jumanji (center), and MinAtar (right).}
    \label{fig:sample_efficiency}
\end{figure}

\begin{table}[h]
\centering
\label{tab:min_max_returns}
\caption{Minimum and maximum returns used for normalization.}
\begin{tabular}{@{}lrr@{}}
\toprule
\textbf{Task}   & \textbf{Min} & \textbf{Max} \\ \midrule
Ant             & -2958.14    & 13466.48    \\
Halfcheetah     & -587.37      & 7859.28     \\
Hopper          & 21.03        & 3697.39     \\
Humanoid        & 207.63       & 11851.71    \\
Humanoidstandup & 6686.00     & 71897.67    \\
Walker2d        & -32.44       & 2558.61     \\
\midrule
2048            & 989.50       & 29084.63    \\
Snake           & 0.00         & 92.55        \\
Rubiks Cube     & 0.00         & 0.66         \\
Maze            & 0.03         & 0.84         \\
\midrule
Asterix         & 0.30         & 64.46        \\
Breakout        & 0.00         & 92.86        \\
Freeway         & 0.00         & 66.13        \\
Space Invaders  & 0.00         & 191.80       \\
\bottomrule
\end{tabular}
\end{table}

\section{Sweep Performances}
\label{app:sweep_performances}

\subsection{Baseline}

\begin{figure}[H]
\centering
\begin{subfigure}{0.3\textwidth}
    \includegraphics[width=\textwidth]{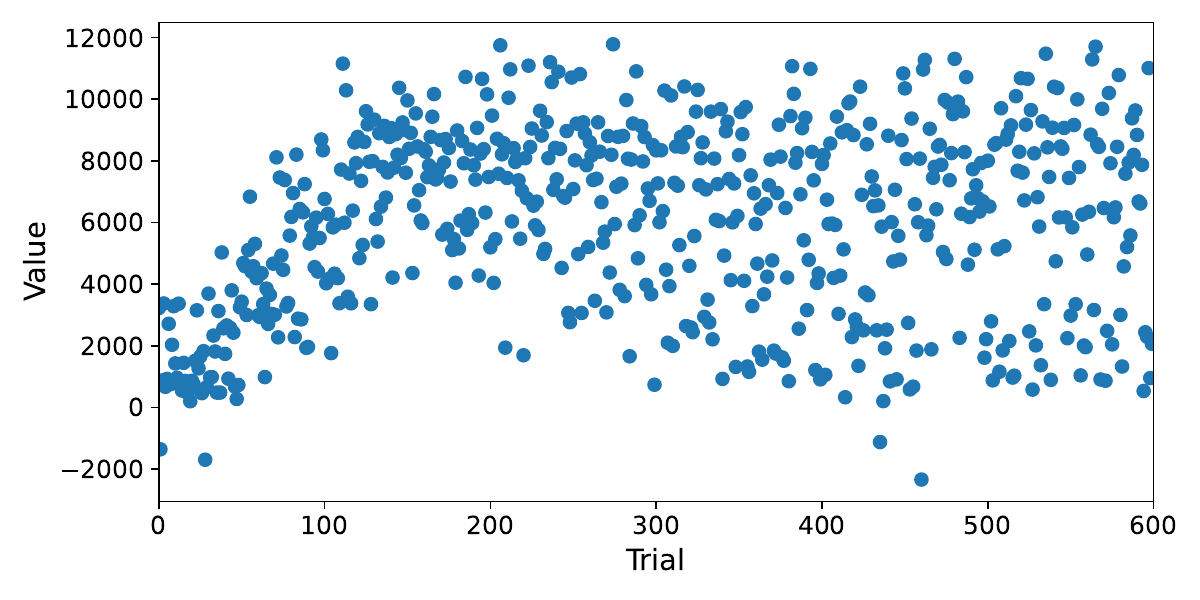}
    \caption{ant}
\end{subfigure}
\hfill
\begin{subfigure}{0.3\textwidth}
    \includegraphics[width=\textwidth]{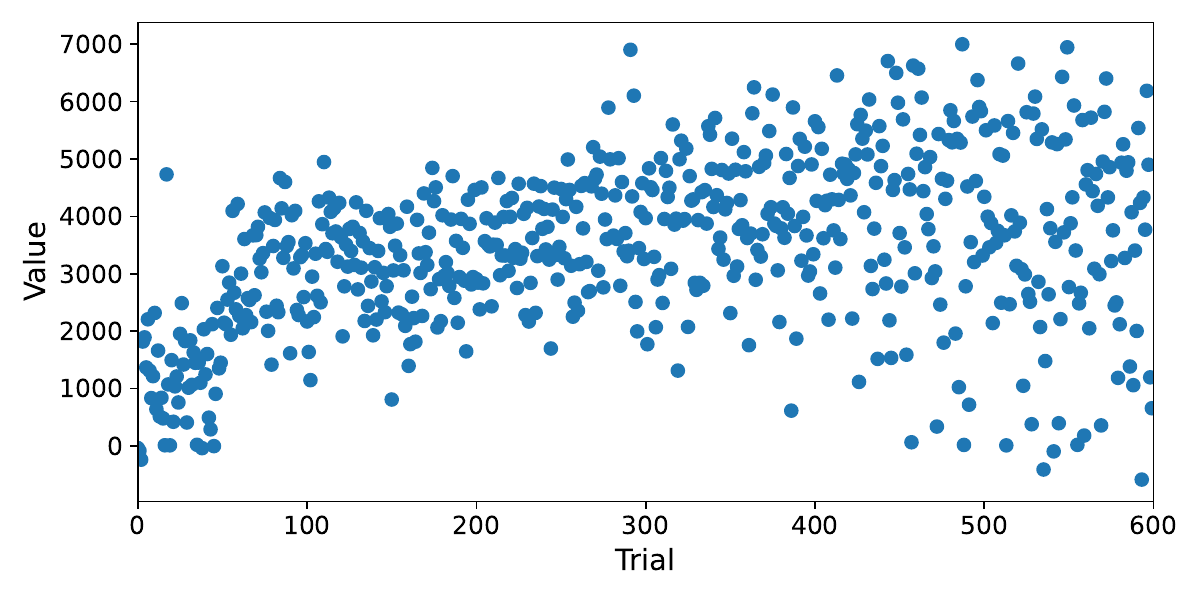}
    \caption{halfcheetah}
\end{subfigure}
\hfill
\begin{subfigure}{0.3\textwidth}
    \includegraphics[width=\textwidth]{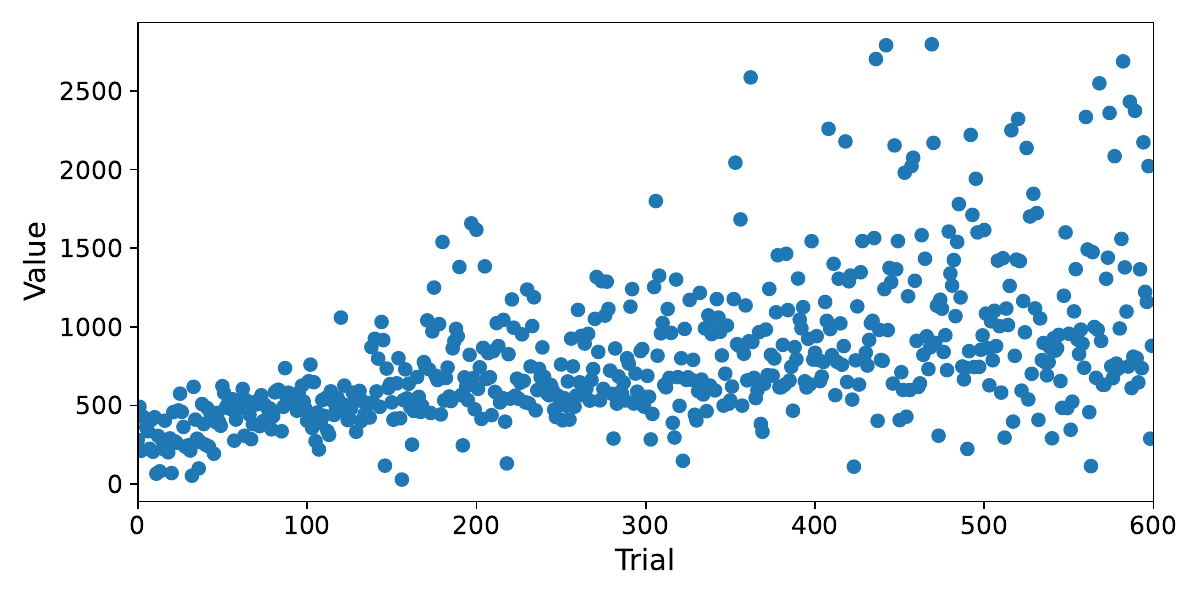}
    \caption{hopper}
\end{subfigure}
\hfill
\begin{subfigure}{0.3\textwidth}
    \includegraphics[width=\textwidth]{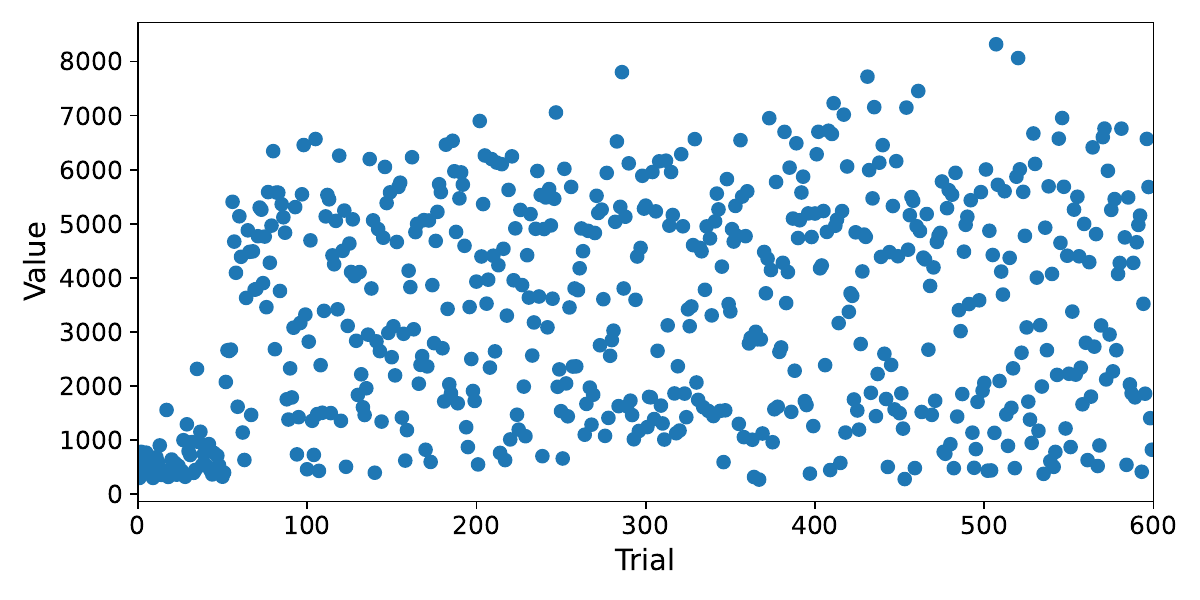}
    \caption{humanoid}
\end{subfigure}
\hfill
\begin{subfigure}{0.3\textwidth}
    \includegraphics[width=\textwidth]{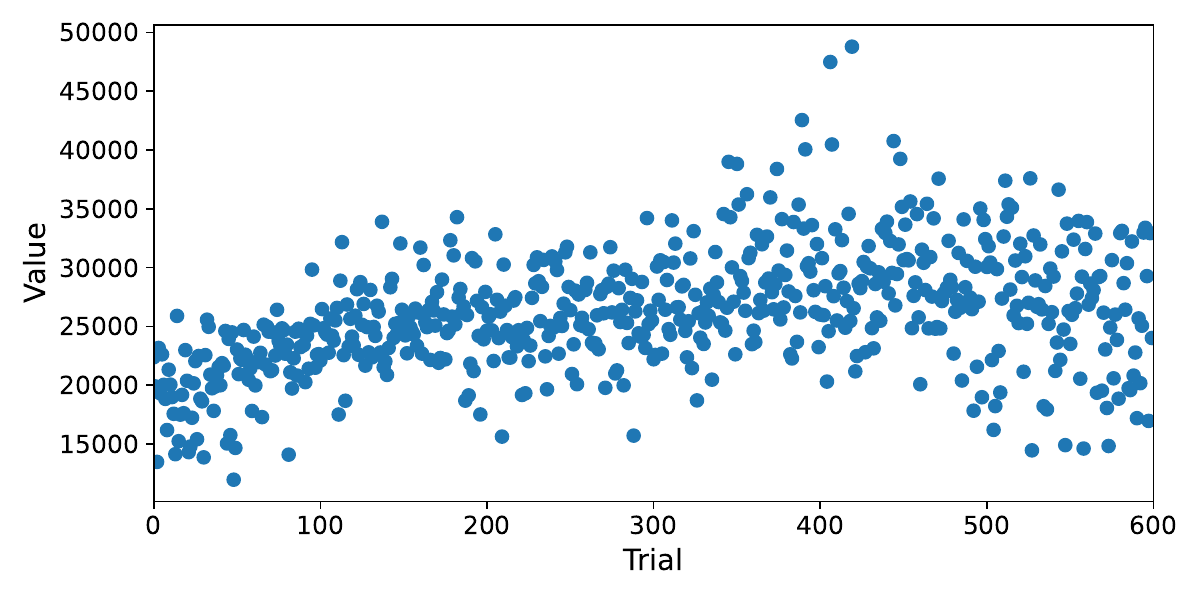}
    \caption{humanoidstandup}
\end{subfigure}
\hfill
\begin{subfigure}{0.3\textwidth}
    \includegraphics[width=\textwidth]{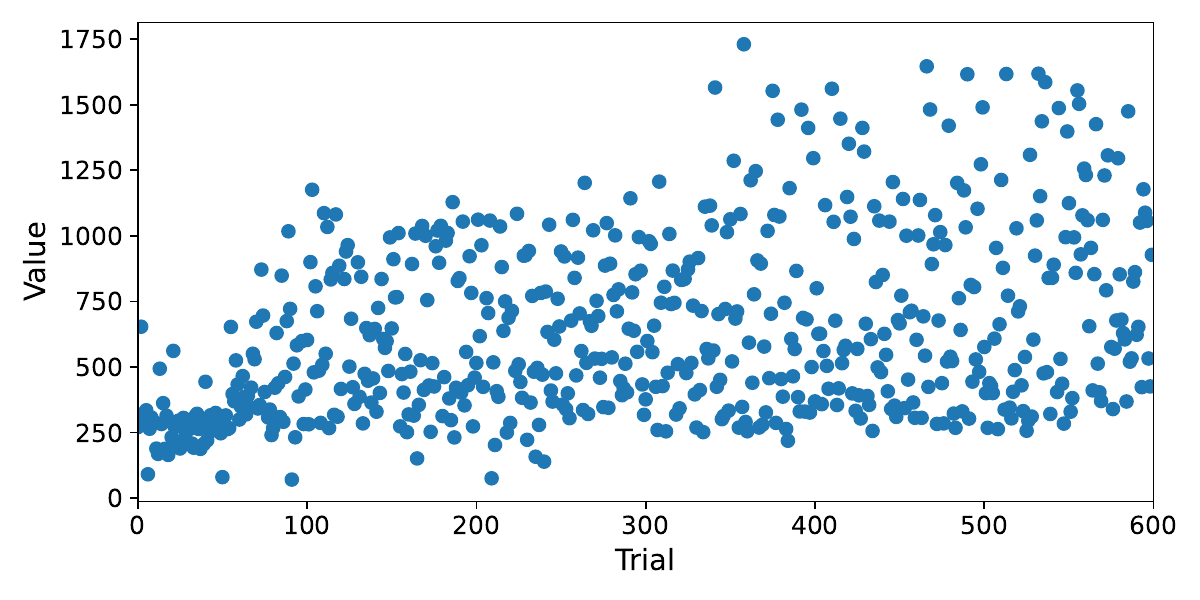}
    \caption{walker2d}
\end{subfigure}
    \caption{\textbf{Baseline sweep performance for Brax tasks.} Each point represents the mean of a 4 seed trial.}
\end{figure}

\begin{figure}[H]
\centering
\begin{subfigure}{0.3\textwidth}
    \includegraphics[width=\textwidth]{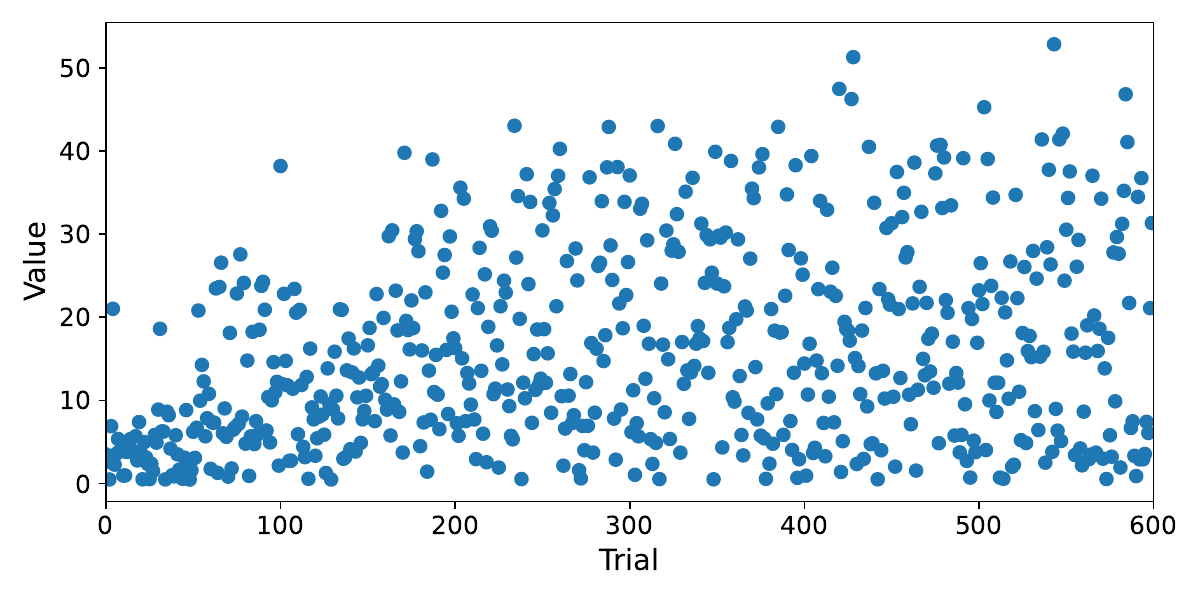}
    \caption{asterix}
\end{subfigure}
\hspace{1.5cm}
\begin{subfigure}{0.3\textwidth}
    \includegraphics[width=\textwidth]{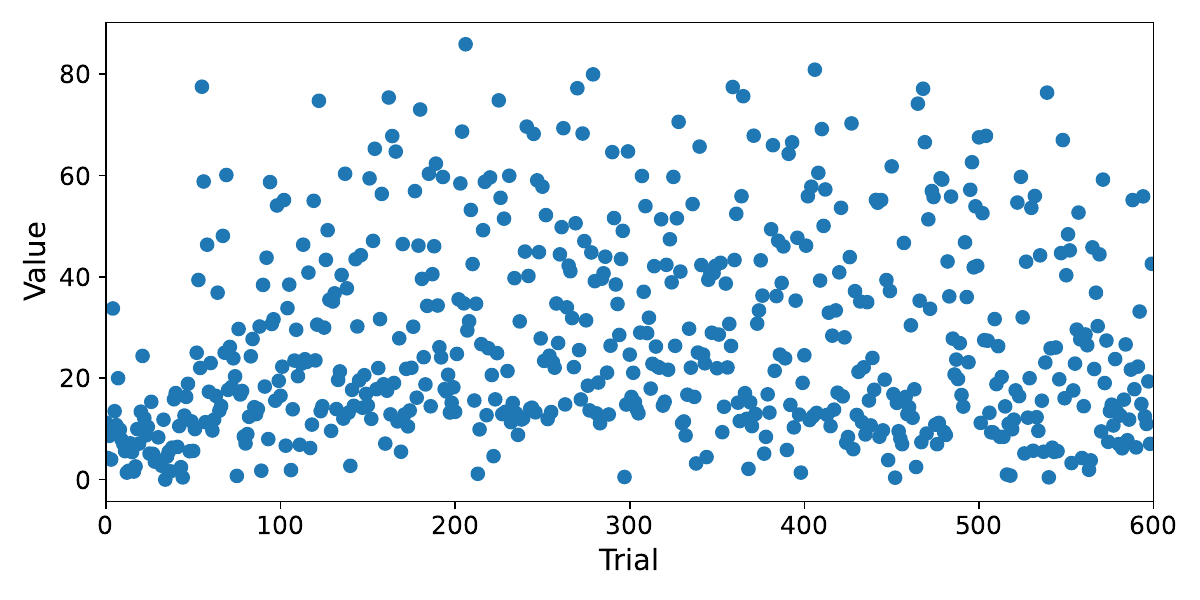}
    \caption{breakout}
\end{subfigure}
\\
\begin{subfigure}{0.3\textwidth}
    \includegraphics[width=\textwidth]{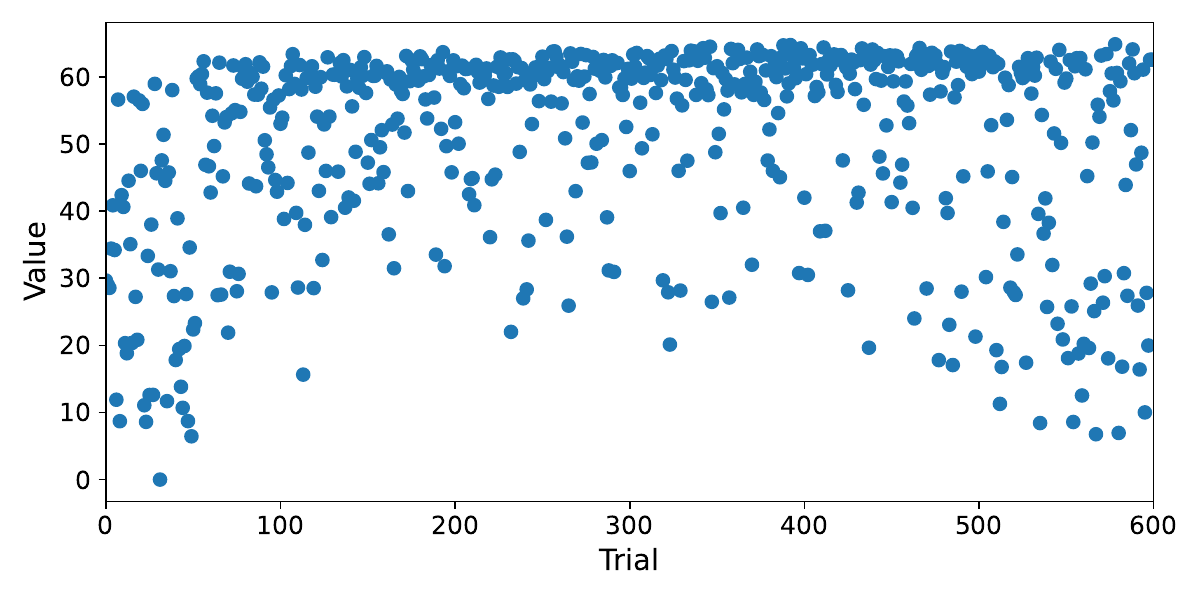}
    \caption{freeway}
\end{subfigure}
\hspace{1.5cm}
\begin{subfigure}{0.3\textwidth}
    \includegraphics[width=\textwidth]{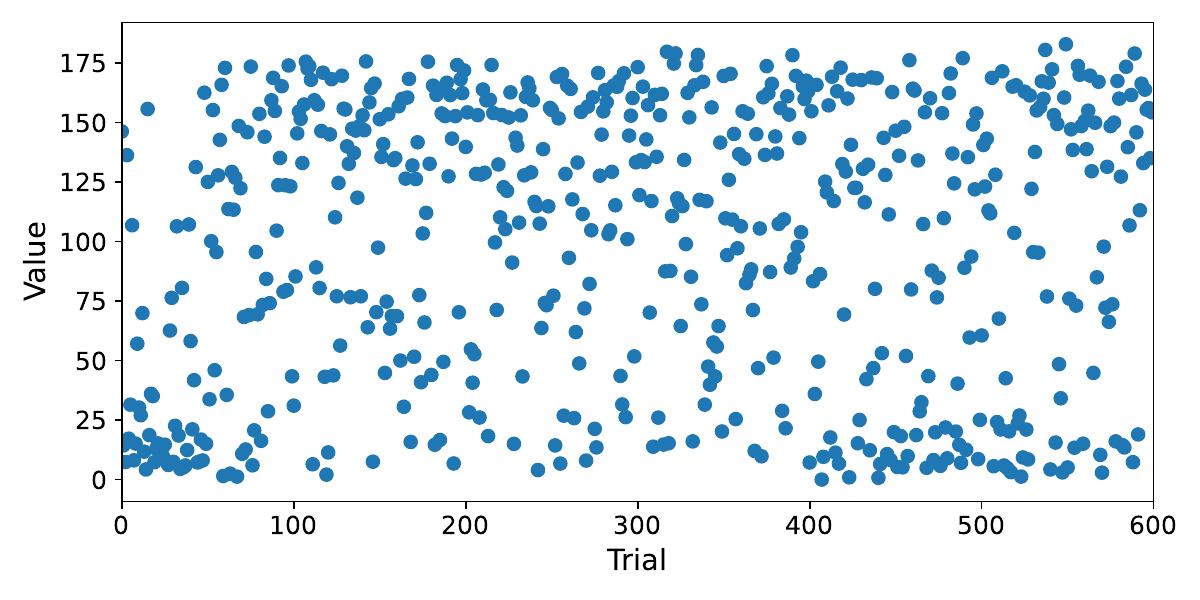}
    \caption{space\_invaders}
\end{subfigure}
    \caption{\textbf{Baseline sweep performance for MinAtar tasks.} Each point represents the mean of a 4 seed trial.}
\end{figure}

\begin{figure}[H]
\centering
\begin{subfigure}{0.3\textwidth}
    \includegraphics[width=\textwidth]{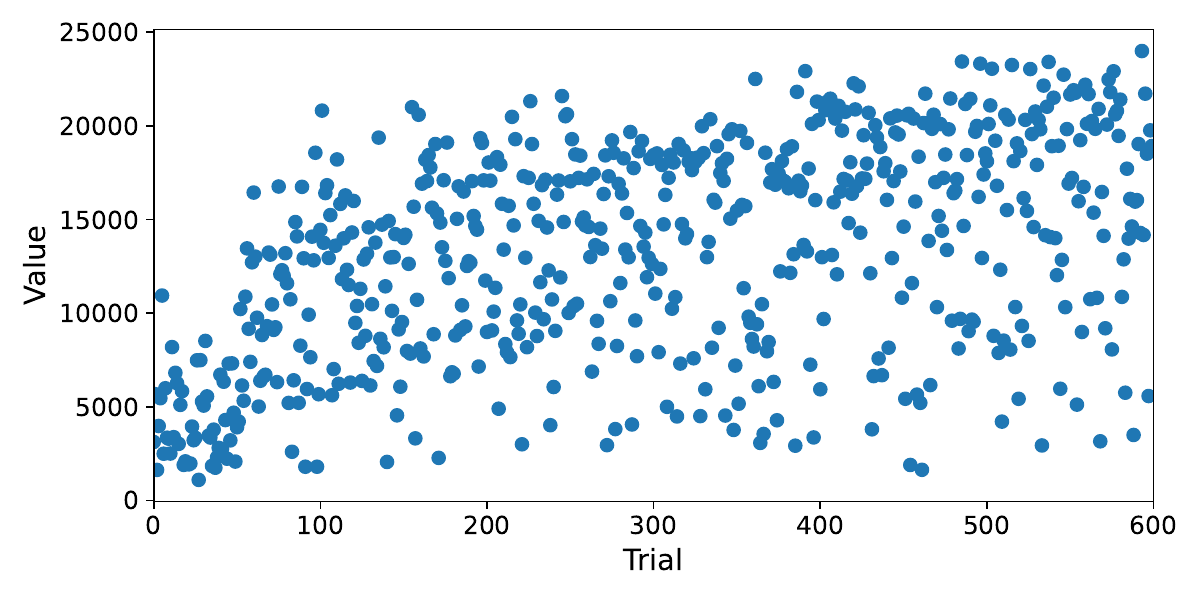}
    \caption{game\_2048}
\end{subfigure}
\hspace{1.5cm}
\begin{subfigure}{0.3\textwidth}
    \includegraphics[width=\textwidth]{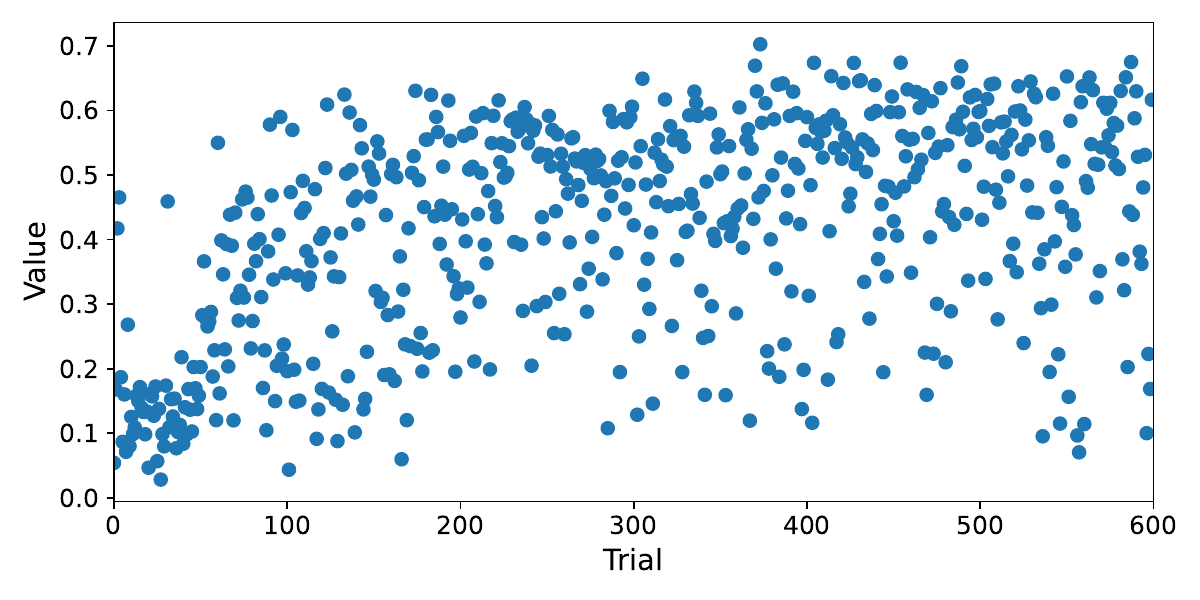}
    \caption{maze}
\end{subfigure}
\\
\begin{subfigure}{0.3\textwidth}
    \includegraphics[width=\textwidth]{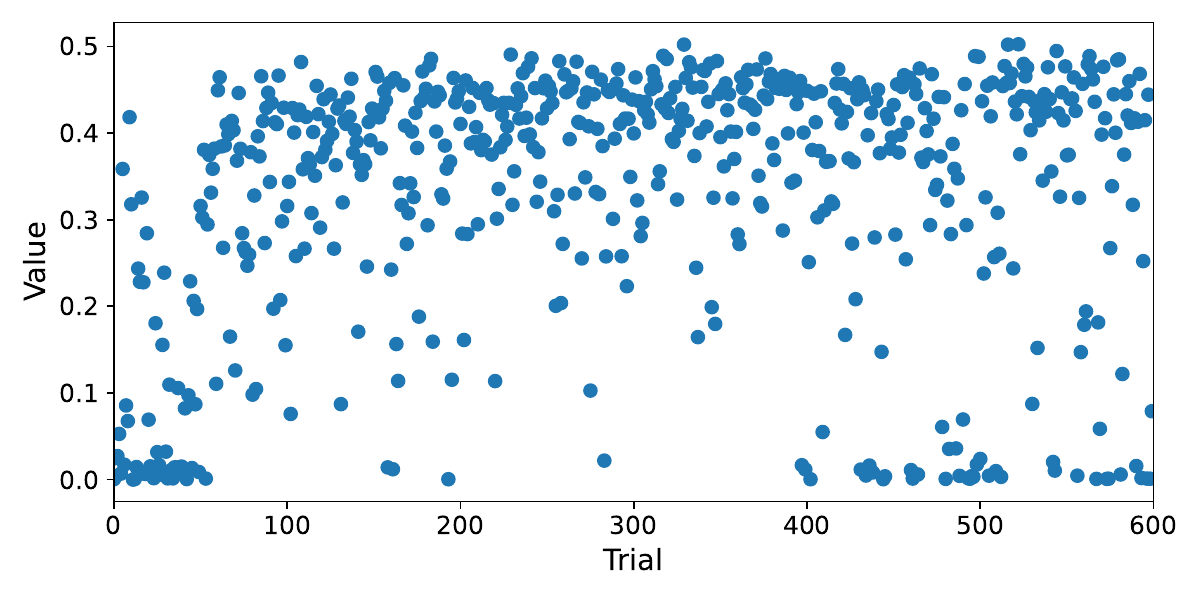}
    \caption{rubiks\_cube}
\end{subfigure}
\hspace{1.5cm}
\begin{subfigure}{0.3\textwidth}
    \includegraphics[width=\textwidth]{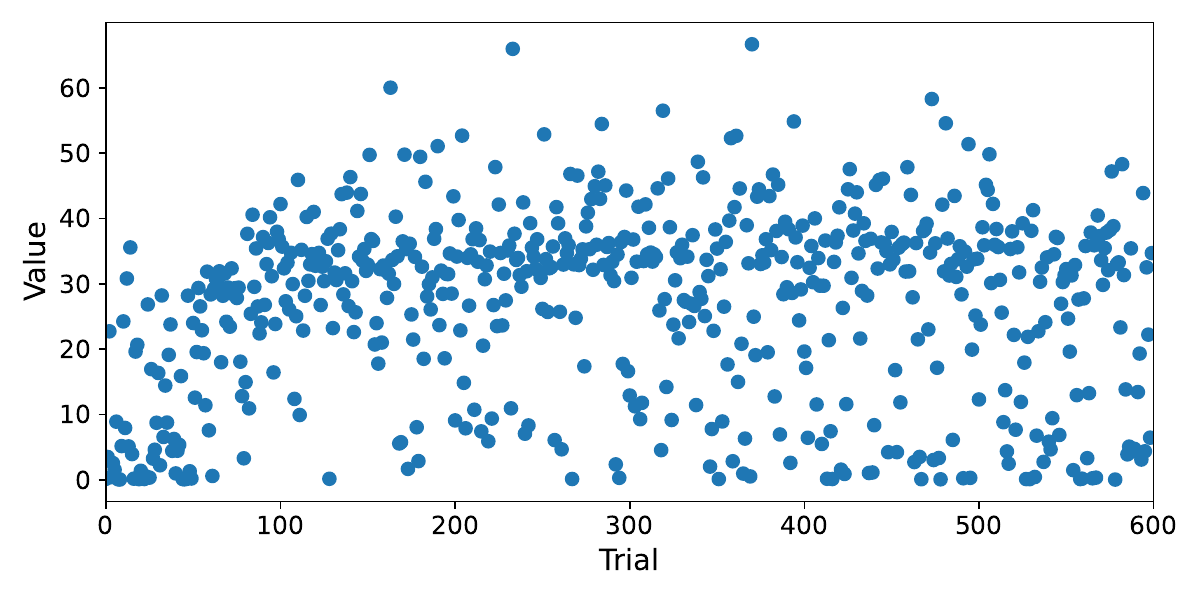}
    \caption{snake}
\end{subfigure}
    \caption{\textbf{Baseline sweep performance for Jumajji tasks.} Each point represents the mean of a 4 seed trial.}
\end{figure}

\clearpage

\subsection{Outer learning rates}

\begin{figure}[H]
\centering
\begin{subfigure}{0.3\textwidth}
    \includegraphics[width=\textwidth]{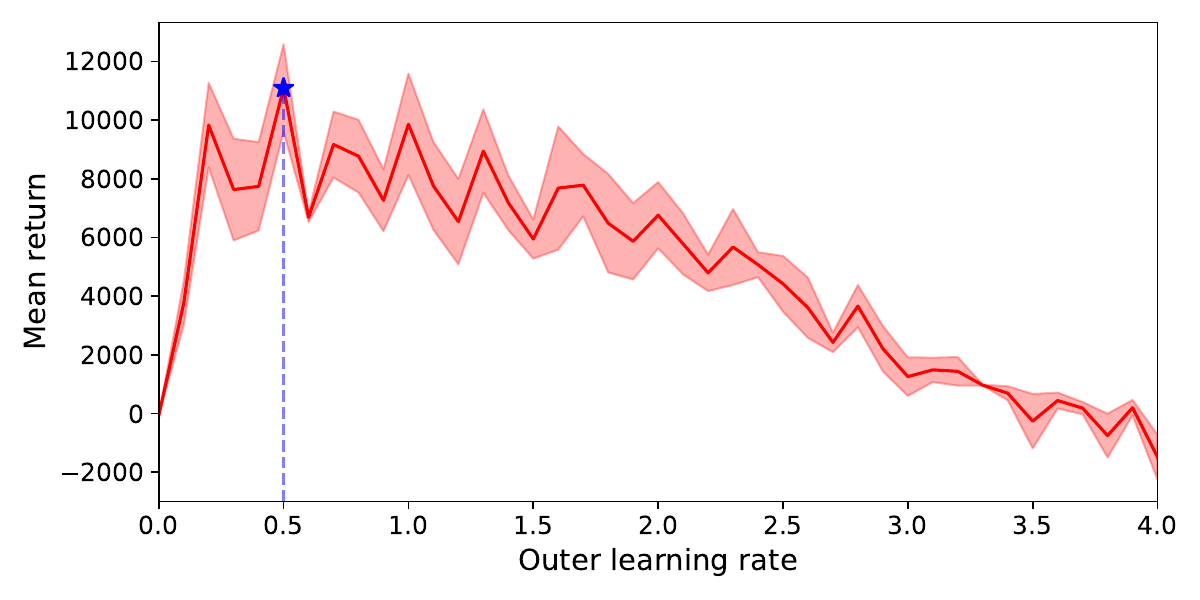}
    \caption{ant}
\end{subfigure}
\hfill
\begin{subfigure}{0.3\textwidth}
    \includegraphics[width=\textwidth]{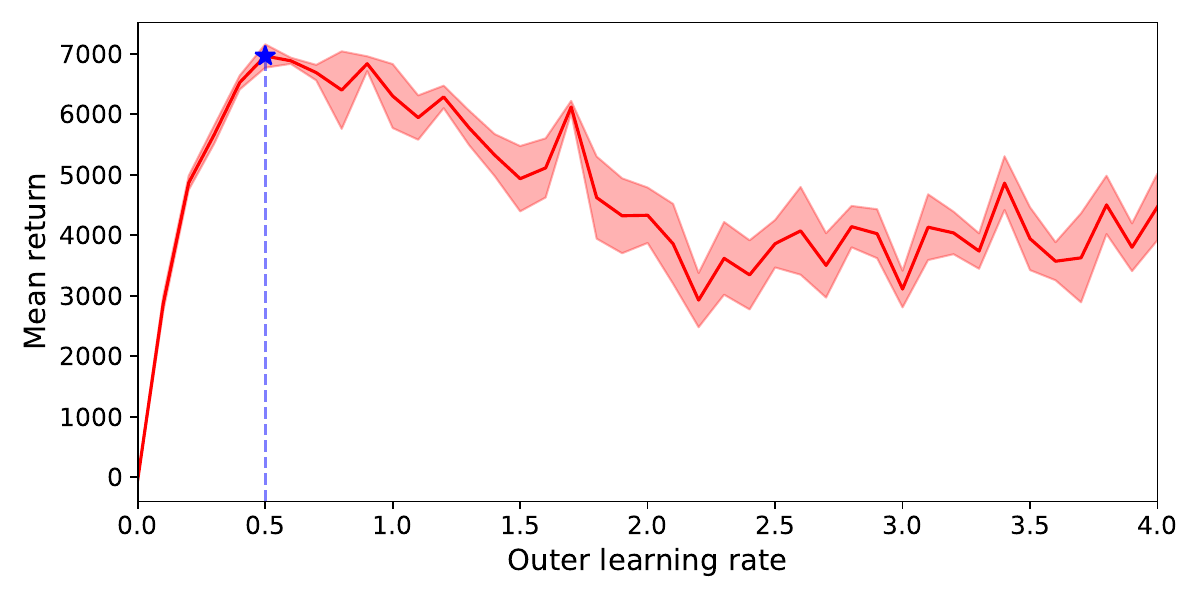}
    \caption{halfcheetah}
\end{subfigure}
\hfill
\begin{subfigure}{0.3\textwidth}
    \includegraphics[width=\textwidth]{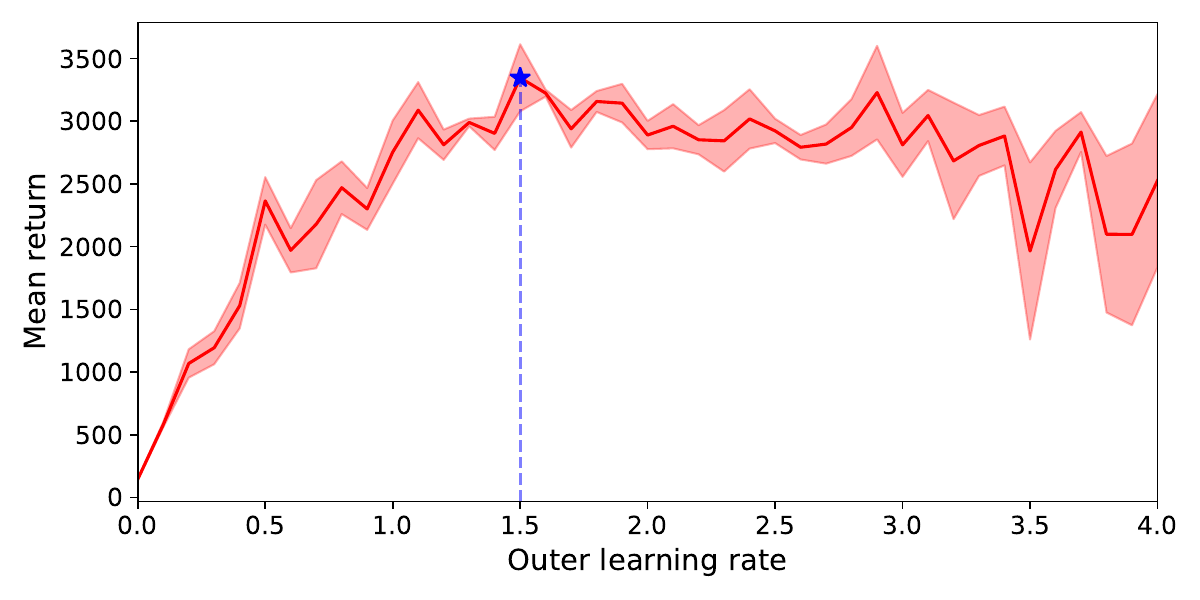}
    \caption{hopper}
\end{subfigure}
\hfill
\begin{subfigure}{0.3\textwidth}
    \includegraphics[width=\textwidth]{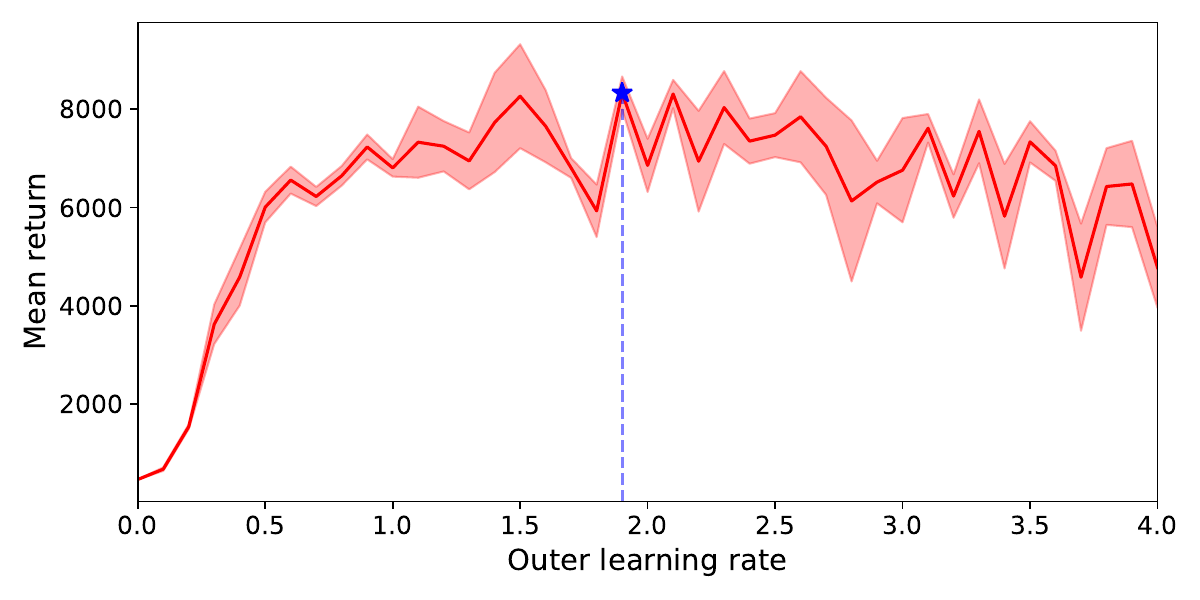}
    \caption{humanoid}
\end{subfigure}
\hfill
\begin{subfigure}{0.3\textwidth}
    \includegraphics[width=\textwidth]{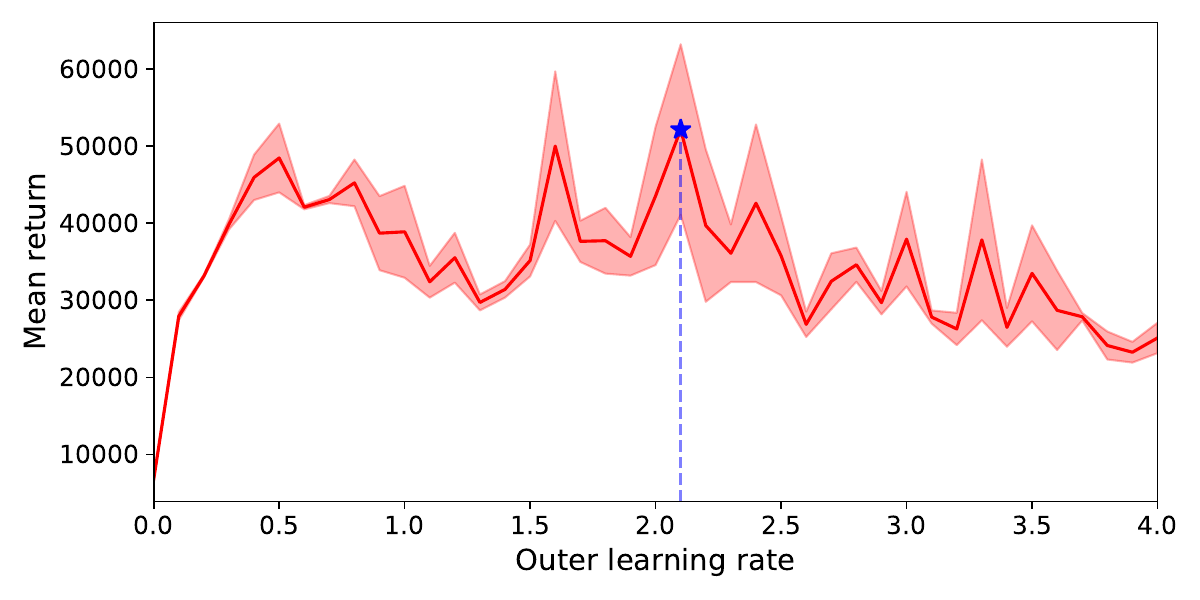}
    \caption{humanoidstandup}
\end{subfigure}
\hfill
\begin{subfigure}{0.3\textwidth}
    \includegraphics[width=\textwidth]{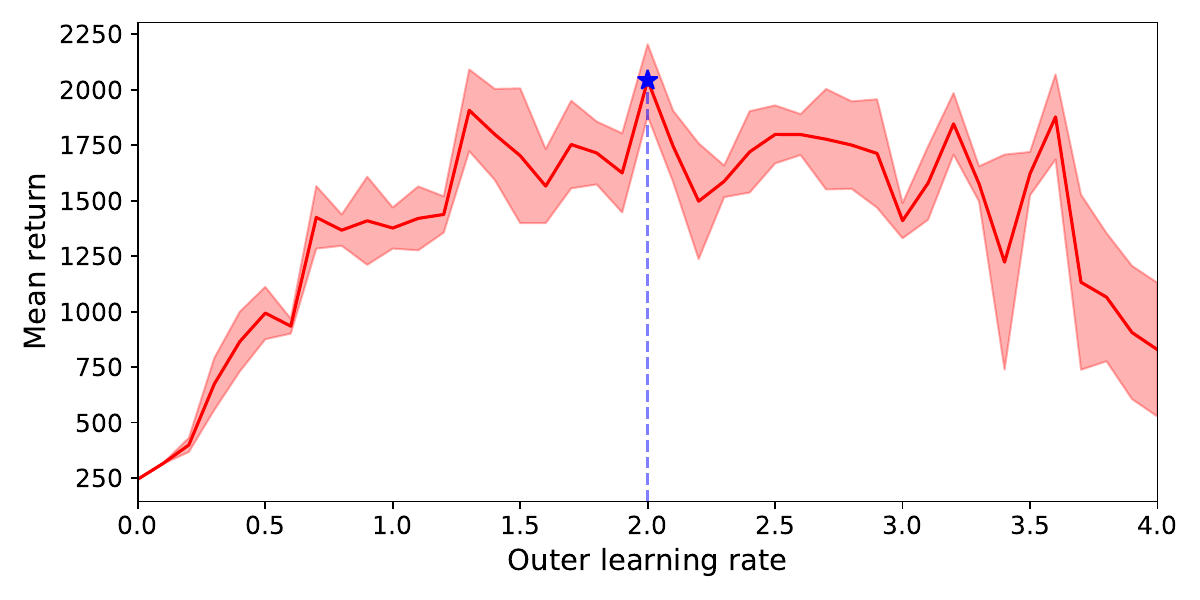}
    \caption{walker2d}
\end{subfigure}
    \caption{\textbf{Outer learning rate sweep performance for Brax tasks.} Mean of 4 seeds shown with standard error shaded. Optimal point marked with blue star.}
\end{figure}

\begin{figure}[H]
\centering
\begin{subfigure}{0.3\textwidth}
    \includegraphics[width=\textwidth]{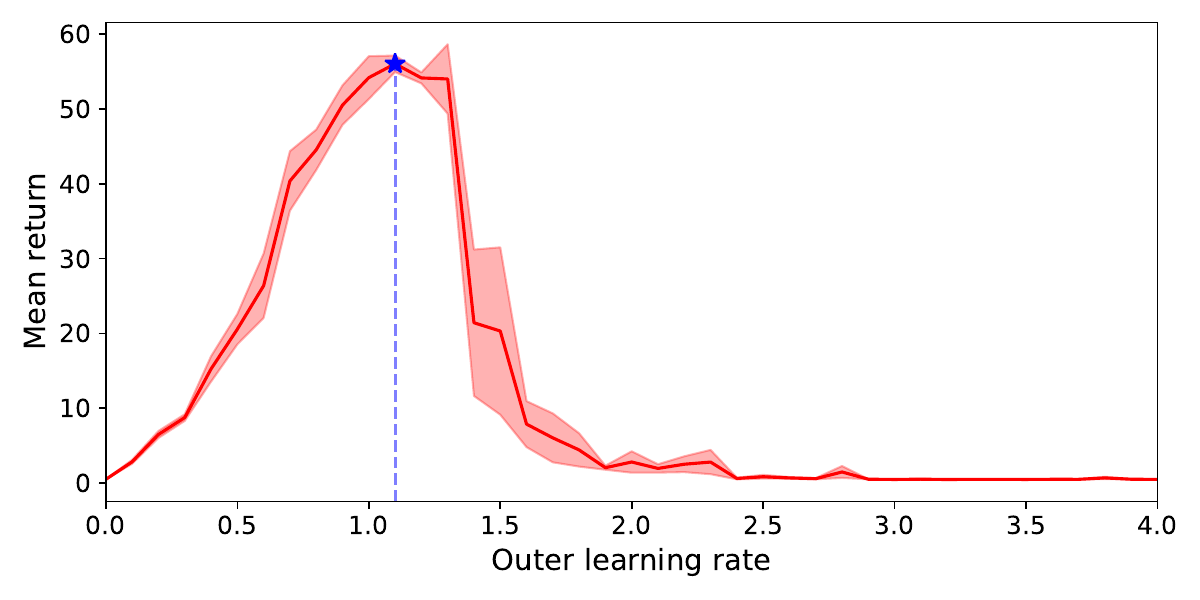}
    \caption{asterix}
\end{subfigure}
\hspace{1.5cm}
\begin{subfigure}{0.3\textwidth}
    \includegraphics[width=\textwidth]{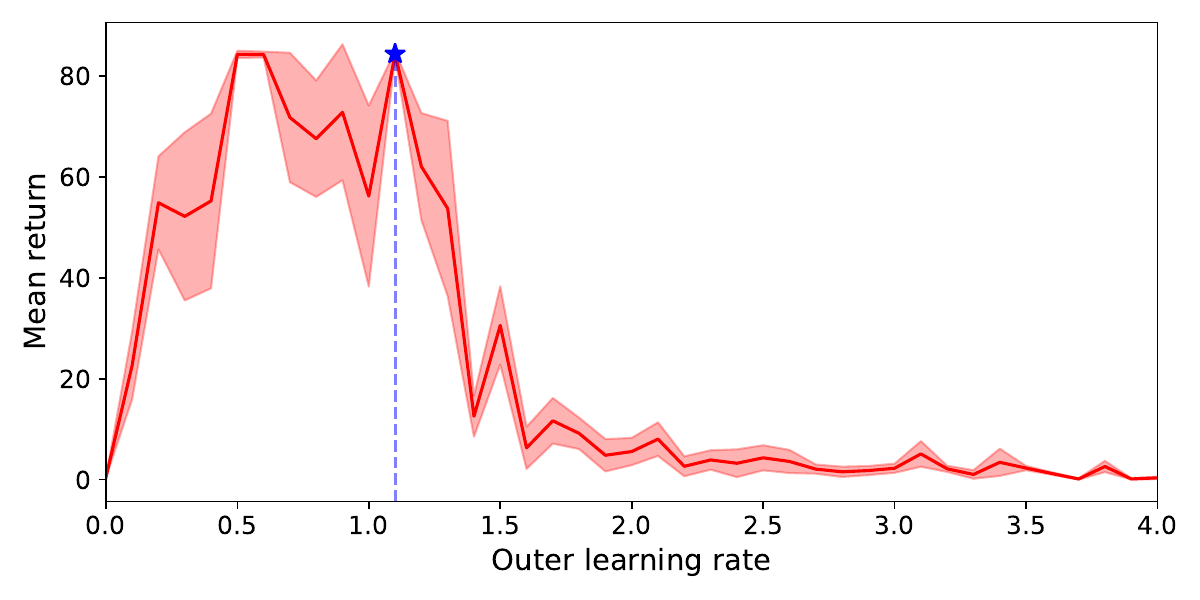}
    \caption{breakout}
\end{subfigure}
\\
\begin{subfigure}{0.3\textwidth}
    \includegraphics[width=\textwidth]{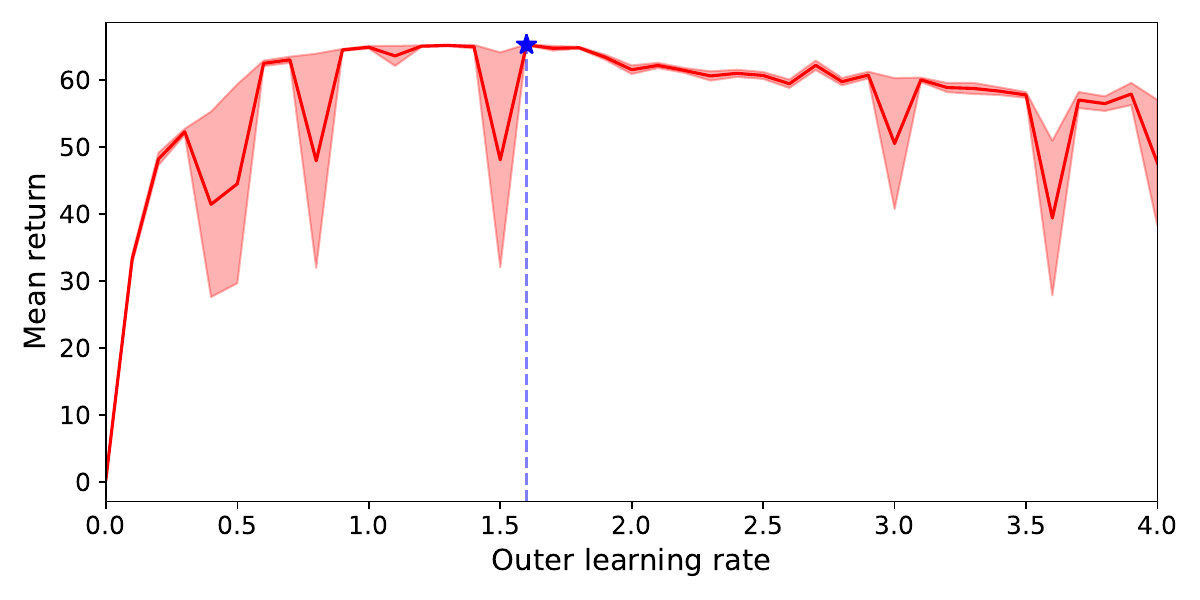}
    \caption{freeway}
\end{subfigure}
\hspace{1.5cm}
\begin{subfigure}{0.3\textwidth}
    \includegraphics[width=\textwidth]{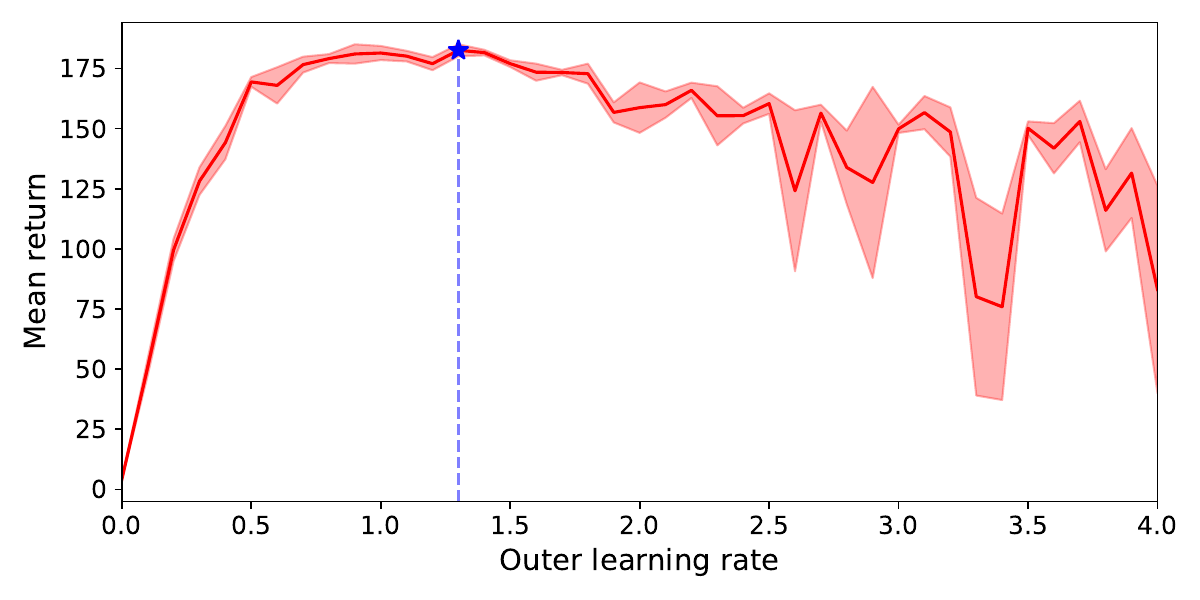}
    \caption{space\_invaders}
\end{subfigure}
    \caption{\textbf{Baseline sweep performance for MinAtar tasks.} Mean of 4 seeds shown with standard error shaded. Optimal point marked with blue star.}
\end{figure}

\begin{figure}[H]
\centering
\begin{subfigure}{0.3\textwidth}
    \includegraphics[width=\textwidth]{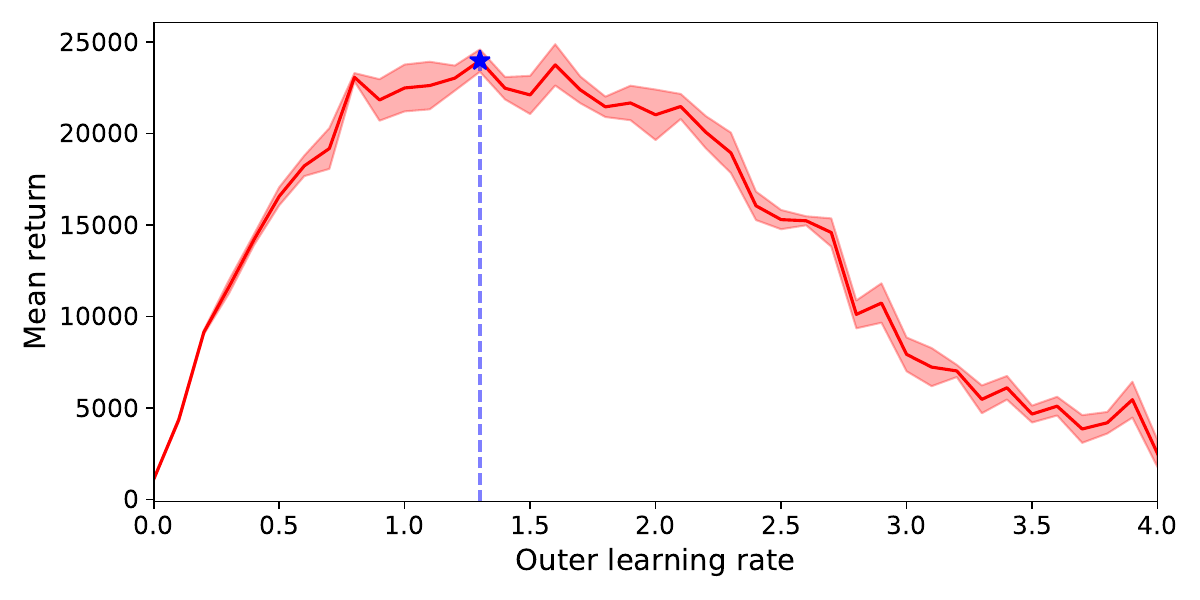}
    \caption{game\_2048}
\end{subfigure}
\hspace{1.5cm}
\begin{subfigure}{0.3\textwidth}
    \includegraphics[width=\textwidth]{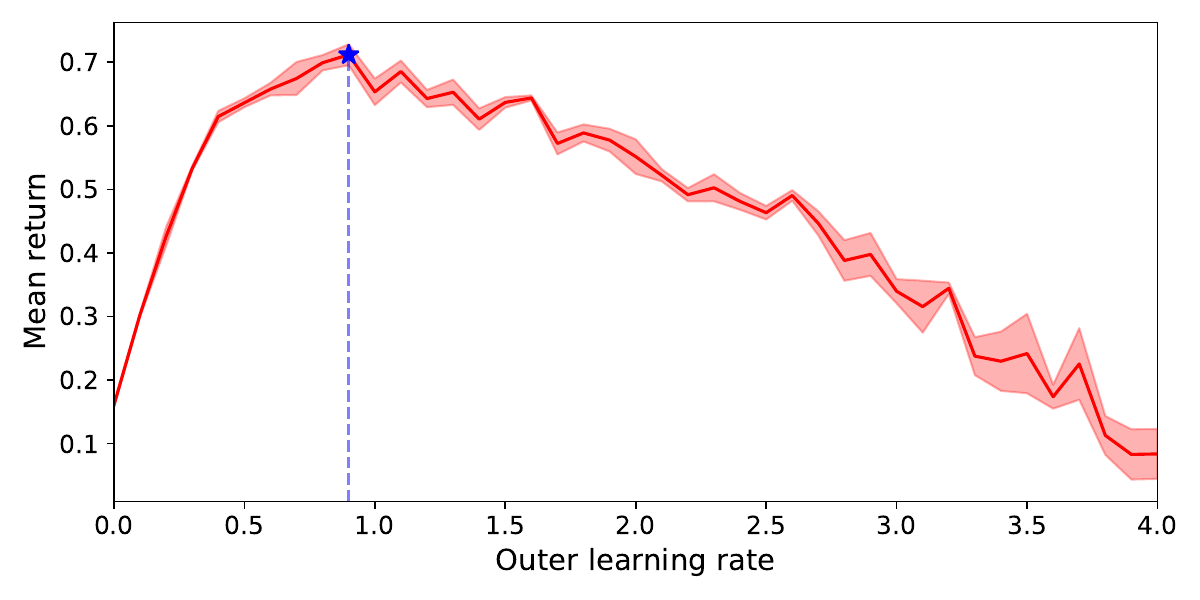}
    \caption{maze}
\end{subfigure}
\\
\begin{subfigure}{0.3\textwidth}
    \includegraphics[width=\textwidth]{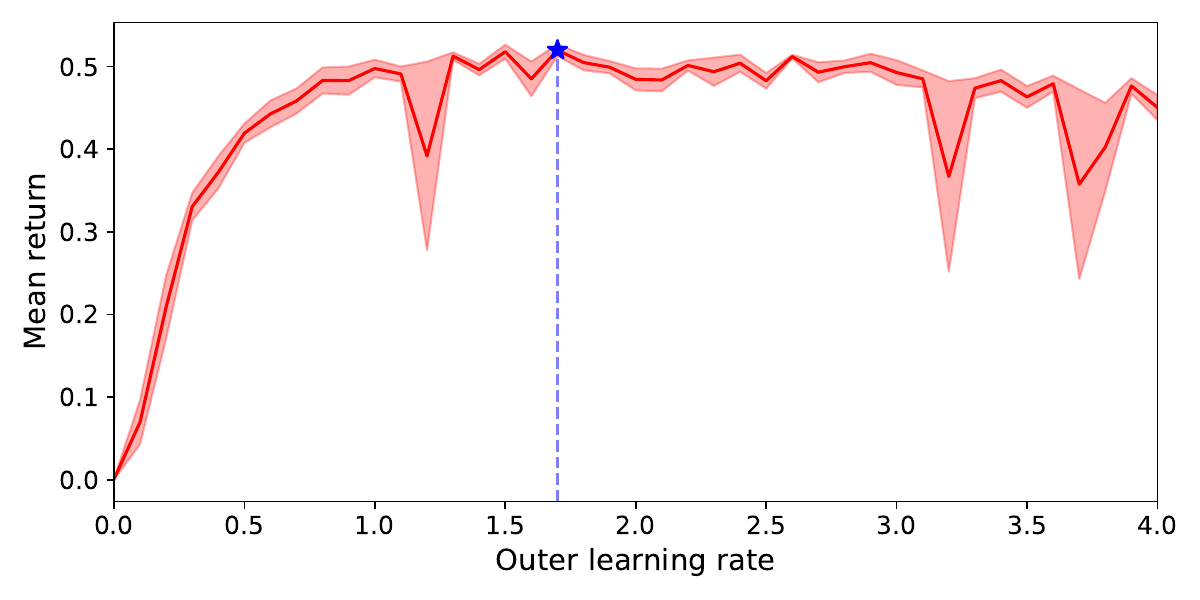}
    \caption{rubiks\_cube}
\end{subfigure}
\hspace{1.5cm}
\begin{subfigure}{0.3\textwidth}
    \includegraphics[width=\textwidth]{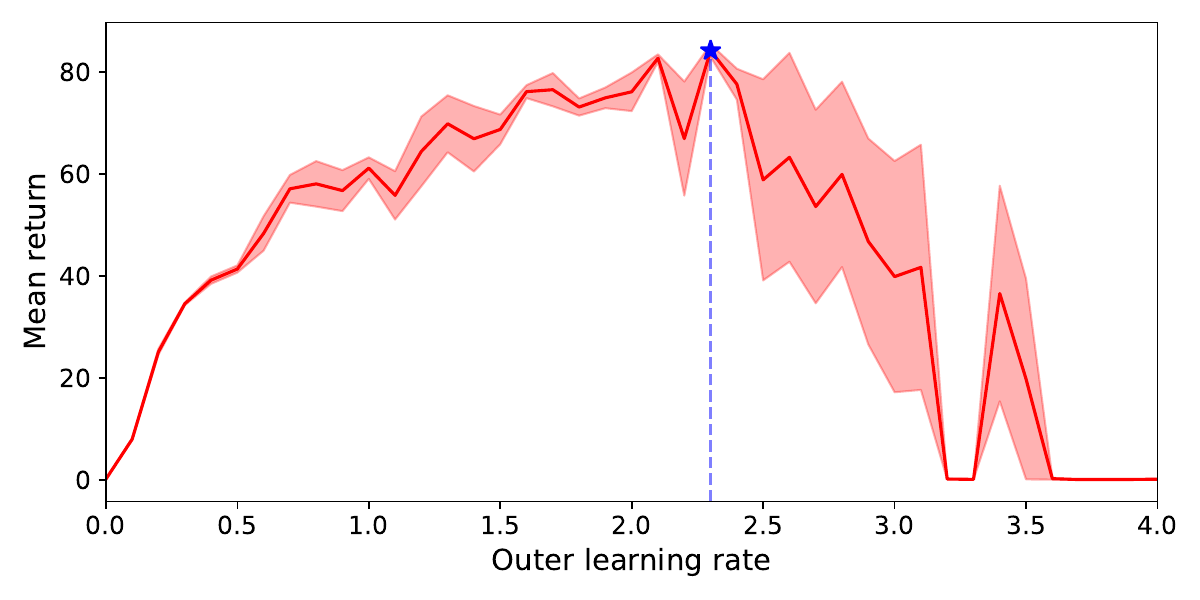}
    \caption{snake}
\end{subfigure}
    \caption{\textbf{Baseline sweep performance for Jumajji tasks.} Mean of 4 seeds shown with standard error shaded. Optimal point marked with blue star.}
\end{figure}

\clearpage
\subsection{Nesterov Momentum}

\begin{figure}[h]
\centering
\begin{subfigure}{0.3\textwidth}
    \includegraphics[width=\textwidth]{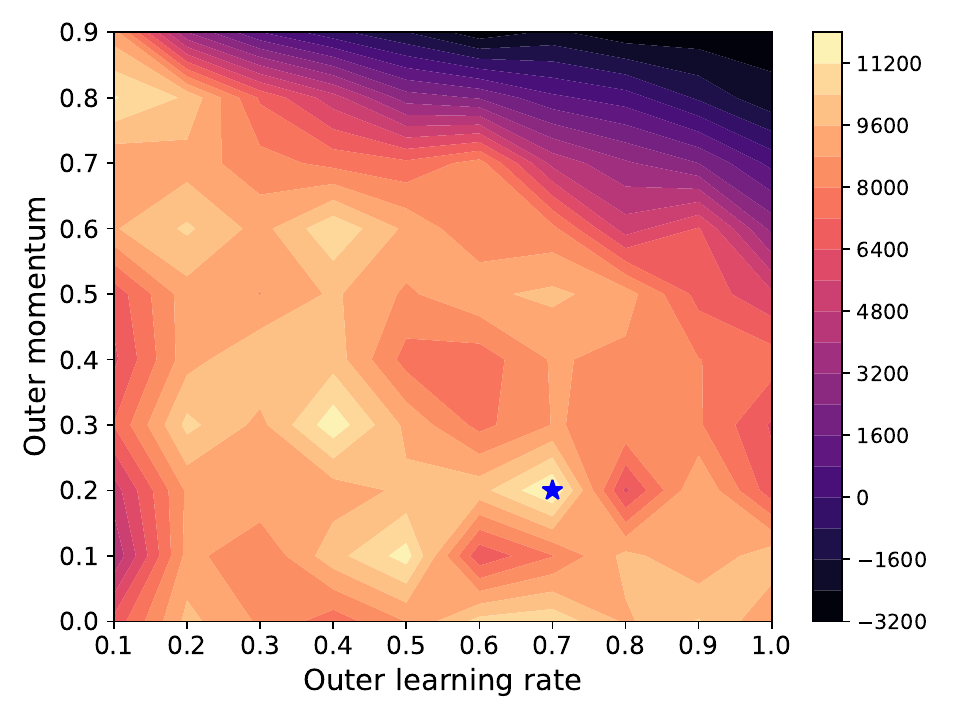}
    \caption{ant}
\end{subfigure}
\hfill
\begin{subfigure}{0.3\textwidth}
    \includegraphics[width=\textwidth]{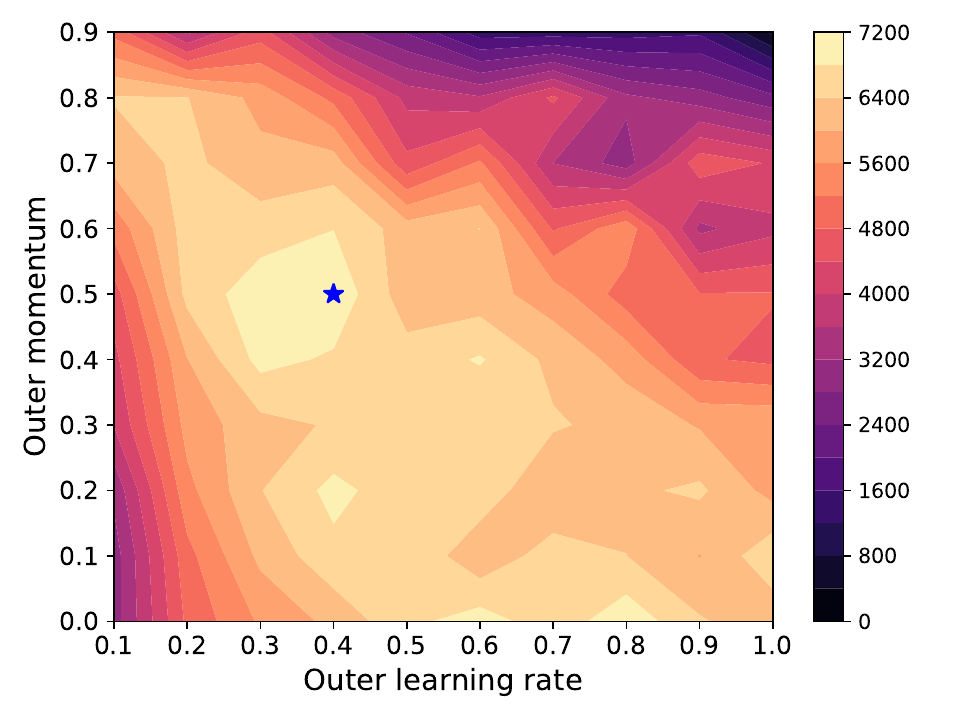}
    \caption{halfcheetah}
\end{subfigure}
\hfill
\begin{subfigure}{0.3\textwidth}
    \includegraphics[width=\textwidth]{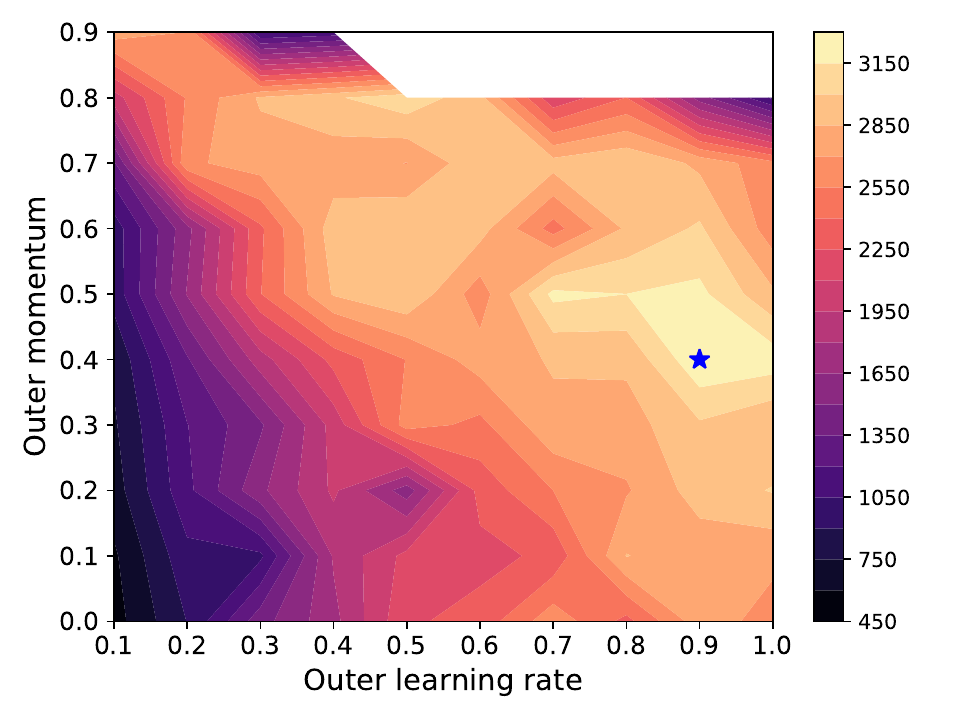}
    \caption{hopper}
\end{subfigure}
\hfill
\begin{subfigure}{0.3\textwidth}
    \includegraphics[width=\textwidth]{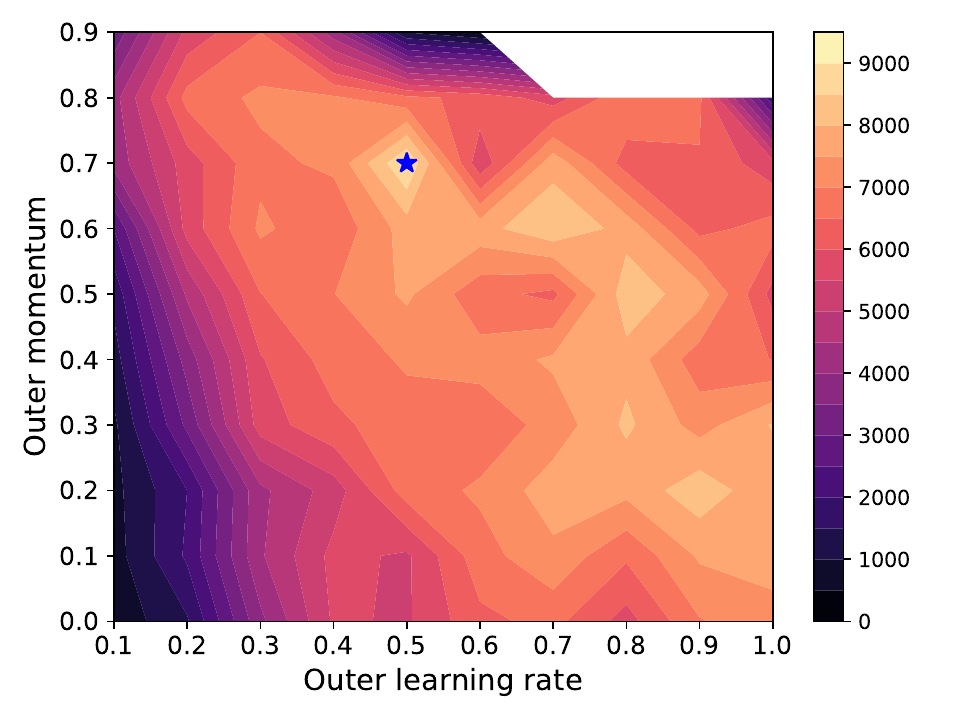}
    \caption{humanoid}
\end{subfigure}
\hfill
\begin{subfigure}{0.3\textwidth}
    \includegraphics[width=\textwidth]{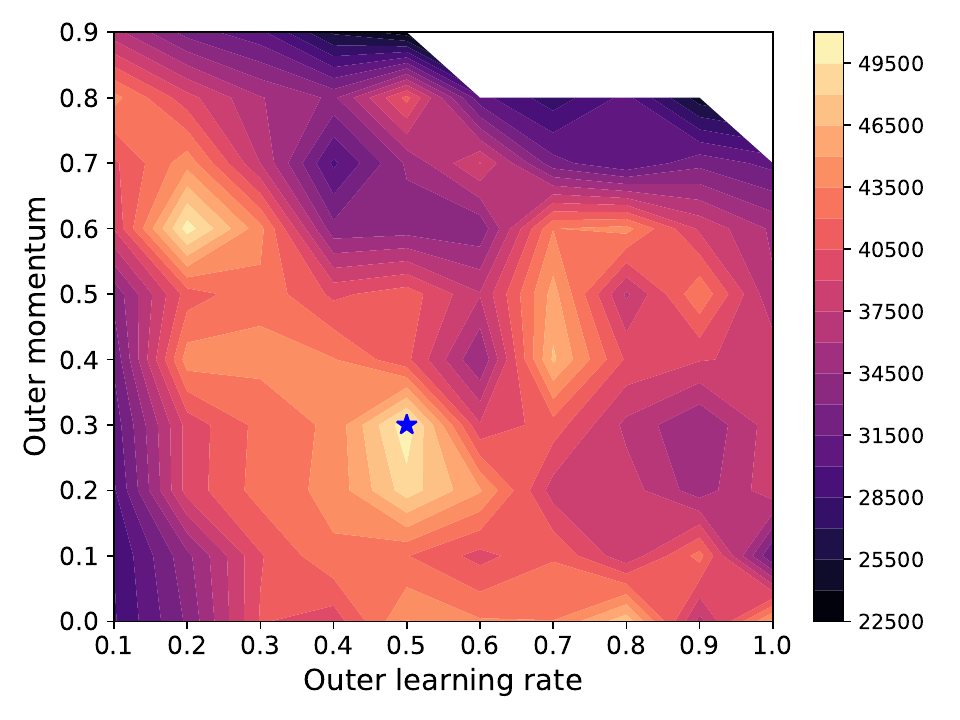}
    \caption{humanoidstandup}
\end{subfigure}
\hfill
\begin{subfigure}{0.3\textwidth}
    \includegraphics[width=\textwidth]{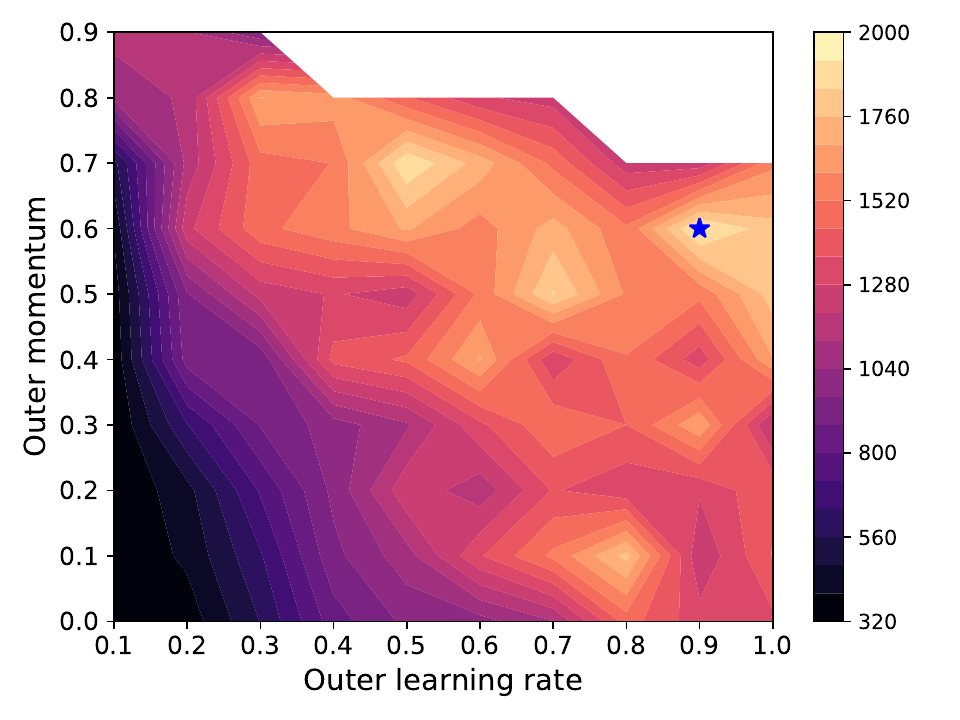}
    \caption{walker2d}
\end{subfigure}
    \caption{\textbf{Nesterov sweep performance for Brax tasks.} Contour plot of mean of 4 seeds. White regions resulted in numerical errors (NaN). Optimal point marked with blue star.}
\end{figure}

\begin{figure}[h]
\centering
\begin{subfigure}{0.3\textwidth}
    \includegraphics[width=\textwidth]{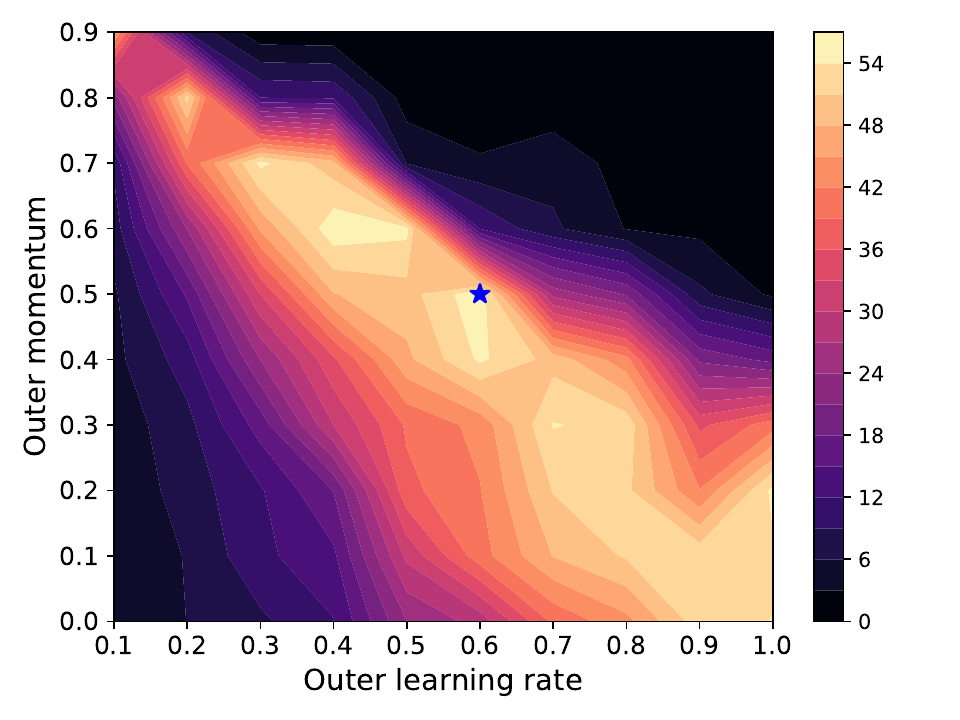}
    \caption{asterix}
\end{subfigure}
\hspace{1.5cm}
\begin{subfigure}{0.3\textwidth}
    \includegraphics[width=\textwidth]{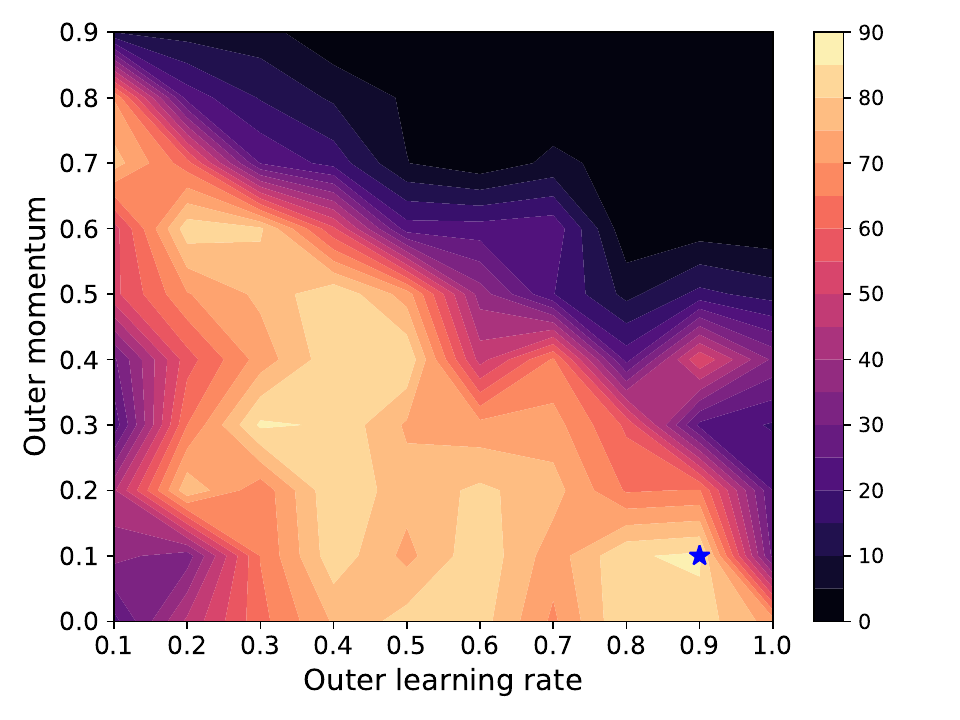}
    \caption{breakout}
\end{subfigure}
\\
\begin{subfigure}{0.3\textwidth}
    \includegraphics[width=\textwidth]{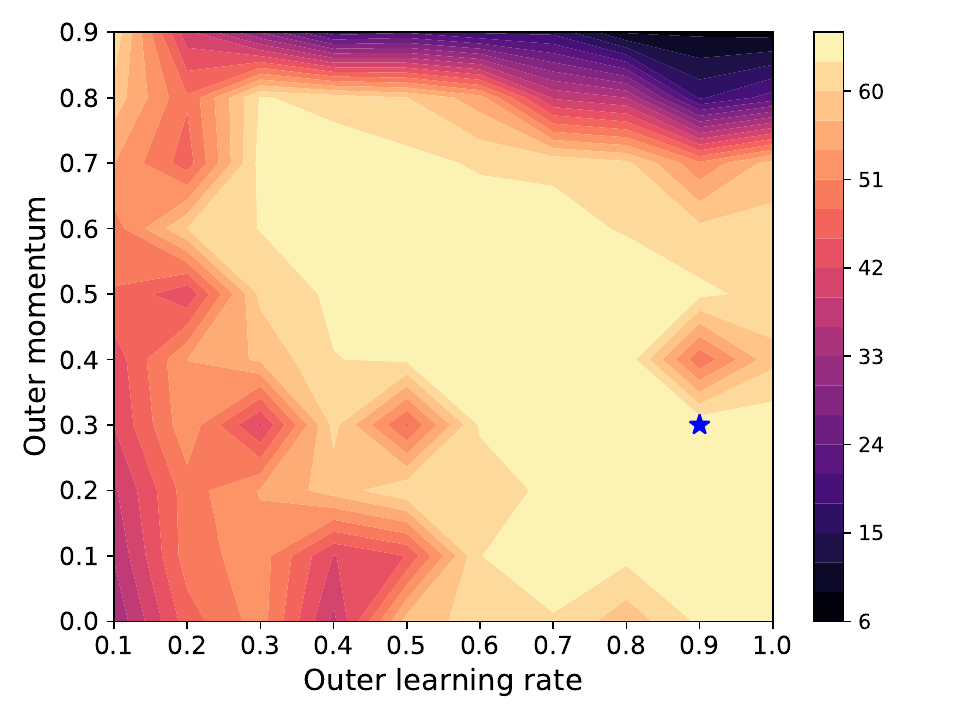}
    \caption{freeway}
\end{subfigure}
\hspace{1.5cm}
\begin{subfigure}{0.3\textwidth}
    \includegraphics[width=\textwidth]{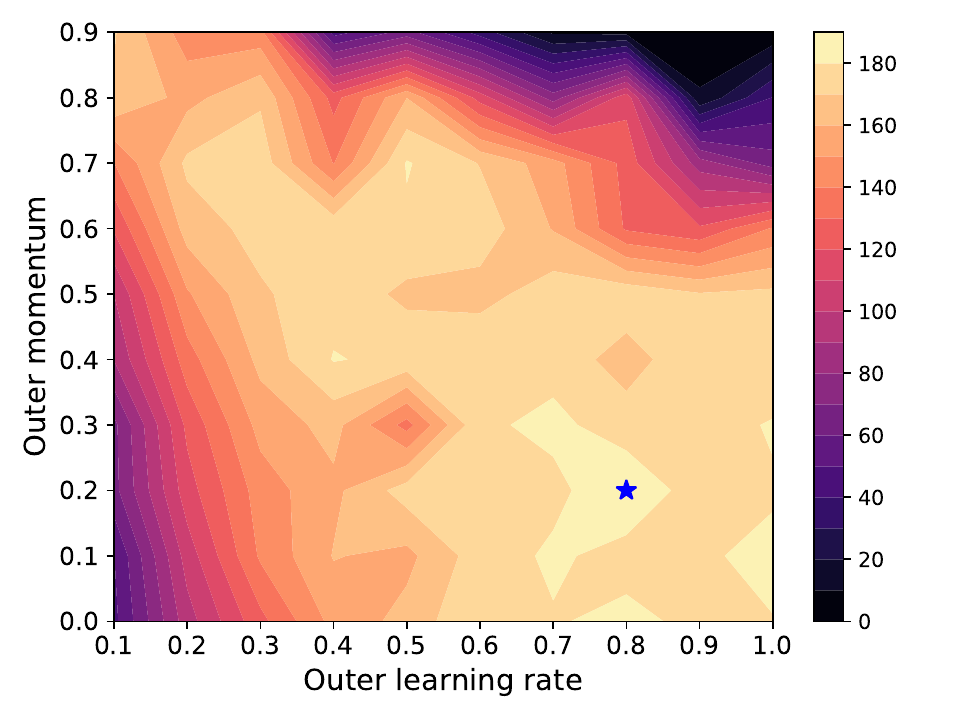}
    \caption{space\_invaders}
\end{subfigure}
    \caption{\textbf{Nesterov sweep performance for MinAtar tasks.} Contour plot of mean of 4 seeds. White regions resulted in numerical errors (NaN). Optimal point marked with blue star.}
\end{figure}

\begin{figure}[h]
\centering
\begin{subfigure}{0.3\textwidth}
    \includegraphics[width=\textwidth]{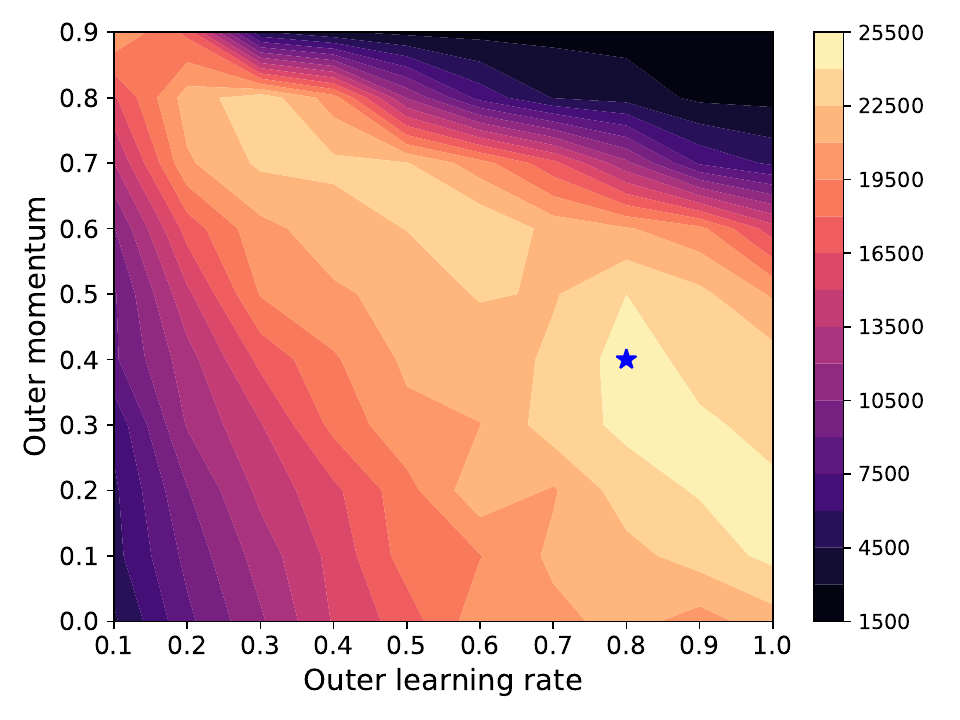}
    \caption{game\_2048}
\end{subfigure}
\hspace{1.5cm}
\begin{subfigure}{0.3\textwidth}
    \includegraphics[width=\textwidth]{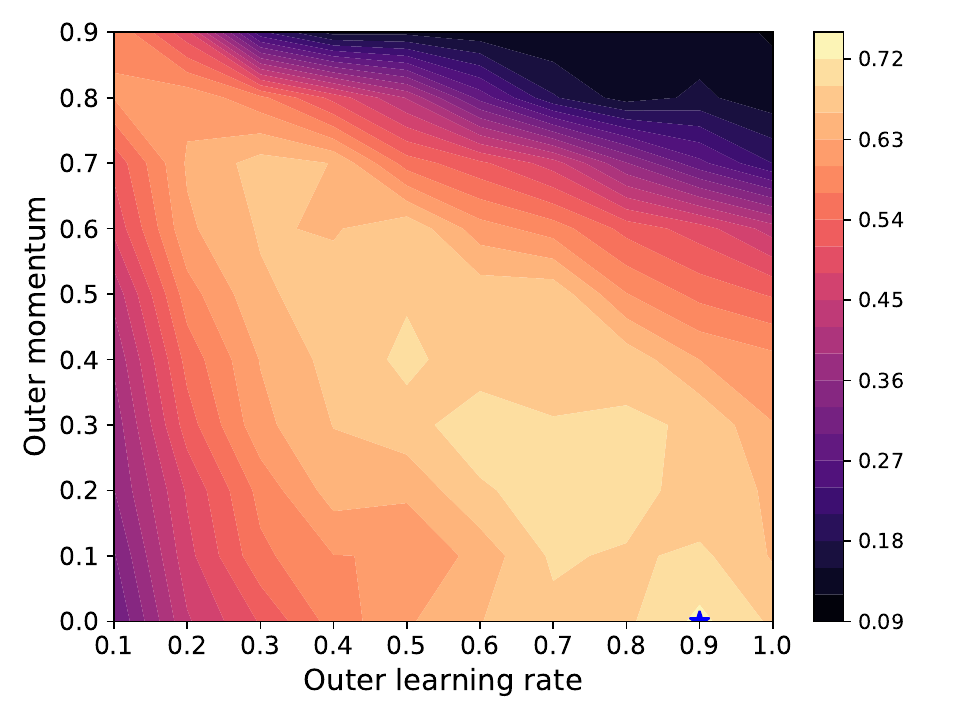}
    \caption{maze}
\end{subfigure}
\\
\begin{subfigure}{0.3\textwidth}
    \includegraphics[width=\textwidth]{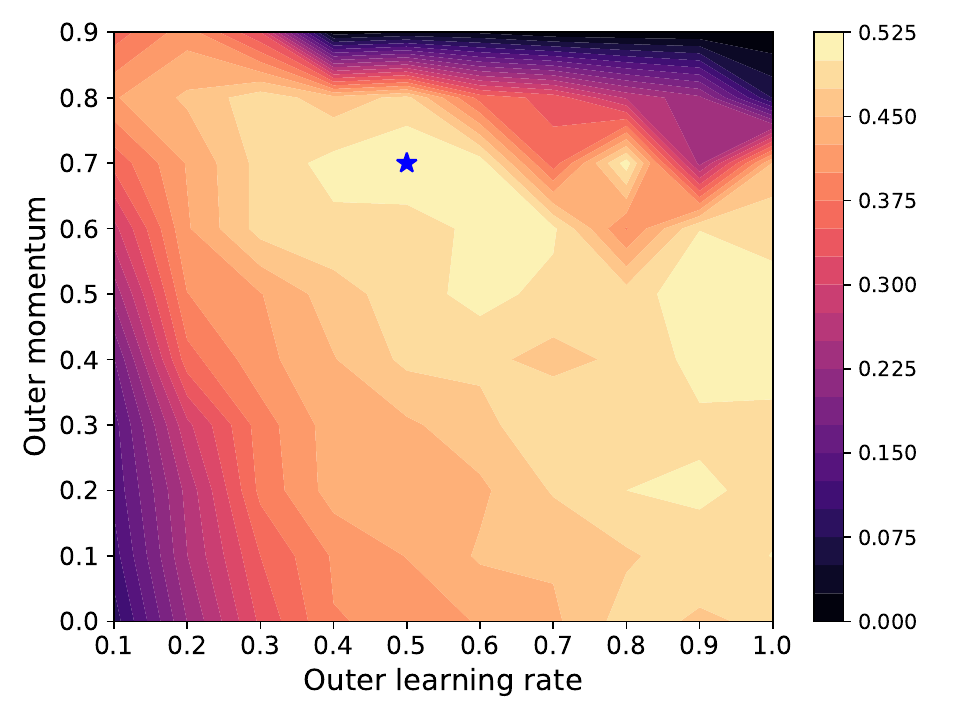}
    \caption{rubiks\_cube}
\end{subfigure}
\hspace{1.5cm}
\begin{subfigure}{0.3\textwidth}
    \includegraphics[width=\textwidth]{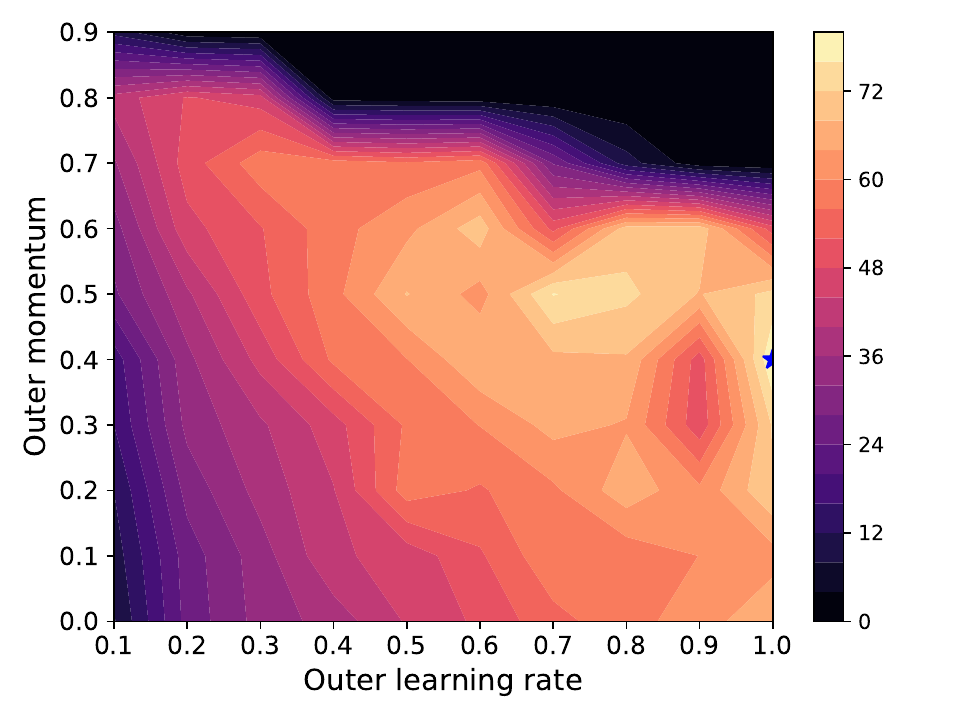}
    \caption{snake}
\end{subfigure}
    \caption{\textbf{Nesterov sweep performance for Jumanji tasks.} Contour plot of mean of 4 seeds. White regions resulted in numerical errors (NaN). Optimal point marked with blue star.}
\end{figure}

\vspace{1cm}

\subsection{Biased Initialization}

\vspace{1cm}

\begin{figure}[H]
\centering
\begin{subfigure}{0.3\textwidth}
    \includegraphics[width=\textwidth]{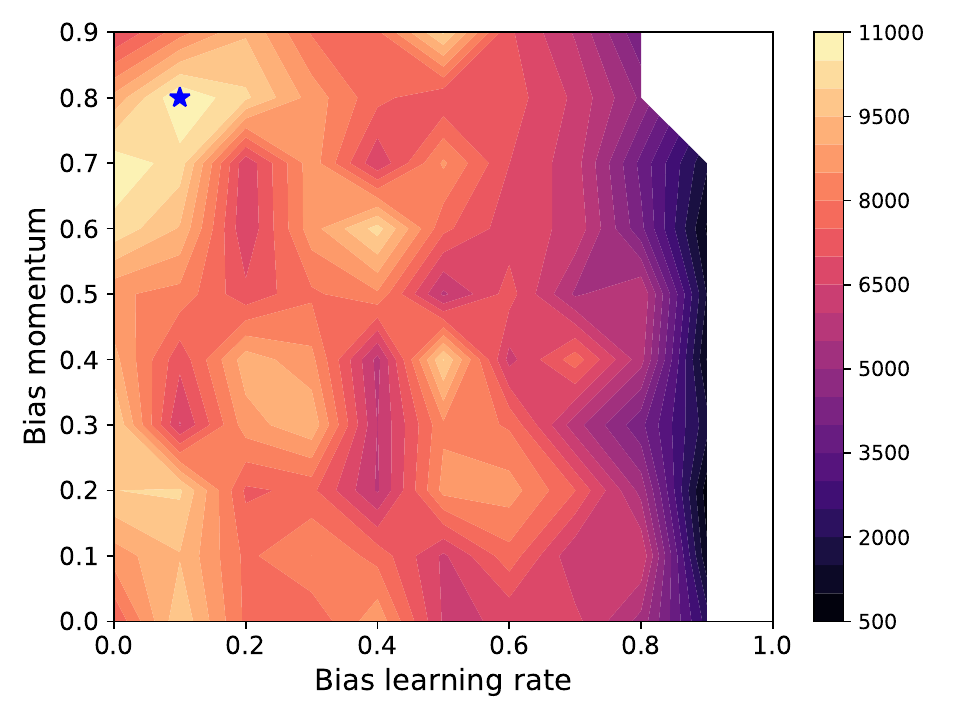}
    \caption{ant}
\end{subfigure}
\hfill
\begin{subfigure}{0.3\textwidth}
    \includegraphics[width=\textwidth]{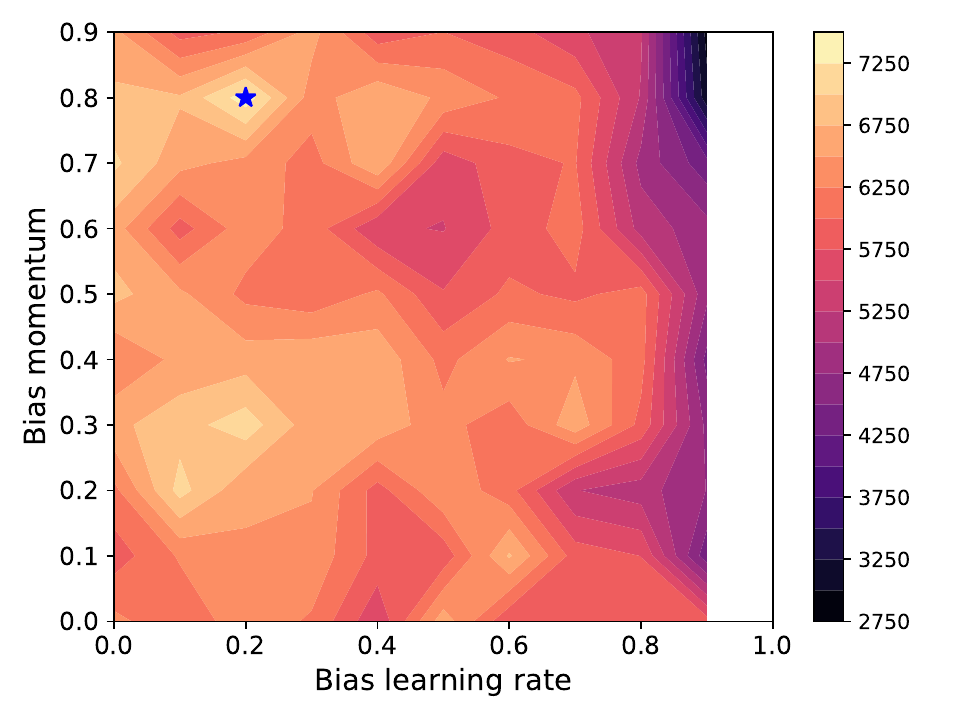}
    \caption{halfcheetah}
\end{subfigure}
\hfill
\begin{subfigure}{0.3\textwidth}
    \includegraphics[width=\textwidth]{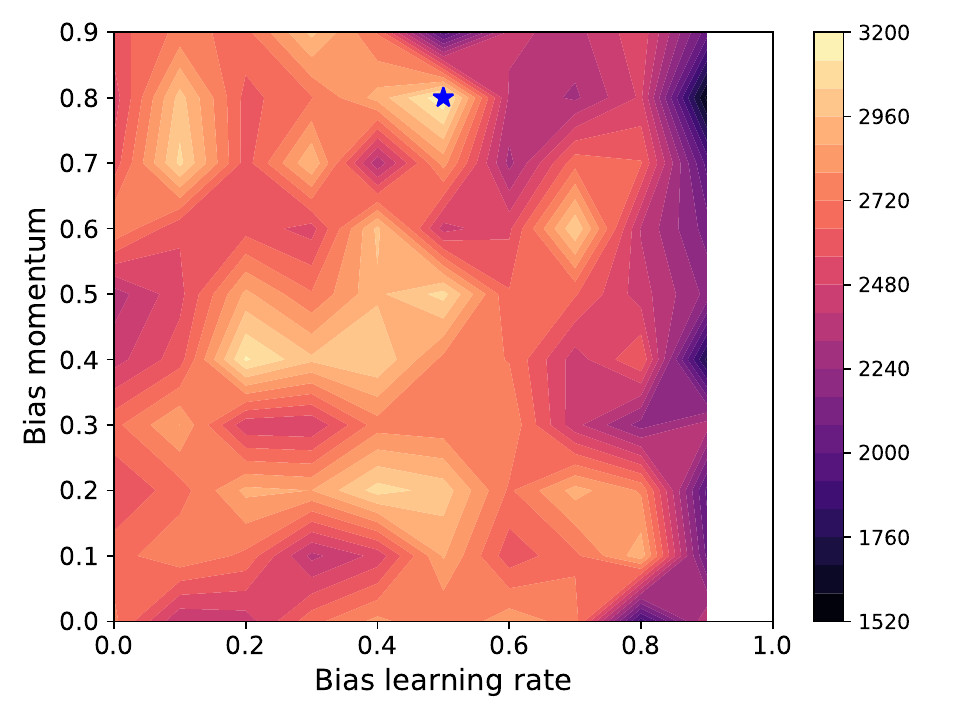}
    \caption{hopper}
\end{subfigure}
\hfill
\begin{subfigure}{0.3\textwidth}
    \includegraphics[width=\textwidth]{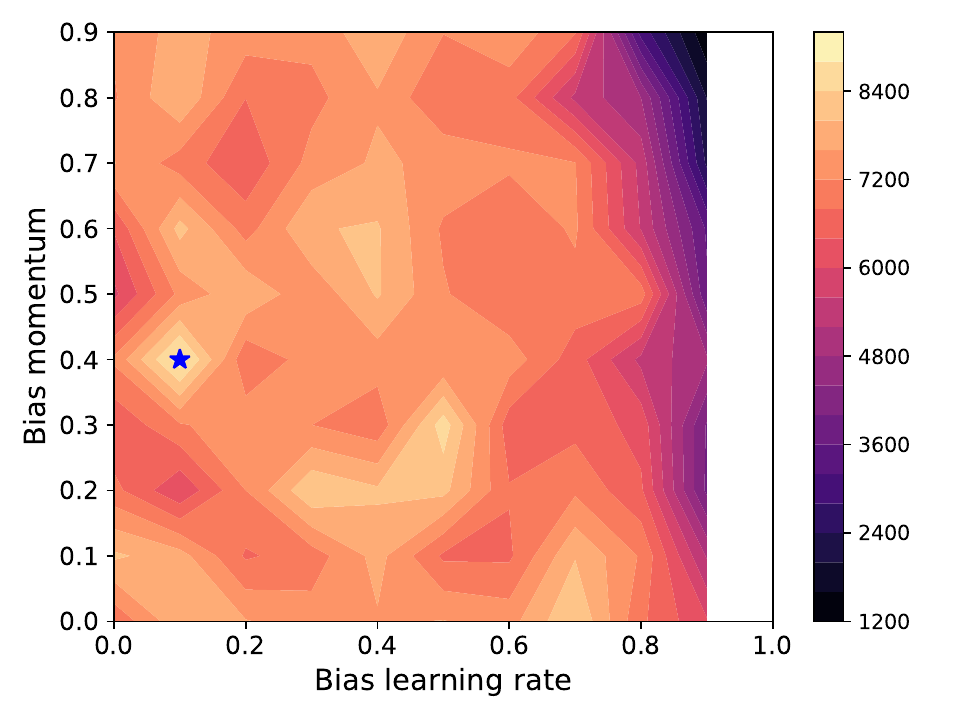}
    \caption{humanoid}
\end{subfigure}
\hfill
\begin{subfigure}{0.3\textwidth}
    \includegraphics[width=\textwidth]{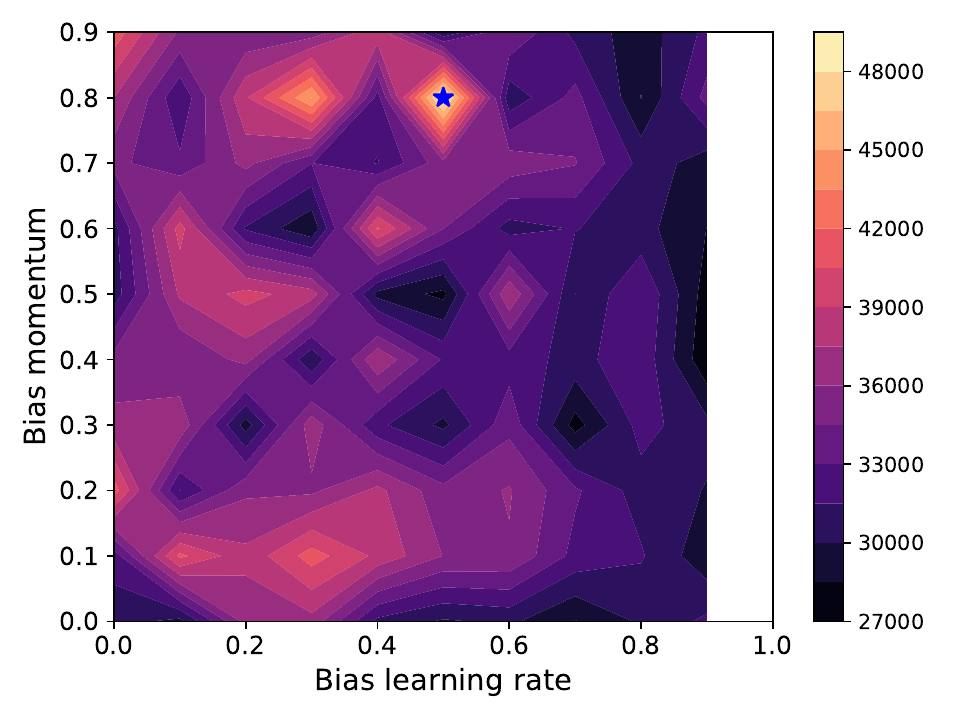}
    \caption{humanoidstandup}
\end{subfigure}
\hfill
\begin{subfigure}{0.3\textwidth}
    \includegraphics[width=\textwidth]{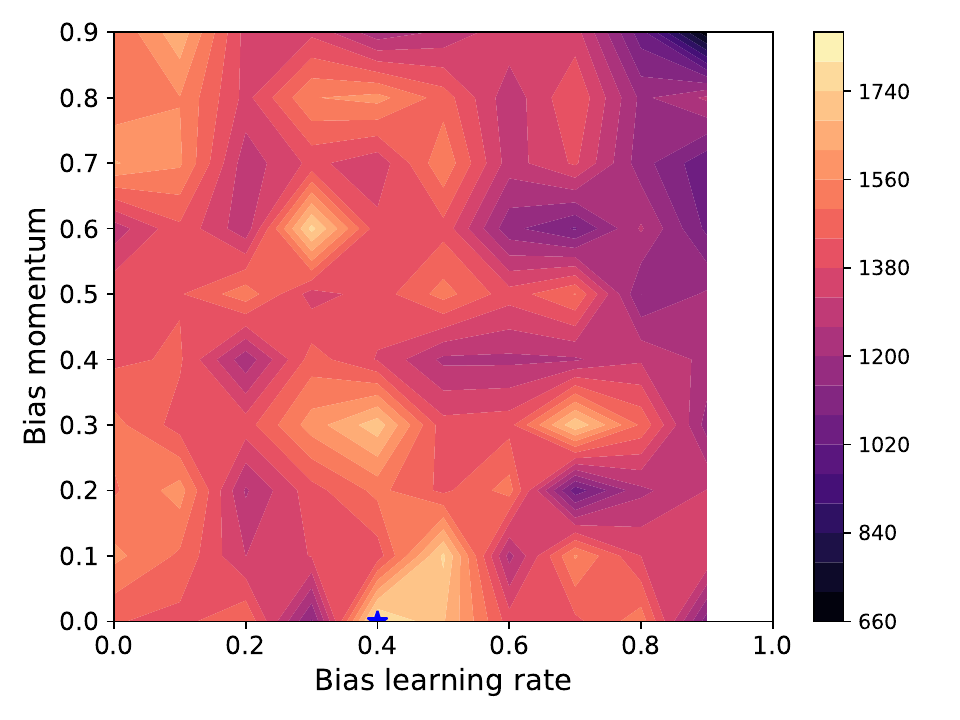}
    \caption{walker2d}
\end{subfigure}
    \caption{\textbf{Biased initialization sweep performance for Brax tasks.} Contour plot of mean of 4 seeds. White regions resulted in numerical errors (NaN). Optimal point marked with blue star.}
\end{figure}

\begin{figure}
\centering
\begin{subfigure}{0.3\textwidth}
    \includegraphics[width=\textwidth]{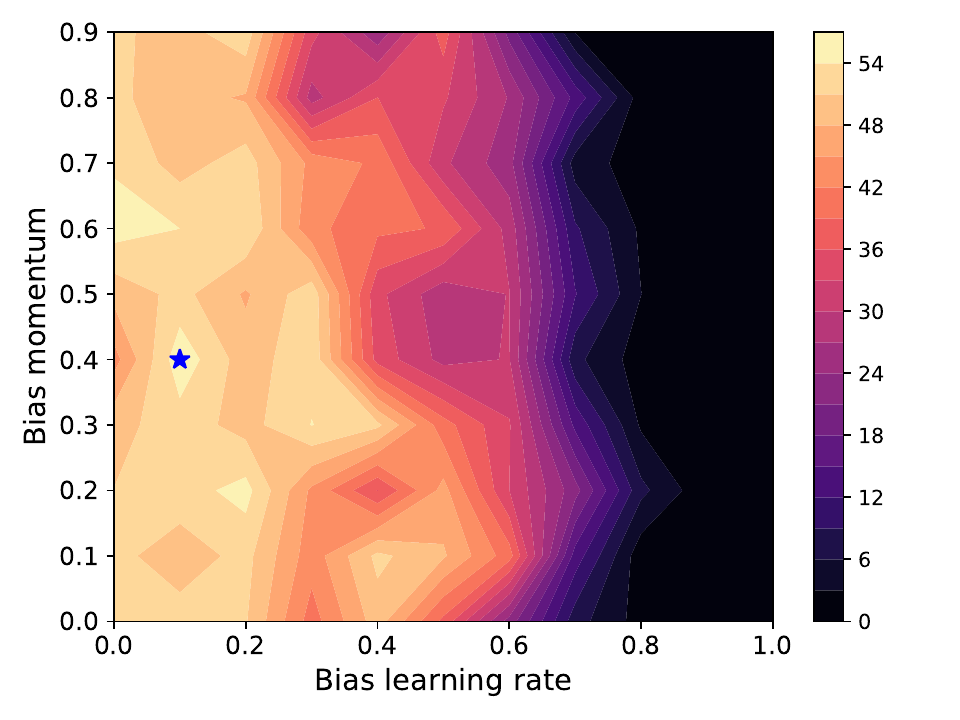}
    \caption{asterix}
\end{subfigure}
\hspace{1.5cm}
\begin{subfigure}{0.3\textwidth}
    \includegraphics[width=\textwidth]{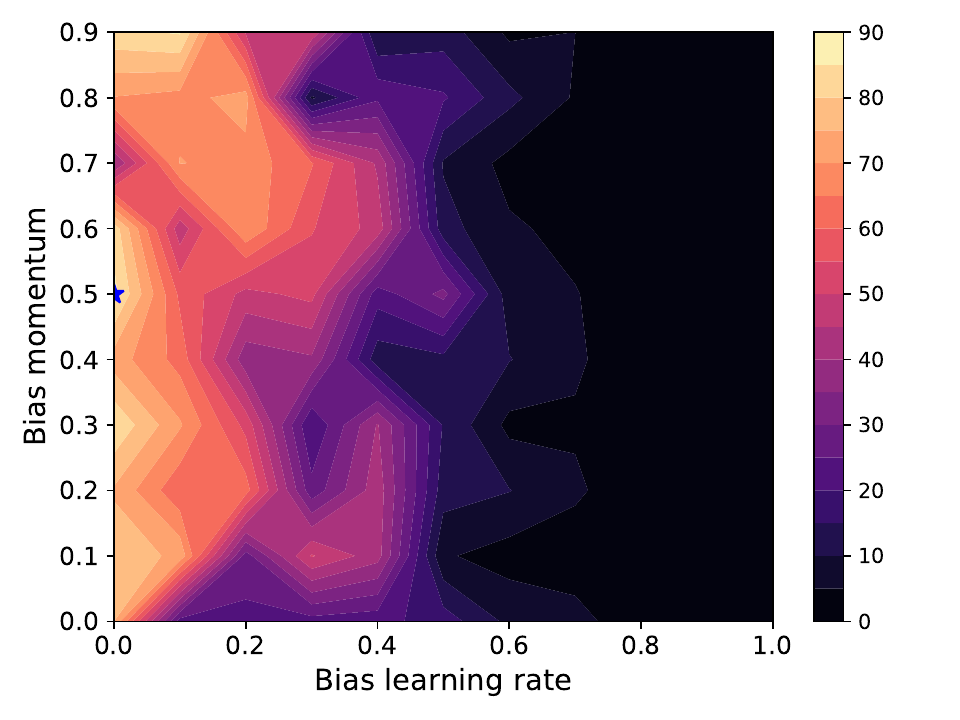}
    \caption{breakout}
\end{subfigure}
\\
\begin{subfigure}{0.3\textwidth}
    \includegraphics[width=\textwidth]{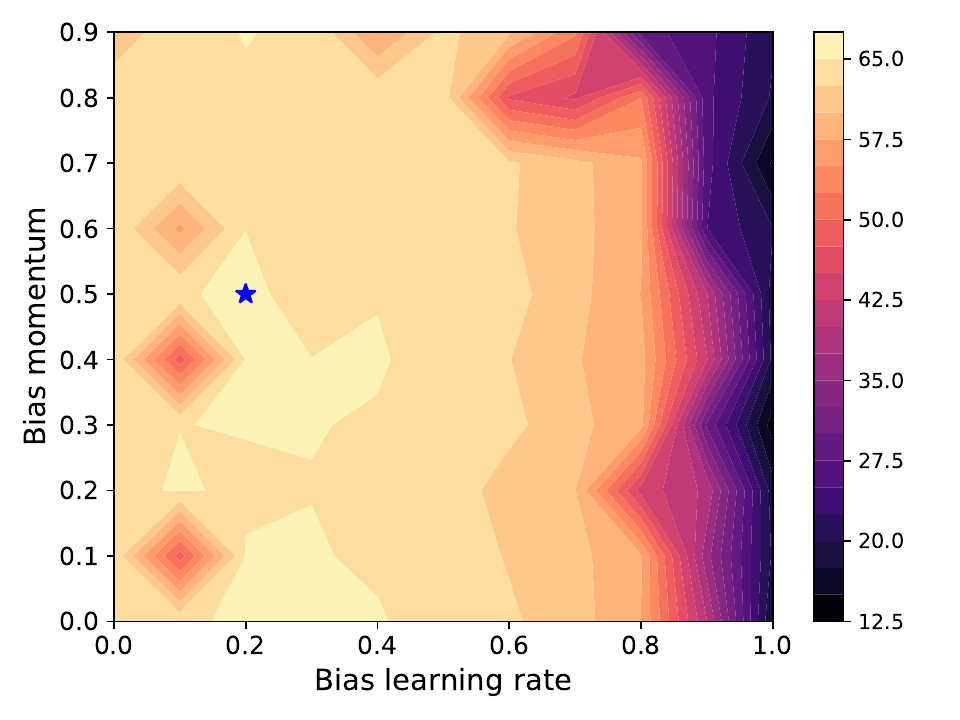}
    \caption{freeway}
\end{subfigure}
\hspace{1.5cm}
\begin{subfigure}{0.3\textwidth}
    \includegraphics[width=\textwidth]{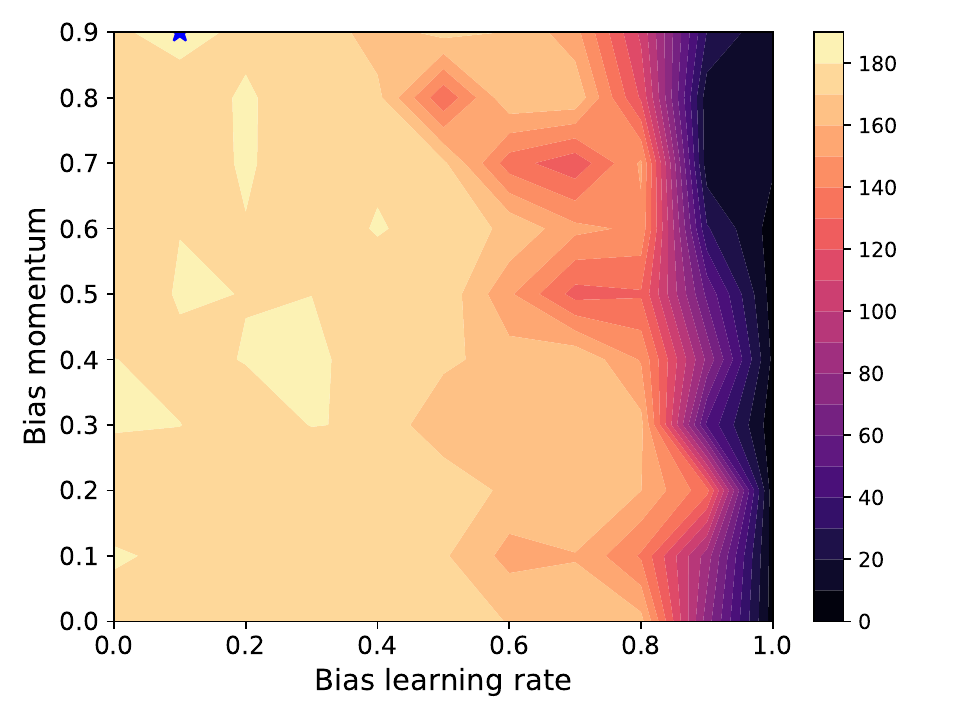}
    \caption{space\_invaders}
\end{subfigure}
    \caption{\textbf{Biased initialization sweep performance for MinAtar tasks.} Contour plot of mean of 4 seeds. White regions resulted in numerical errors (NaN). Optimal point marked with blue star.}
\end{figure}

\begin{figure}
\centering
\begin{subfigure}{0.3\textwidth}
    \includegraphics[width=\textwidth]{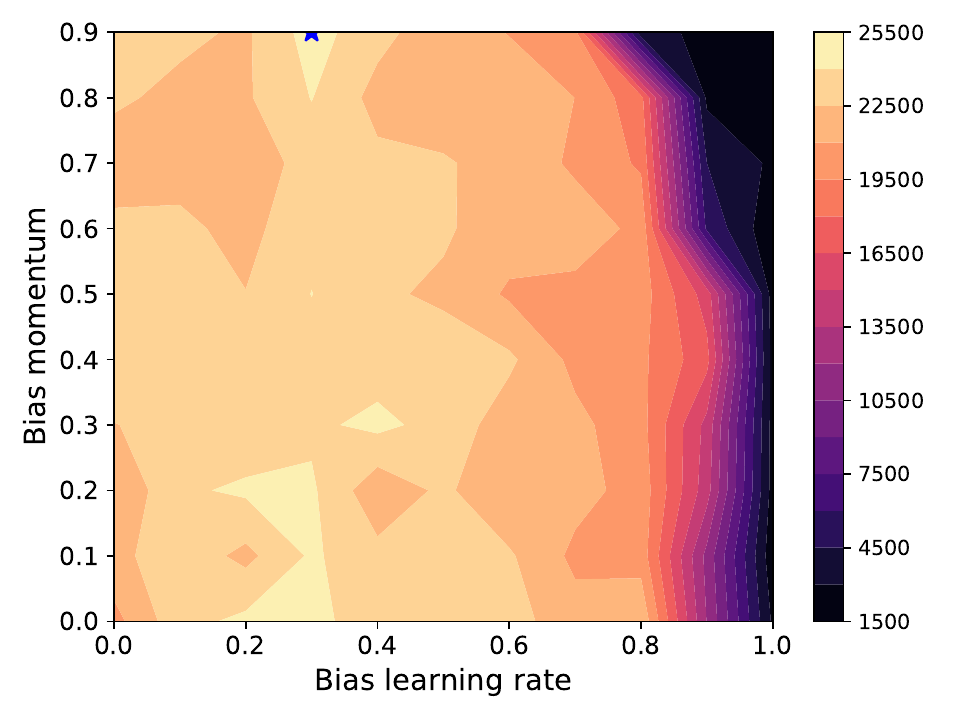}
    \caption{game\_2048}
\end{subfigure}
\hspace{1.5cm}
\begin{subfigure}{0.3\textwidth}
    \includegraphics[width=\textwidth]{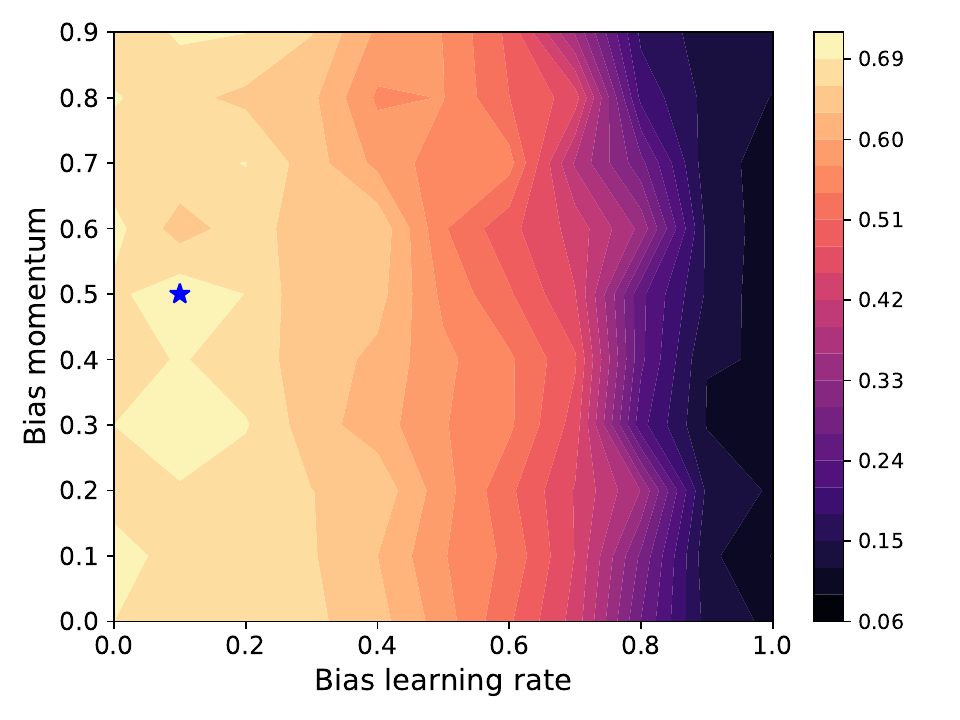}
    \caption{maze}
\end{subfigure}
\\
\begin{subfigure}{0.3\textwidth}
    \includegraphics[width=\textwidth]{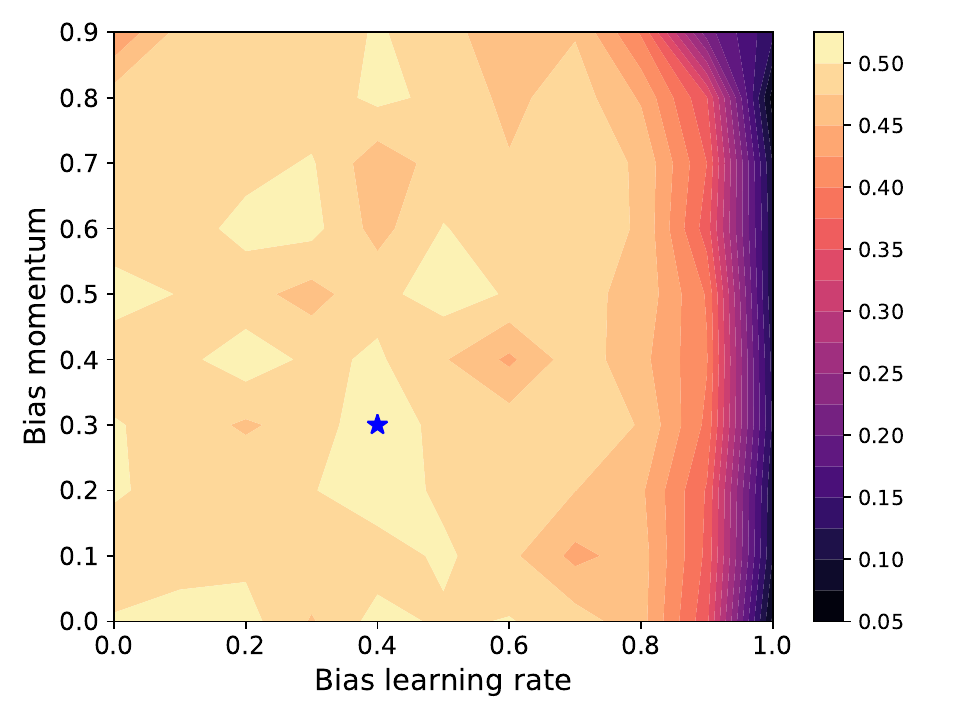}
    \caption{rubiks\_cube}
\end{subfigure}
\hspace{1.5cm}
\begin{subfigure}{0.3\textwidth}
    \includegraphics[width=\textwidth]{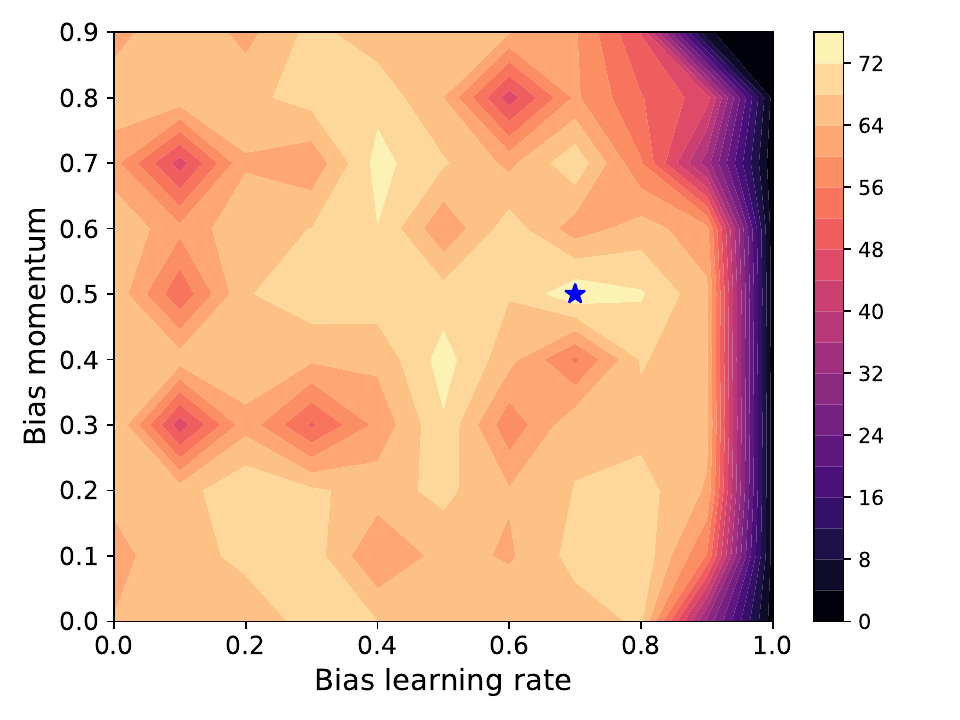}
    \caption{snake}
\end{subfigure}
    \caption{\textbf{Biased initialization sweep performance for Jumanji tasks.} Contour plot of mean of 4 seeds. White regions resulted in numerical errors (NaN). Optimal point marked with blue star.}
\end{figure}

\clearpage
\section{Individual Task Performances}

\begin{figure}[h]
\centering
\begin{subfigure}{0.25\textwidth}
    \includegraphics[width=\textwidth]{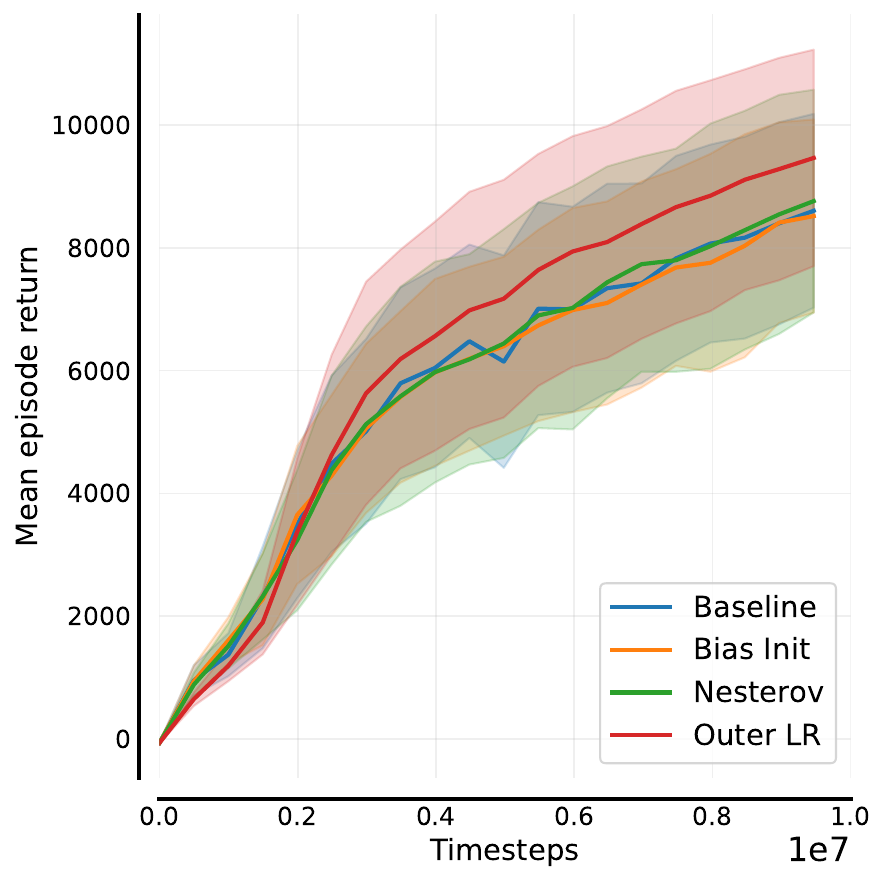}
    \caption{ant}
\end{subfigure}
\hfill
\begin{subfigure}{0.25\textwidth}
    \includegraphics[width=\textwidth]{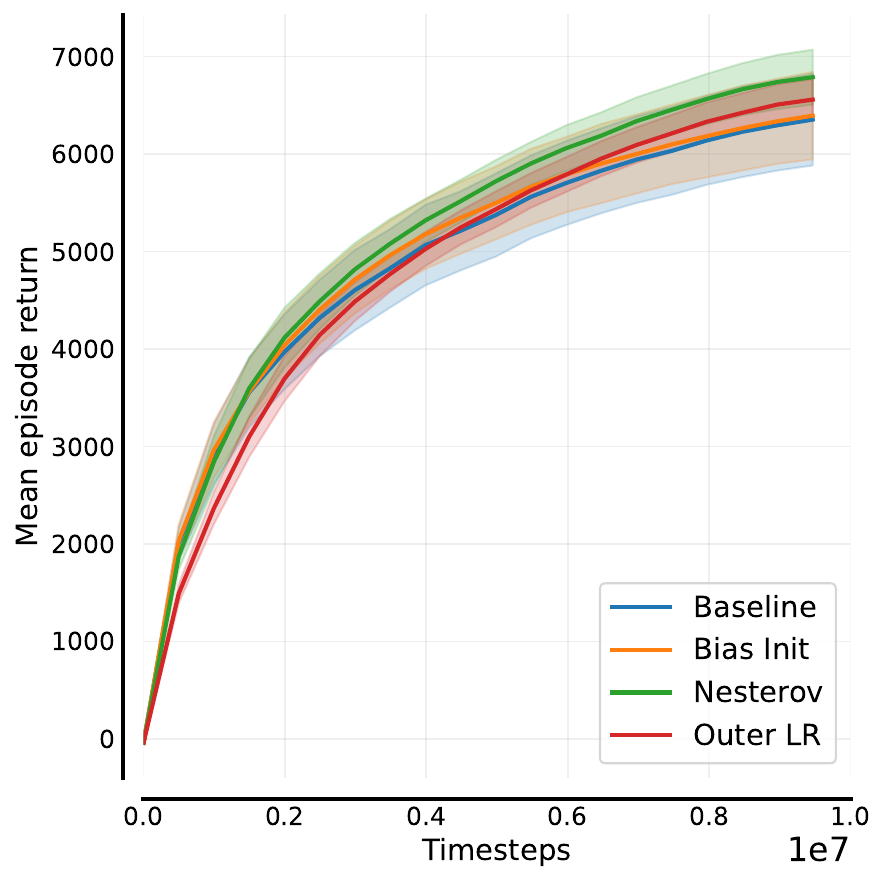}
    \caption{halfcheetah}
\end{subfigure}
\hfill
\begin{subfigure}{0.25\textwidth}
    \includegraphics[width=\textwidth]{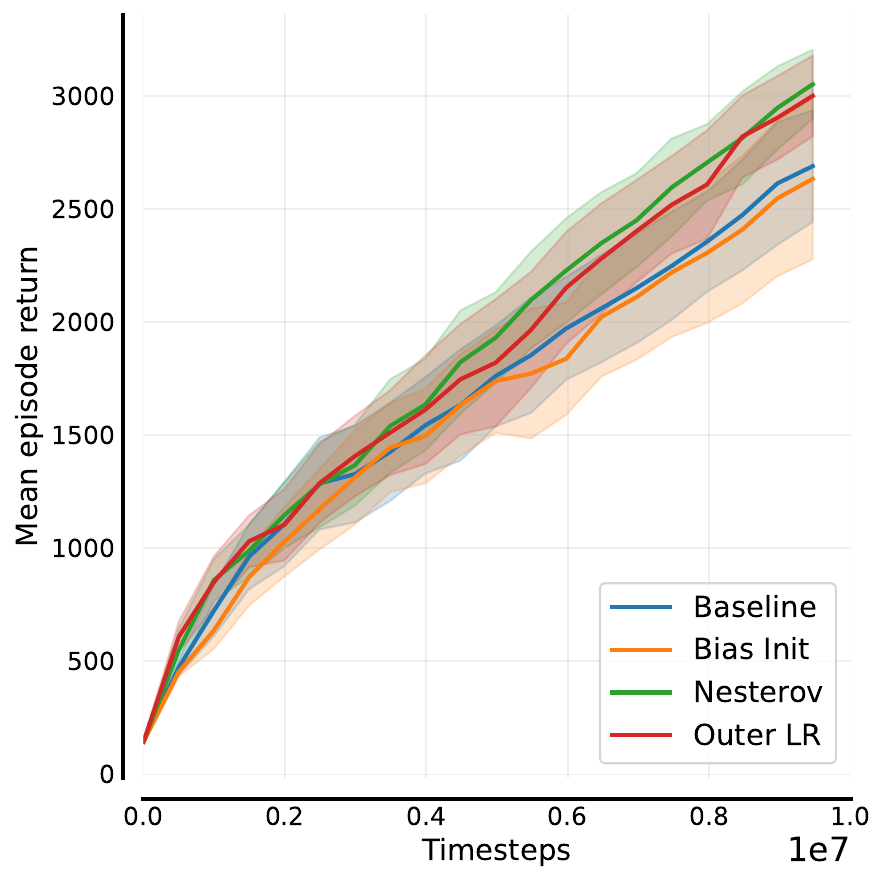}
    \caption{hopper}
\end{subfigure}
\hfill
\begin{subfigure}{0.25\textwidth}
    \includegraphics[width=\textwidth]{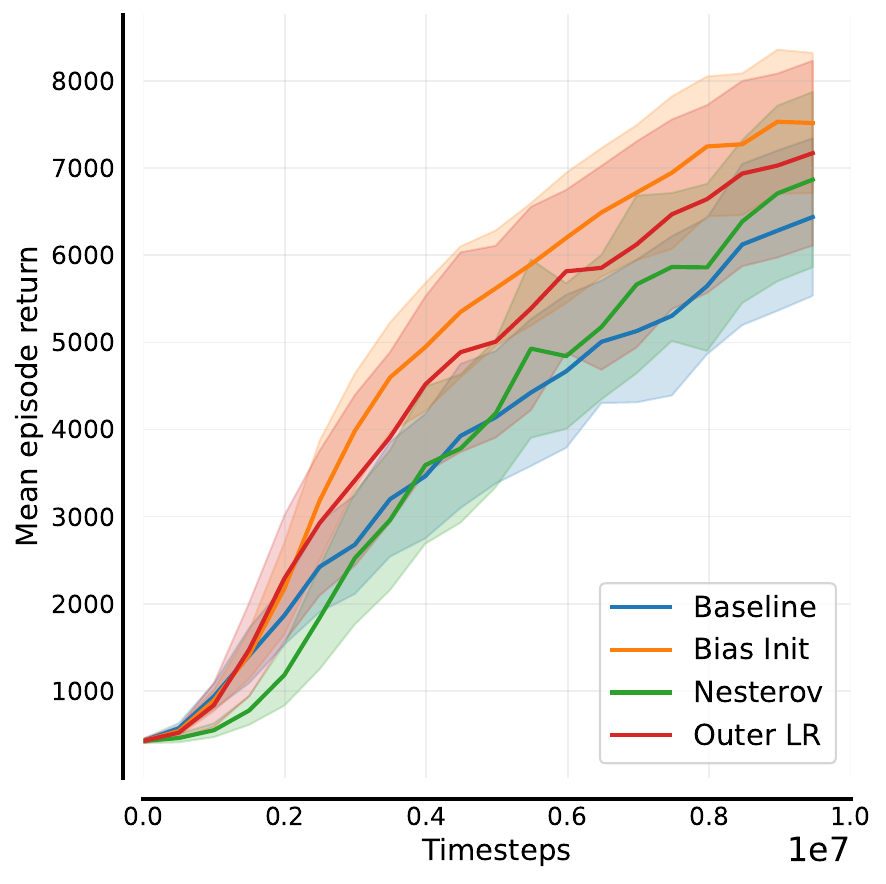}
    \caption{humanoid}
\end{subfigure}
\hfill
\begin{subfigure}{0.25\textwidth}
    \includegraphics[width=\textwidth]{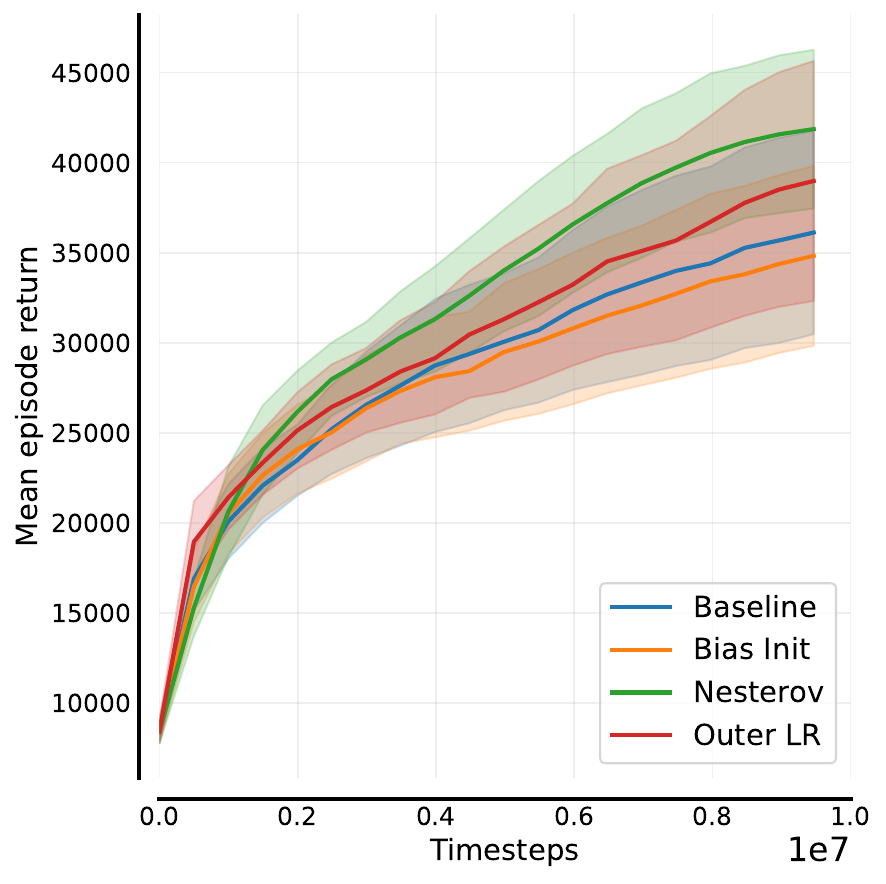}
    \caption{humanoidstandup}
\end{subfigure}
\hfill
\begin{subfigure}{0.25\textwidth}
    \includegraphics[width=\textwidth]{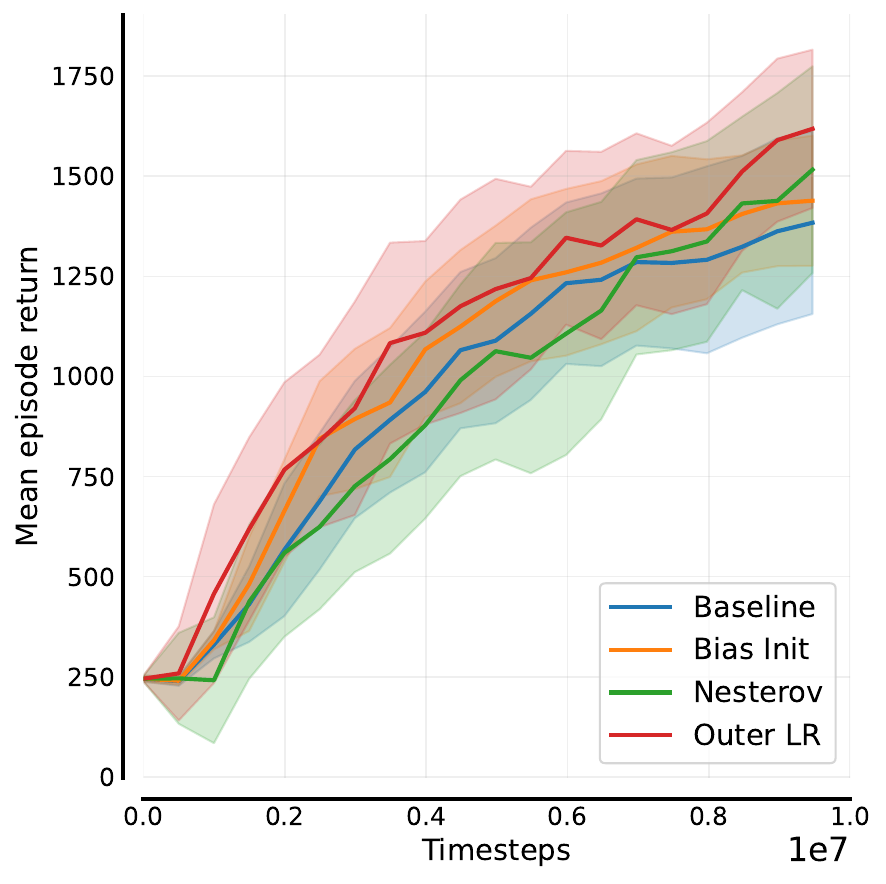}
    \caption{walker\_2d}
\end{subfigure}
    \caption{\textbf{Individual task performance for Brax.} For each task mean of 64 seeds is presented with standard deviation shaded.}
\end{figure}

\begin{figure}[h]
\centering
\begin{subfigure}{0.25\textwidth}
    \includegraphics[width=\textwidth]{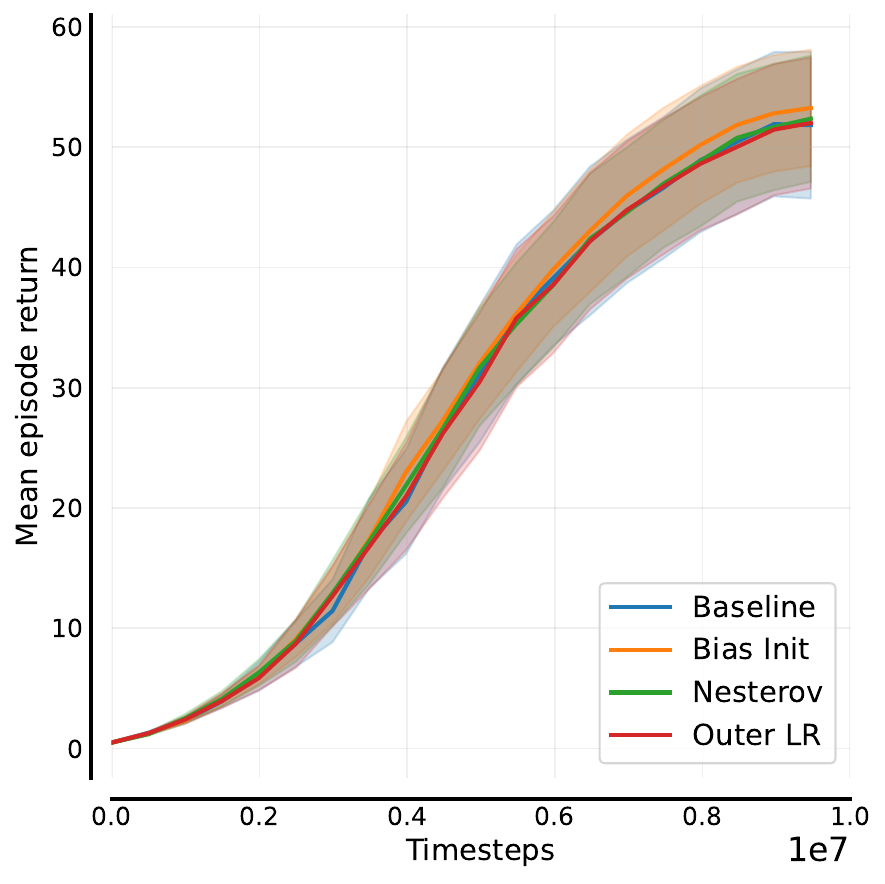}
    \caption{asterix}
\end{subfigure}
\hspace{1.5cm}
\begin{subfigure}{0.25\textwidth}
    \includegraphics[width=\textwidth]{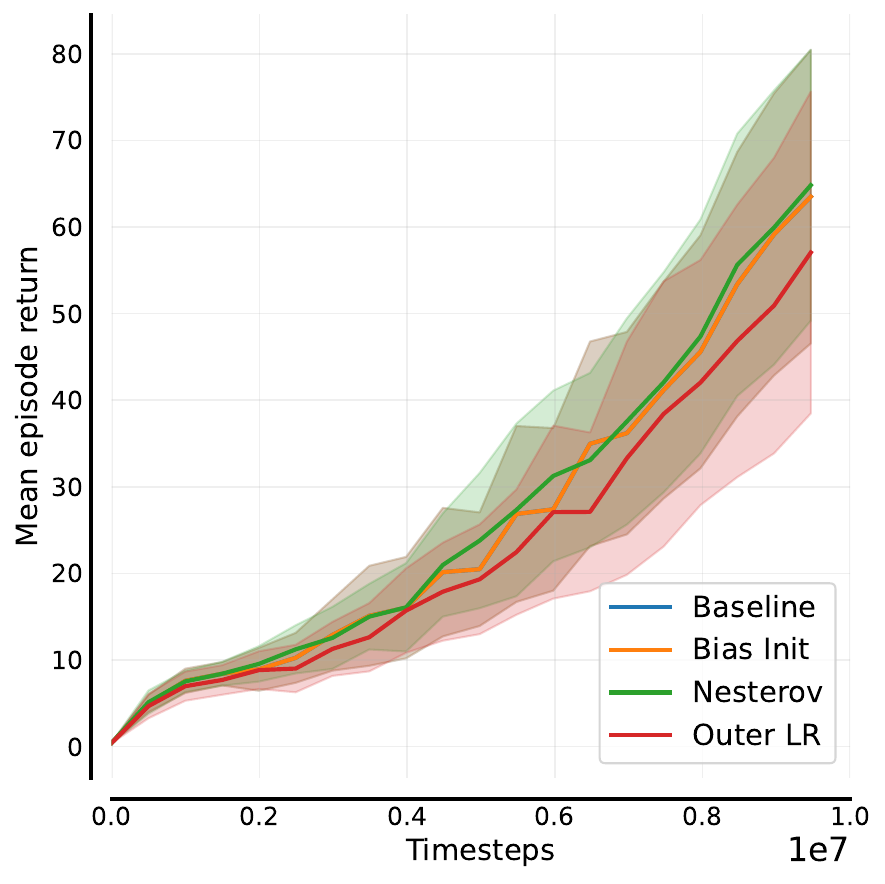}
    \caption{breakout}
\end{subfigure}
\\
\begin{subfigure}{0.25\textwidth}
    \includegraphics[width=\textwidth]{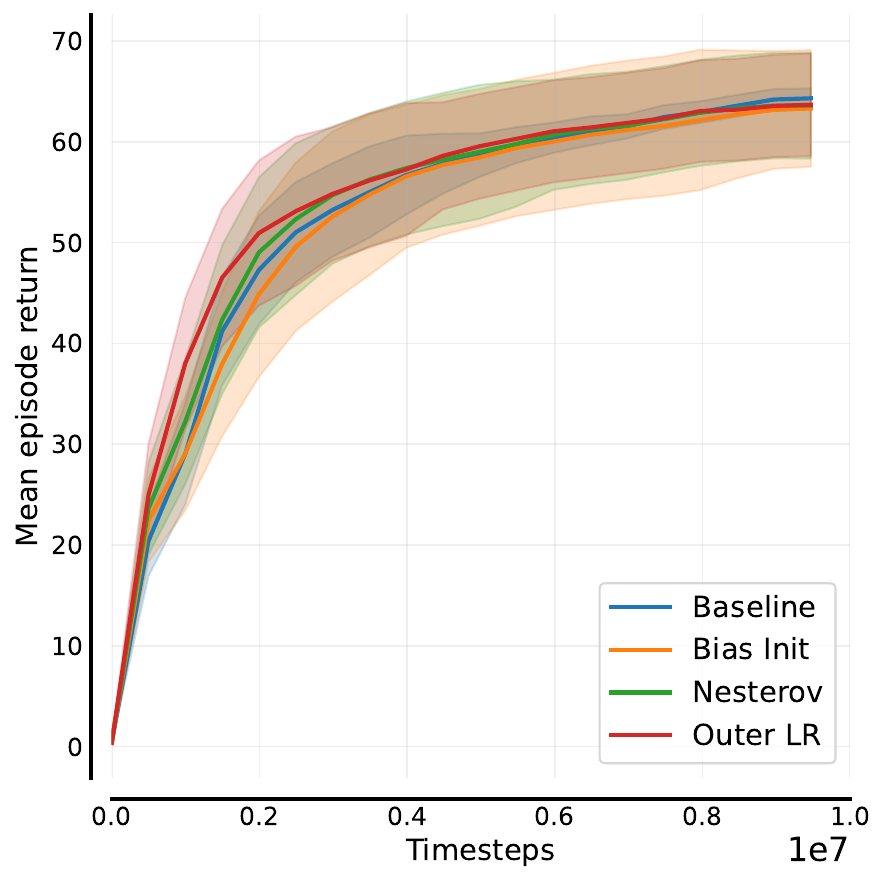}
    \caption{freeway}
\end{subfigure}
\hspace{1.5cm}
\begin{subfigure}{0.25\textwidth}
    \includegraphics[width=\textwidth]{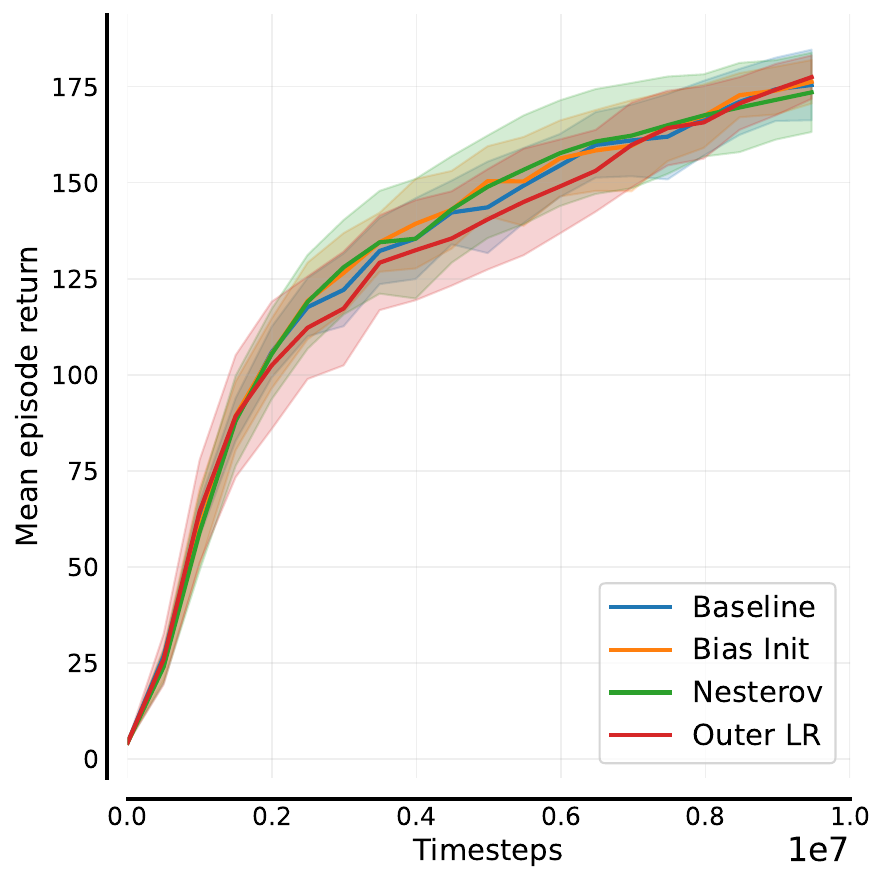}
    \caption{space\_invaders}
\end{subfigure}
    \caption{\textbf{Individual task performance for MinAtar.} For each task mean of 64 seeds is presented with standard deviation shaded.}
\end{figure}

\begin{figure}[h]
\centering
\begin{subfigure}{0.25\textwidth}
    \includegraphics[width=\textwidth]{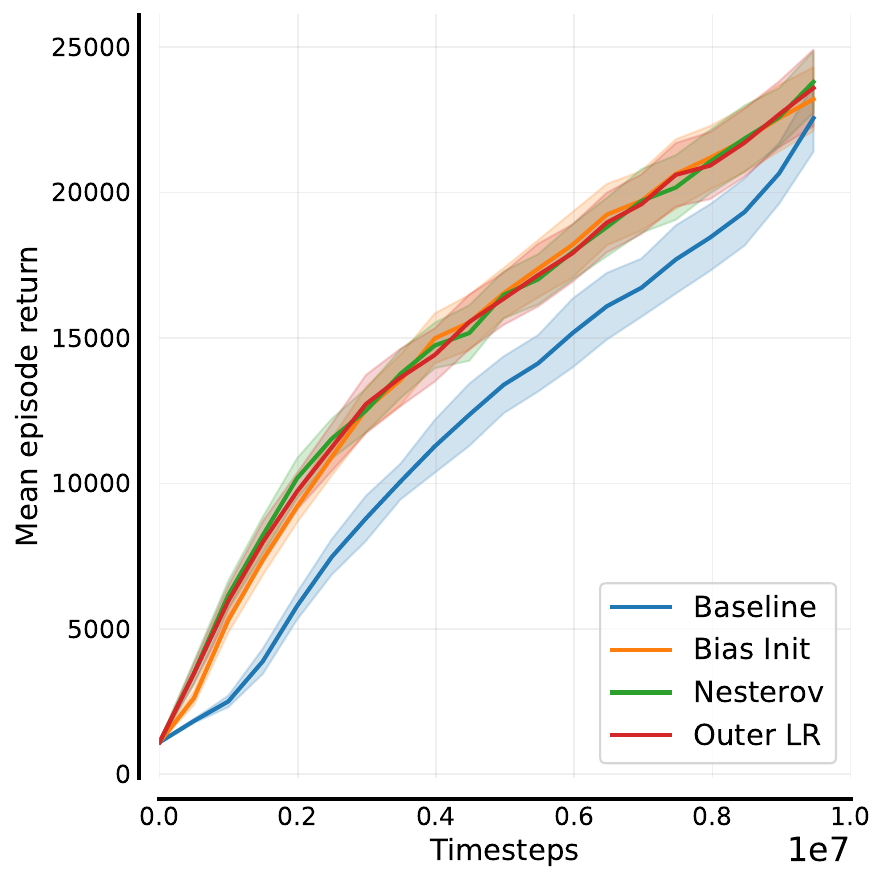}
    \caption{game\_2048}
\end{subfigure}
\hspace{1.5cm}
\begin{subfigure}{0.25\textwidth}
    \includegraphics[width=\textwidth]{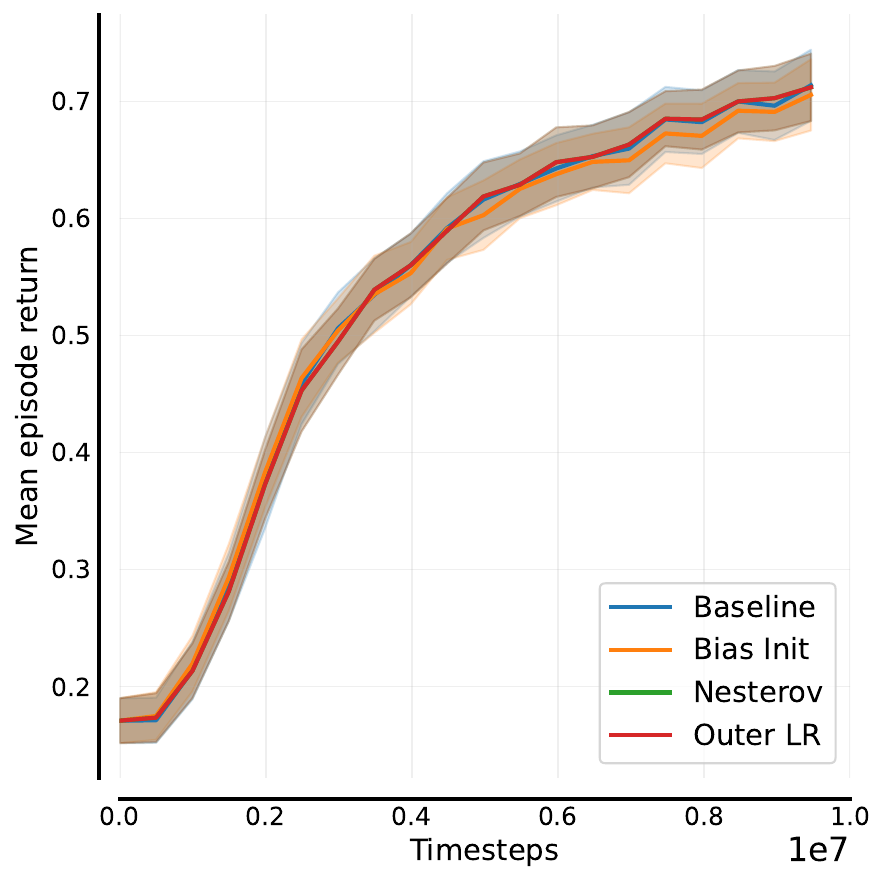}
    \caption{maze}
\end{subfigure}
\\
\begin{subfigure}{0.25\textwidth}
    \includegraphics[width=\textwidth]{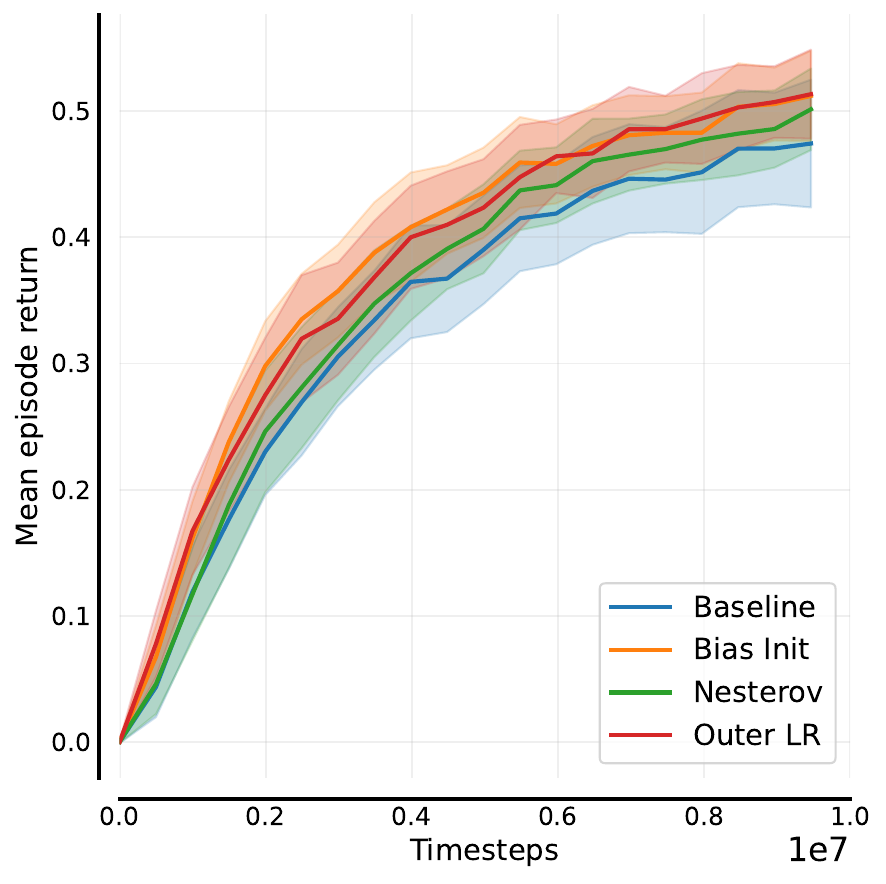}
    \caption{rubiks\_cube}
\end{subfigure}
\hspace{1.5cm}
\begin{subfigure}{0.25\textwidth}
    \includegraphics[width=\textwidth]{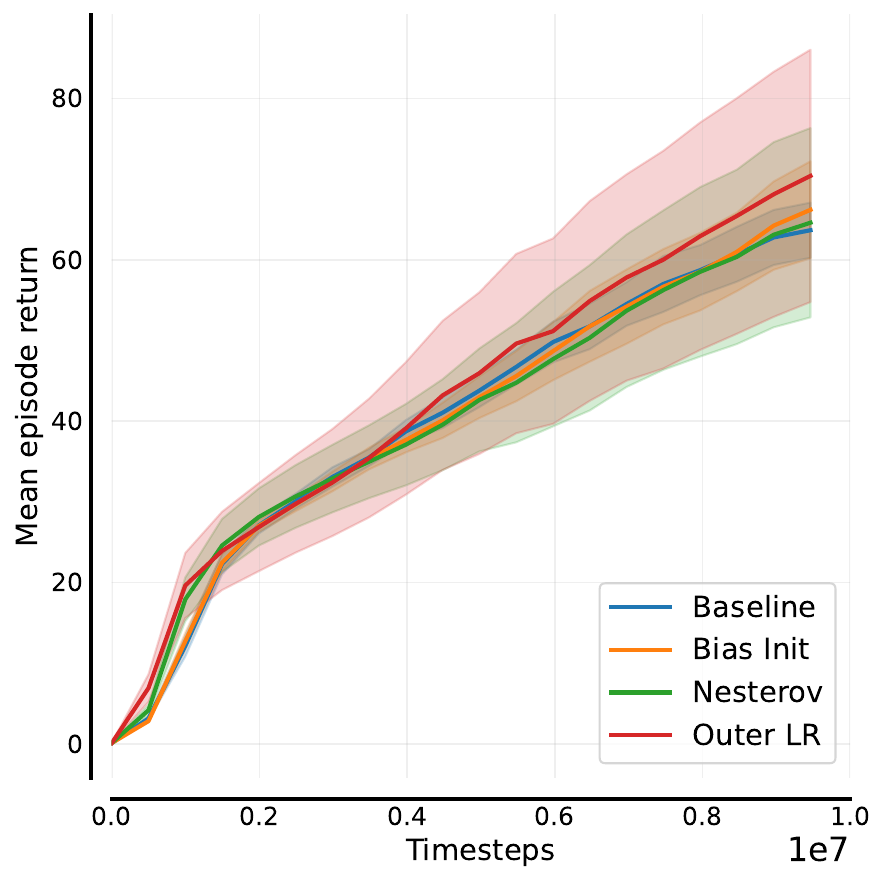}
    \caption{snake}
\end{subfigure}
    \caption{\textbf{Individual task performance for Jumanji.} For each task mean of 64 seeds is presented with standard deviation shaded.}
\end{figure}

\end{document}